\documentclass[10pt]{article} % For LaTeX2e
\usepackage[accepted]{tmlr}
\usepackage{amsmath,amsfonts,bm}

\def\eqref#1{equation~\ref{#1}}
\def\1{\bm{1}}

\DeclareMathAlphabet{\mathsfit}{\encodingdefault}{\sfdefault}{m}{sl}
\SetMathAlphabet{\mathsfit}{bold}{\encodingdefault}{\sfdefault}{bx}{n}

\usepackage[symbol]{footmisc}
\usepackage{hyperref}
\usepackage{url}
\usepackage{amsmath}
\usepackage{amssymb}
\usepackage{mathtools}
\usepackage{amsthm}
\usepackage{pifont}
\usepackage{wrapfig}
\usepackage{tabularx}
\usepackage{xr}
\usepackage{graphicx}
\usepackage{booktabs}
\usepackage{microtype}
\usepackage{graphicx}
\usepackage{subfigure}
\usepackage{adjustbox} 
\usepackage{multirow}
\usepackage[linesnumbered,ruled]{algorithm2e}
\usepackage{algorithmicx}
\usepackage{algcompatible}

\usepackage{booktabs,tabularx,array,graphicx}
\usepackage{subcaption}     % for subfigure panels
\usepackage{pifont}         % ✓ and ✗ symbols
\newcommand{\cmark}{\ding{51}}   % check
\newcommand{\xmark}{\ding{55}}   % cross
\newcolumntype{M}[1]{>{\centering\arraybackslash}m{#1}}
\makeatletter
\newcommand*{\suppressmaintoctoc}{%
  \let\orig@addcontentsline\addcontentsline
  \def\addcontentsline##1##2##3{}%
}
\newcommand*{\restoremaintoctoc}{%
  \let\addcontentsline\orig@addcontentsline
}
\makeatother
\title{M4GN: Mesh-based Multi-segment Hierarchical Graph Network for Dynamic Simulations}

\author{\name Bo Lei \email lei4@llnl.gov \\
      \addr Lawrence Livermore National Laboratory \\
      \AND
      \name Victor M. Castillo \email castillo3@llnl.gov \\
      \addr Lawrence Livermore National Laboratory \\
      \AND
      \name Yeping Hu \email hu25@llnl.gov\\
      \addr Lawrence Livermore National Laboratory \\}

\def\month{09}  % Insert correct month for camera-ready version
\def\year{2025} % Insert correct year for camera-ready version
\def\openreview{\url{https://openreview.net/forum?id=R3vDbqWa1v}} % Insert correct link to OpenReview for camera-ready version

\begin{document}
\suppressmaintoctoc

\maketitle

\begin{abstract}

{Mesh-based graph neural networks (GNNs) have become effective surrogates for PDE simulations, yet their deep message passing incurs high cost and over‑smoothing on large, long‑range meshes; hierarchical GNNs shorten propagation paths but still face two key obstacles: (i) building coarse graphs that respect mesh topology, geometry, and physical discontinuities, and (ii) maintaining fine-scale accuracy without sacrificing the speed gained from coarsening. We tackle these challenges with M4GN—a three‑tier, segment‑centric hierarchical network. M4GN begins with a hybrid segmentation strategy that pairs a fast graph partitioner with a superpixel‑style refinement guided by modal‑decomposition features, producing contiguous segments of dynamically consistent nodes. These segments are encoded by a permutation‑invariant aggregator, avoiding the order sensitivity and quadratic cost of aggregation approaches used in prior works. The resulting information bridges a micro‑level GNN—which captures local dynamics—and a macro‑level transformer that reasons efficiently across segments, achieving a principled balance between accuracy and efficiency. Evaluated on multiple representative benchmark datasets, M4GN improves prediction accuracy by up to 56\% while achieving up to 22\% faster inference than state‑of‑the‑art baselines.}

\end{abstract}

% \begin{figure}[htbp]
%     \centering
%     \includegraphics[width=0.5\textwidth]{image/ICML_first_image.png}
%     \vskip -0.1in
%     \caption{Visualization of multi-scale dynamics—encompassing micro-, meso-, and macro-level interactions—during a solid mechanics simulation from time $t$ to $t+1$. In this scenario, a rigid object is interacting with a hyperelastic beam that is anchored to the ground, resulting in deformation.}
%     \label{fig: first_image}
% \end{figure}

% \begin{wrapfigure}{r}{0.5\textwidth}
%     \centering
%     \includegraphics[scale=0.12]{image/ICML_first_image.png}
%     \caption{Visualization of multi-scale dynamics—encompassing micro-, meso-, and macro-level interactions—during a solid mechanics simulation from time $t$ to $t+1$. In this scenario, a rigid object is interacting with a hyperelastic beam that is anchored to the ground, resulting in deformation.}
%     \label{fig: first_image}
% \end{wrapfigure}

\section{Introduction}\label{introduction}
{Numerically solving partial differential equations (PDEs) to model dynamical systems is fundamental in science and engineering but is often computationally intensive, especially in time-sensitive applications requiring rapid inference. This has prompted increased attention on adopting learning-based surrogate models \citep{sun2020surrogate} to expedite numerical simulations, addressing the computational challenges associated with traditional solvers. Among these methods, mesh-based Graph Neural Network (GNN) methods \citep{pfaff2020learning, gao2022physics, hu2023graph} have proved highly effective for simulating dynamical systems discretized on unstructured meshes. Information is propagated by stacking successive message‐passing layers across mesh edges. When the mesh graph becomes very large—or when the underlying physics couples spatially distant regions (e.g., vortex–vortex interactions in fluids or boundary loads transmitted along an elastic beam)—a substantial number of message-passing iterations is required for information to traverse the graph \citep{fortunato2022multiscale}. This growth in node count and propagation depth inflates computational cost, while the deeper propagation over-smooths node embeddings and erodes accuracy \citep{chen2020measuring,yang2020revisiting,keriven2022not}. To mitigate these effects, recent work has introduced hierarchical GNNs that learn coarse graph representations and pass messages across multiple resolutions, thereby shortening information-propagation paths and reducing the depth required for expressive receptive fields \citep{gao2019graph,li2020multipole,lino2022towards}. Such models have achieved state-of-the-art accuracy on challenging benchmarks and some of them have shown substantial speed-ups over single-scale mesh-based GNNs. Nevertheless, two limitations persist, which we examine in detail in the following subsections.}

%To address the over-smoothing issue, several hierarchical models have been introduced recently with the aim of ......These kind of method have shown promising results....However, two challenges remains for hierarchical GNNs.

 \begin{table}[ht]
  \captionsetup{width=\textwidth}
  \caption{{Comparison of strategies used to construct coarse-level graphs in hierarchical GNN models for physics-based simulations. Here are the definitions for each evaluation criterion: \textit{Heuristic}: the method is  driven by a fixed, hand-crafted rule rather than by parameters learned from data; \textit{Contiguity}: the method produces coarse elements that remain internally connected and never bridge gaps, holes, or material interfaces; \textit{Geometric Fidelity}: the method maintains the original element shapes and sizes closely enough to avoid severe geometric distortion; \textit{Physics-Aware}: the method incorporates physical information—such as material domains, boundary conditions, or regions of nearly uniform fields.}}
  %None of (a–d) embeds material or boundary physics, so the coarse graph can mix incompatible states and degrade surrogate accuracy.}
  \label{tab: pooling_methods}
  \centering
  %–––––––––––––––– The table ––––––––––––––––
  \setlength{\tabcolsep}{4pt}           % tighten horizontal padding
  \begin{tabularx}{\textwidth}{
      >{\raggedright\arraybackslash}p{7.5cm}  % Method
      %>{\raggedright\arraybackslash}p{3cm}                                        % Description
      *{4}{M{2cm}}                           % Three properties
  }
    \toprule
    \textbf{Method} & \textbf{Heuristic}
                    & \textbf{Contiguity} & \textbf{Geometric Fidelity} & \textbf{Physics Aware}\\
    \midrule
    (a) Learnable pooling  \citep{gao2019graph}
                                & \xmark & \xmark & \xmark & \xmark\\
    (b) Spatial proximity pooling  \citep{lino2022towards}
                                & \cmark & \xmark & \cmark & \xmark\\
    (c) Bi-Stride pooling  \citep{cao2023efficient}
                                & \cmark & \cmark & \xmark & \xmark\\
    (d) Same size $k$-means  \citep{janny2023eagle}
                                & \cmark & \xmark & \cmark & \xmark\\
    (e) Hybrid mesh graph segmentation (Ours)  
                                & \cmark & \cmark & \cmark & \cmark\\
    % … add more rows …
    \bottomrule
  \end{tabularx}

  %–––––––––––––––– Small gap, then the merged image ––––––––––––––––
  \vspace{4pt}                         % tweak to taste (0–6 pt is typical)

  \includegraphics[height=2cm,keepaspectratio]{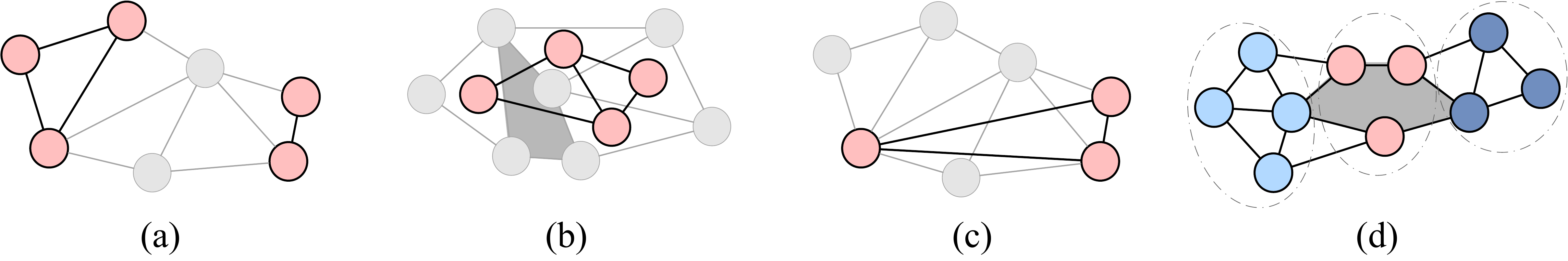}

  % Optional mini-caption for the strip (no “Figure x” number, thanks to \caption*)
  \caption*{\footnotesize
            (a) Learnable pooling can indiscriminately merge distance vertices, disrupting connectivity, and—because it lacks a fixed heuristic—precludes offline sub-graph generation, shifting additional computation into the model runtime; (b) Spatial proximity pooling and (d) Same size $k$-means clustering ignore holes or interfaces (shaded areas), connecting/grouping nodes across gaps; (c) Bi-Stride keeps connectivity but its hop-count stripes warp geometry into elongated, jagged stripes with poor aspect ratios.}
\vspace{-0.2in}
\end{table}
\subsection{Challenge in sub-graph construction}
{One of the key elements of these hierarchical methods is to generate coarse-level graphs, yet existing strategies each have different drawbacks (Table \ref{tab: pooling_methods}). For example, learnable \citep{gao2019graph} or random pooling \citep{li2020multipole} can introduce artificial partitions in the sub-level graphs, which impedes information exchange across partitions. Methods like spatial proximity pooling \citep{lino2022towards, fortunato2022multiscale} can lead to wrong connections across the boundaries at the coarser level. While Bi-Stride \citep{cao2023efficient} does guarantee 2-hop connectivity and avoids cross-boundary edges, its hop-count frontiers can severely warp the geometric metric. Alternatively,  \citep{janny2023eagle} preserves the original mesh and utilizes same size k-means to cluster mesh nodes by treating the mesh graph as point clouds. However, it ignores edge topology entirely: two nodes that are Euclidean-close yet separated by a crack, thin wall, or hole can be grouped together. Because their physical states are incompatible, pooling them into a single ‘super-node’ yields segment embeddings that misrepresent local physics; these distorted features propagate through the model and ultimately amplify prediction error \citep{cao2023efficient, chen2020measuring}. \textit{Hence, the first challenge is to design a graph-coarsening strategy that simultaneously preserves mesh topology and geometric fidelity while respecting physical discontinuities.}}

%The aggregated feature is now an average of disconnected sub-meshes. Nodes that are physically disjoint but forced into a single “super-node” share one feature vector. \YH{Their incompatible states cause... and eventually raising prediction error.} \textit{Therefore, it is necessary to design a segmentation methods that can reduce physics leakage across interfaces and resolve poor locality for message passing \textbf{(Challenge 1)}}.

\subsection{Challenge in balancing accuracy and efficiency}

{Having an appropriate segmentation or pooling strategy addresses only part of the problem; the organization of sub-level graphs into a multi-resolution hierarchy and the way information exchanges between levels are equally decisive for speed and fidelity. Most pooling-based hierarchies \citep{gao2019graph, cao2023efficient} adopt a U-Net-style encoder–decoder \citep{ronneberger2015u}: a lightweight gating function scores vertices, the top-$k$ fraction is retained, and the rest are discarded.  Each pooling step shrinks the graph rapidly, so deeper levels operate on far fewer nodes and enjoy substantial computational savings.  The downside is that aggressive pooling or poorly designed coarsening levels act like an irreversible low-pass (and sometimes aliasing) filter on the graph signal \citep{chen2020measuring, li2020multipole}, so the high-frequency physics “smears or vanishes” when the prediction is mapped back to the fine mesh, resulting in accuracy reduction. By contrast, segment‑based approaches such as EAGLE \citep{janny2023eagle} keep the original mesh intact within each segment, ensuring that no local geometric or physical detail is discarded. However, their reliance on gated recurrent units (GRU) \citep{cho2014learning} means that every node in every segment incurs three gating operations per time step, so computational cost grows with both segment size and feature dimension. Beyond the runtime and memory overhead on large meshes, long GRU chains also suffer from order sensitivity and information dilution \citep{vinyals2015order, bengio1994learning}, which can erode predictive accuracy even when enough compute is available. \textit{The second challenge, therefore, is to design a hierarchical architecture that retains fine‑grained geometric and physical cues like segment‑based methods, yet extracts and propagates segment representations through an aggregator that remains permutation‑invariant, information‑preserving, and computationally light—thus achieving a principled balance between accuracy and efficiency.}}

\subsection{Contributions}
{To overcome the challenges outlined above and to comprehensively evaluate different surrogates, we make three primary contributions in this paper:}
\vspace{-0.15in}
\begin{itemize}
    \item \textbf{Hybrid mesh–graph segmentation.}  
    {We propose a two-stage mesh-graph segmentation strategy: (i) a lightweight graph partitioner first produces a coarse segmentation, and (ii) each partition is then adaptively refined by a superpixel-inspired algorithm guided by modal-decomposition features encoding local physics. The resulting segments preserve contiguity and geometric fidelity while encompassing nodes with coherent physical behavior.}
    \vspace{-0.1in}
    \item \textbf{Multi-segment hierarchical graph network.} 
    {We introduce M4GN, a three‑tier, segment‑centric hierarchy with two crucial innovations compared to previous works: (i) integration of a hybrid mesh–graph segmentation scheme that yields segments of higher geometric and physical fidelity, facilitating communication between adjacent hierarchy levels; and (ii) a permutation‑invariant aggregator for extracting segment features, which is insensitive to mesh node orders and computationally efficient. As a result, M4GN preserves fine‑scale physics (micro-level), compress them into faithful segment tokens (intermediate-level), and enables efficient inter-segment reasoning (macro-level), yielding an effective balance between predictive accuracy and computational efficiency.}
        
    \vspace{-0.1in}
    \item \textbf{Additional dataset and its scaled-up version.} 
    {We contribute DeformingBeam and its scaled counterpart DeformingBeam (large), the first public 3-D Lagrangian contact-deformation benchmark that includes a scale-up version. The meshes’ elongated geometry produces graph diameters several times larger than those in previous solid-mechanics benchmarks, exposing explicit long-range interactions. It enables rigorous testing of hierarchical surrogates and serves as a benchmark for probing model scalability and cross-scale generalization.}

\end{itemize}  

\begin{figure*}[htbp]
    \centering
    \includegraphics[width=0.95\textwidth]{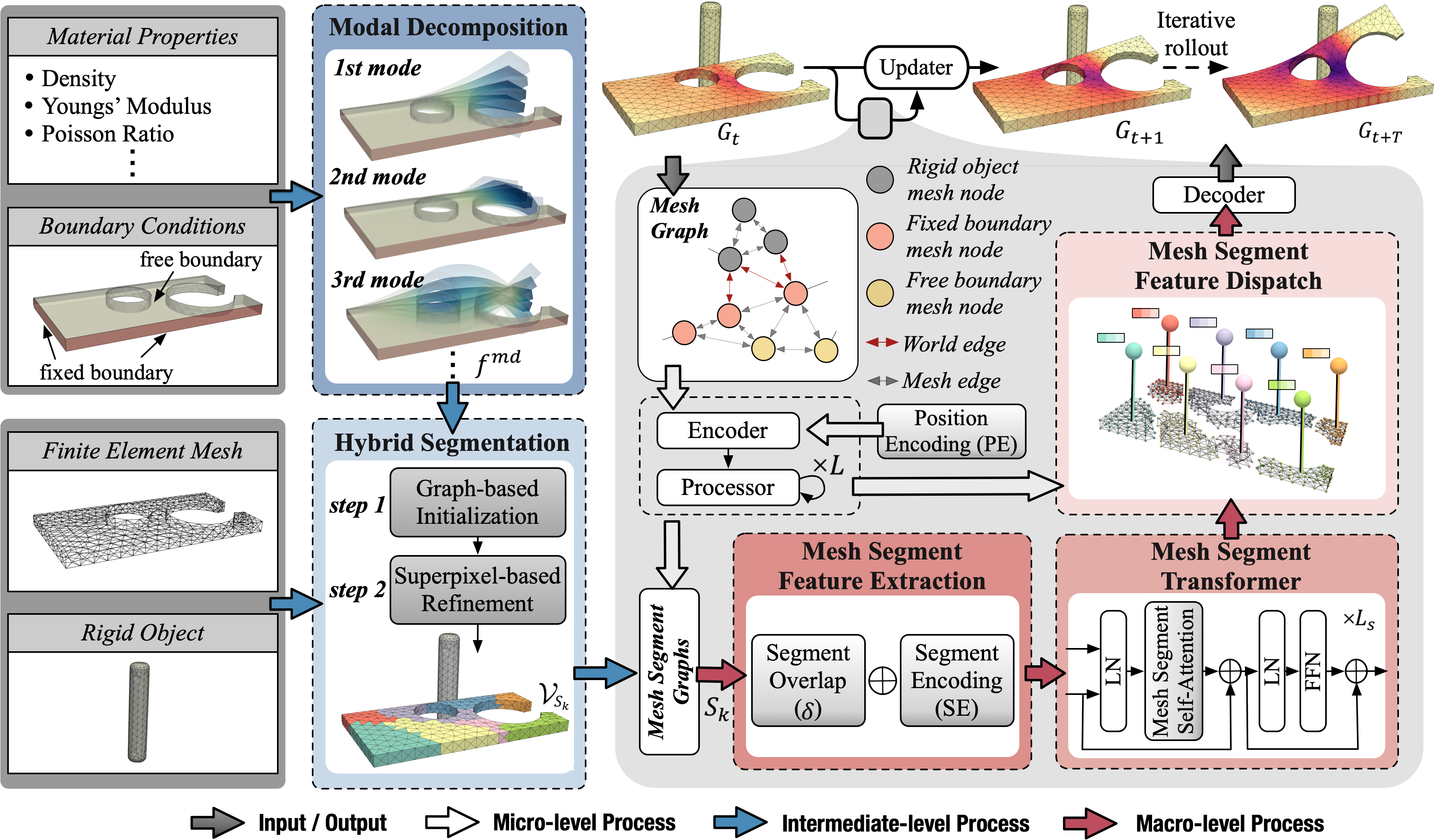}
    \vskip -0.1in
    \caption{{Architecture of the proposed M4GN framework: \textbf{M}esh-based \textbf{M}ulti-Segment Hierarchical (\textbf{M}icro-Intermediate-\textbf{M}acro) \textbf{G}raph \textbf{N}etwork. The framework operates on three modules: micro-level module to capture fine-scale dynamic, intermediate-level module to generate mesh-based segmentation, and macro-level module to model segment-level interactions. The colored arrows trace the data flow within each module.}}
    \label{fig: framework}
\end{figure*}

\section{{Hybrid Mesh-graph Segmentation}}\label{hybrid_segmentation}
This section is organized as follows. Section \ref{segmentation_motivation} motivates the hybrid mesh–graph segmentation approach. Section \ref{segmentation_math_notation} introduces the mathematical notation, and Section \ref{modal_decomposition} illustrates the modal-decomposition technique. Section~\ref{segmentation_methodology} presents the complete segmentation pipeline, which serves as the intermediate-level module of the M4GN framework shown in Figure~\ref{fig: framework}.

\subsection{Motivation}\label{segmentation_motivation}
To avoid the uninterpretable and potentially erroneous dynamics that coarsened graphs or added edges might introduce, we propose preserving the original mesh structure and facilitating long-range information exchange through communication between segmented mesh graphs. Traditional graph segmentation methods \citep{alpert1995spectral, delingette1999general} often prioritize geometric properties and computational efficiency over underlying physical attributes. Conversely, superpixel approaches \citep{veksler2010superpixels, achanta2012slic} group pixels based on user-defined similarity measures but rely on careful cluster-center initialization to maintain segmentation quality. To merge the strengths of both, we apply a \textit{graph-based method} ($f_{gb}$) for initial mesh segmentation and refine it using a \textit{superpixel-based method} ($f_{sb}$), guided by leveraging features associated with dominant modes identified in the modal decomposition module (Section~\ref{modal_decomposition}). In this way, we ensure these segments remain physically coherent and well-structured for effective communication. Grouping elements with similar physical properties enhances model convergence by minimizing discontinuities within each segment \citep{diao2023solving}, while grouping nodes with similar behaviors streamlines learning and ensures uniform handling of similar interactions \citep{dolean2015introduction}. Eventually, this hybrid approach offers efficient and reliable geometric partitioning alongside adaptive, feature-based refinement, producing high-quality mesh segments adaptable to diverse dynamical systems.

\subsection{Mathematical Notation}\label{segmentation_math_notation}
We define the segmentation policy $\pi(G) = f_{s}(G, I)$, where the segmentation function $f_s$ takes the input graph $G$ and prior physical information $I$ (e.g., boundary conditions, material properties), and outputs a set of graph segments $\{S_1^0, S_2^0 \dots, S_K^0\}$. The superscript $0$ denotes non-overlapping segmentation. For each segment $S_k = ({\mathcal{V}_{S_k}}, \mathcal{E}_{S_k})$, the set of nodes $\mathcal{V}_{S_k} \subseteq \mathcal{V}$ and $\mathcal{E}_{S_k} \subseteq \mathcal{E}$ are subsets of the original graph $G$. The union of all segments reconstructs the original graph, such that $\mathcal{V} = \cup \mathcal{V}_{S_k}^0$ and $\mathcal{E} = \cup \mathcal{E}_{S_k}^0$. In some cases, it may be beneficial to allow for overlapping segments, where nodes in $\mathcal{V}$ can belong to more than one segment. This overlap helps create smoother transitions between segments and reduces discontinuities at segment boundaries. We define the overlap amount by $\delta \in \mathbb{N}$, with $\delta = 0$ representing no overlap. For $\delta > 0$, the node set $\mathcal{V}_{S_k}^\delta$ is defined recursively as $\mathcal{V}_{S_k}^\delta = \mathcal{V}_{S_k}^{\delta-1} \cup \{\textit{Adj}(i) \ | \ i \in  \mathcal{V}_{S_k}^{\delta-1}\}$. To simplify the presentation, we disregard the superscript $\delta$ in the remainder of this paper and use $\delta=1$ for all experiments with overlapping. The effect of adding overlapping segments is discussed in our ablation study, as shown in Table~\ref{tab:ablation_mp_hop}.

\subsection{Modal Decomposition}\label{modal_decomposition}
Modal decomposition is a fundamental technique for extracting dominant spatiotemporal patterns, or \emph{modes}, from complex physical systems \citep{fu2001modal, schmid2011applications, taira2017modal}. Each mode encapsulates coherent behavior—such as a characteristic deformation shape or flow structure—allowing a reduced but meaningful representation of the underlying dynamics. In complex physical simulations, these dominant modes can effectively guide downstream tasks such as mesh segmentation, where the domain is subdivided based on physical coherence \citep{yang2016coupling, huang2009shape}. In this work, we employ two different modal decomposition approaches to address \emph{solid} and \emph{fluid} problems separately, given their distinct physical behaviors \citep{bathe2001computational}. The pseudocode of the modal decomposition module can be found in Algorithm~\ref{alg: modal_decomp}.

\textbf{Structural Modal Analysis:}~For solids, the decomposition naturally arises from the mass–stiffness relationship in elastodynamics, capturing genuine dynamic displacements \citep{andersen2006linear}. Let \( \mathbf{K} \) be the global stiffness matrix and \( \mathbf{M} \) the global mass matrix arising from finite element assembly. The free vibration modes of a structure are obtained by solving the generalized eigenvalue problem:
    \begin{equation}\label{eq:eig_solid}
        \mathbf{K}\,\boldsymbol{\phi} \;=\; \lambda \,\mathbf{M}\,\boldsymbol{\phi}\,,
    \end{equation}
    where $\lambda$ represents the square of the natural frequency, and $\boldsymbol{\phi} =(\phi_1,\phi_2,\dots,\phi_{\dim})$ is the corresponding \emph{structural modes}, whose dimension ($\dim$) matches the number of displacement components (e.g., 2D or 3D). Physically, each mode shape indicates a fundamental deformation pattern under vibrational motion \citep{fu2001modal, wilson2002three}, which is tied to the solid’s geometry, boundary conditions, and material parameters. In practice, it is typical to select the first $m$ modes $(\lambda_1 \le \lambda_2 \le \dots \le \lambda_m)$ to construct an $m$-dimensional feature at each mesh node $i$: $f_i^{md} 
        \;=\;
        \bigl(\boldsymbol{\phi}_1(i), \boldsymbol{\phi}_2(i),\dots,\boldsymbol{\phi}_m(i)\bigr).$
        
\textbf{Laplacian Eigenfunctions:}~In fluid contexts, particularly when lacking multiple snapshots or a steady base flow \citep{wang2024recent}, Laplacian eigenfunctions \citep{grebenkov2013geometrical} are used to capture geometry- and boundary-driven harmonic modes by solving:
    \begin{equation}\label{eq:eig_fluid}
        -\,\nabla^2 \phi \;=\; \lambda \,\phi, \; \text{subject to boundary constraints,}
    \end{equation}
    yielding \emph{harmonic modes} \(\phi_1,\dots,\phi_m\). These modes serve as a practical proxy for flow-related structures, providing a minimal but informative decomposition that respects the domain shape and boundary conditions \citep{de2012fluid, taira2017modal}. Similar to the solid case, each node $i$ in the fluid mesh is associated with a feature vector:$
        f_i^{md}
        \;=\;
        \bigl(\phi_1(i),\phi_2(i),\dots,\phi_m(i)\bigr).$

\subsection{Detailed Methodology}\label{segmentation_methodology}
In the hybrid segmentation module, we first use METIS~\citep{karypis1998fast} for initial mesh segmentation due to its great balance of partition quality and speed. Formally, given a graph $G$, the partition function $f_{gb}$ will split it into $K$ non-overlapped mesh-segment graphs: $\{S_1, \dots, S_K \ | \ S_{i} \cap S_{j} = \varnothing, \forall i \neq j\} = f_{gb}(G)$. Then, we apply SLIC~\citep{achanta2012slic}, the state-of-the-art superpixel-based clustering methods, to these mesh segments to iteratively update the segmentation centroids $\{C_1, \dots, C_K \}$ and corresponding node assignments using information obtained from modal decomposition (Section~\ref{modal_decomposition}). It is worth noting that standard modal decomposition does not account for external obstacles \citep{fu2001modal}. Therefore, in models with moving rigid objects, this information will need to be incorporated separately. For node $i$ in graph $G$, we represent it by its spatial coordinates $\mathbf{x}_i$, features related to rigid objects or obstacles $f_i^{{obs}}$, and features obtained from modal decomposition $f_i^{md}$. For a given mesh segment $S_k$ containing $|\mathcal{V}_{S_k}|$ nodes, we define its centroid $C_k$ as its mean value along the features: 
\begin{equation}
C_k = [\mathbf{x}_{C_k}, f^{obs}_{C_k}, f^{md}_{C_k}]^T = \frac{1}{|\mathcal{V}_{S_k}|}\sum_{i \in \mathcal{V}_{S_k}}[\mathbf{x}_i, f^{obs}_i, f^{{md}}_i]^T.
\end{equation}
Within each iteration, we improve the mesh segmentation by minimizing a distance measure that considers {both physical similarity and spatial proximity.} The distance measure $d(i, C_k)$ between a node $i \in \mathcal{V}$ and a segment's centroid $C_k$ is defined as:
\begin{equation}
    d(i, C_k) = \|f^{obs}_i - f^{obs}_{C_k}\| + \|f^{md}_i - f^{md}_{C_k}\| + \tau \|\mathbf{x}_i - \mathbf{x}_{C_k}\|,
\end{equation}
where $\tau$ is used to control the compactness of a mesh segment. The pseudo code of the hybrid segmentation module can be found in Algorithm~\ref{alg: slic}. In Appendix~\ref{appendix: mesh_segmentation}, we present a comprehensive comparison of various segmentation methods and their variants based on different distance measures. Additionally, we evaluate the impact of varying the number of mesh segments on model performance in Appendix~\ref{appendix: ablation_macro_2} and Appendix~\ref{appendix: generalization_num_seg}. {We also introduce several metrics to measure the quality of different mesh segmentations, specifically to understand the intra-segment and inter-segment characteristics, which can be found in Appendix~\ref{appendix: segment_quality_metrics} }

\section{M4GN: Multi-segment Hierarchical Graph Network}\label{methodology}
%We begin this section with a concise formulation of the problem, followed by a detailed description of the three key hierarchical stages of M4GN.
This section details our proposed hierarchical framework (Figure \ref{fig: framework}). After a formal problem statement in Section \ref{problem_definition}, we describe its three-level modules: (i) a micro-level module (Section \ref{Micro_intro}) that performs message passing along mesh edges to capture fine-scale dynamics; (ii) an intermediate-level segment module (Section \ref{hybrid_segmentation}) that constructs mesh segments offline using our heuristic hybrid segmentation algorithm; and (iii) a macro-level module (Section \ref{Macro_intro}) that aggregates segment features and exchanges information across segments to model long-range interactions.

%Meso-level Information Alignment (Section~\ref{Meso_intro}) explains the process of synchronizing mesh structures with the underlying physical properties. Finally, Macro-level Information Exchange (Section~\ref{Macro_intro}) addresses large-scale global interactions that dictate emergent system behaviors.%Taken together, these stages showcase how M4GN integrates hierarchical processes to effectively model and simulate complex dynamic systems.

\subsection{Problem Definition}\label{problem_definition}
Let $G = (\mathcal{V}, \mathcal{E})$ be a mesh graph with $\mathcal{V}$ being the set of nodes and $\mathcal{E}$ being the set of edges. The graph has $N = |\mathcal{V}|$ nodes and $E = |\mathcal{E}|$ edges, with adjacency matrix $A \in \mathbb{R}^{N \times N}$ representing graph connectivity. The dynamic simulation task is to learn a forward model of the dynamic quantities of the mesh graph at the next time step $\hat{G}_{t+1}$ given the current mesh graph $G_t$ and (optionally) a history of previous mesh graphs $\{G_{t-1}, \dots, G_{t-h}\}$. Finally, the rollout trajectory can be generated through the simulator iteratively based on the previous prediction: $G_t, \hat{G}_{t+1}, \dots, \hat{G}_{t+T}$, where $T$ is the total simulation steps. In this paper, the proposed model (M4GN) can simulate both Eulerian and Lagrangian systems \citep{bontempi1998lagrangian}. {For Eulerian systems examined in this paper, where continuous fields such as velocity evolve on a stationary mesh, the graph $\mathcal{E}$ includes only mesh-related edges $\mathcal{E}^M$. Conversely, for Lagrangian systems considered in this paper, where the mesh represents a moving and deforming surface or volume, additional world edges $\mathcal{E}^W$ are incorporated into the graph.} These edges enable the model to learn external dynamics such as collision and contact. The node features of node $i$ are denoted by $\mathbf{x}_{i}$, while the features for an edge between node $i$ and $j$ are indicated by $\mathbf{e}_{ij}$.

% \subsection{Physics-informed Hierarchical Architecture}
\subsection{Micro-level Module}
\label{Micro_intro}
Within the micro-level module, each node engages in the exchange of information with its neighboring nodes. This process holds particular significance in dynamical systems, where the behavior of adjacent nodes is closely intertwined \citep{booij1987propagation, emanuel1994atmospheric, fahy2007sound, kennett2009seismic}. Furthermore, this module serves a crucial role in addressing discontinuities that may arise at the boundaries of adjacent mesh segments \citep{lai2009introduction}. By prioritizing micro-level information exchange, we effectively mitigate discontinuities introduced by subsequent macro-level operations. %For detailed information, refer to Section~\ref{MGN}.
%\subsubsection{Mesh-based Graph Neural Networks} \label{MGN}
{We follow the Encoder-Process-Decoder (EPD)\citep{pfaff2020learning} architecture for our micro-level information exchange as it has shown great performance in dealing with mesh-based graphs. Specifically, we use the encoder and processor blocks for micro-level message passing, while the decoder head is detached and applied only after the macro-level module (Section~\ref{Macro_intro}).} For a given graph $G_t$ at time $t$, the model begins with extracting node and edge features through two separate Multi-Layer Perceptrons (MLPs):
\begin{equation}\label{eq:MGN_encoding}
    \mathbf{h}_{i,t}^{0} = f_{n}(\mathbf{x}_{i,t}), \quad \mathbf{h}_{ij,t}^{M,0} = f_{e}^M(\mathbf{e}_{ij,t}^M), \quad \mathbf{h}_{ij,t}^{W,0} = f_{e}^W(\mathbf{e}_{ij,t}^W),
\end{equation}
where $\mathbf{x}_{i,t} \in \mathcal{V}$, $\mathbf{e}_{ij,t}^M \in \mathcal{E}^M$, and $\mathbf{e}_{ij,t}^W \in \mathcal{E}^W$ denote node feature, mesh edge feature, and world edge feature vector at time $t$, respectively. For Lagrangian systems, world edges are created by spatial proximity, where for a fixed radius $r_W$, a world edge is added between nodes $i$ and $j$ when $|\mathbf{x}_i - \mathbf{x}_j| < r_W$, excluding node pairs already connected in the mesh. The outputs of two MLPs (i.e. $f_{n}$ and $f_{e}$) for node and edge are denoted as $\mathbf{h}_{i,t}^{0}$ and $\mathbf{h}_{ij,t}^{0}$, respectively. Then, a $L$-step message passing (MP) is performed such that each node can receive and aggregate information from neighboring nodes within $L$ steps of edge traversing. For each MP from 1 to $L$, the node and edge representations are updated as:
\begin{align}
    \mathbf{h}_{i,t}^{l} = f_{n}^{l}(\mathbf{h}_{i,t}^{l-1}, \sum_{j\in\textit{Adj}(i)}\mathbf{h}_{ij,t}^{M,l-1}, \sum_{j\in\textit{Adj}(i)}\mathbf{h}_{ij,t}^{W,l-1}), \\
    \mathbf{h}_{ij,t}^{M,l} = f_{e}^{l}(\mathbf{h}_{ij,t}^{M,l-1}, \mathbf{h}_{i,t}^{l-1}, \mathbf{h}_{j,t}^{l-1}), \quad
    \mathbf{h}_{ij,t}^{W,l} = f_{e}^{l}(\mathbf{h}_{ij,t}^{W,l-1}, \mathbf{h}_{i,t}^{l-1}, \mathbf{h}_{j,t}^{l-1}),
\end{align}
where $\textit{Adj}(i)$ denotes all adjacent nodes of node $i$. Up until this point, the node and edge information of the graph $G_t$ are updated. Additionally, we implement a technique from \citep{godwin2021simple}, which involves corrupting the input graph with noise and adding a noise-correcting node-level loss. We evaluate the impact of varying the number of message passing steps during micro-level information exchange step, where details can be found in Appendix~\ref{appendix: ablation_micro}. 

% \subsection{Meso-level Information Alignment}\label{Meso_intro}
% In the meso-level information alignment stage, the model facilitates interactions among clusters of nodes, capturing intermediate-scale structures that bridge local and global dynamics. This level is crucial for representing transitional processes where groups of interacting components form coherent substructures within the mesh.

\subsection{Macro-level Module} \label{Macro_intro}
%{At the macro-level information exchange stage, the model captures long-range dependencies and global interactions that govern the emergent behavior of the entire system. This stage is essential for understanding large-scale trends and patterns that arise from the collective dynamics of the system’s components \citep{svedin2005micro}. The macro-level module aggregates information from the meso-level, allowing the model to synthesize comprehensive system-wide insights and make predictions that reflect the overall state of the dynamic system. }

\subsubsection{Mesh Segment Feature Extraction}\label{Feature_Extraction}
\textbf{Segment Encoding (SE)} -- In order to extract a global feature for each mesh segment, we perform average pooling on all node vectors in $S_k$ and apply a MLP ($f_s$) to get the fixed-sized segment embedding: $
\mathbf{h}_{S_k,t} = f_s(\frac{1}{|\mathcal{V}_{S_k}|}\sum_{i \in \mathcal{V}_{S_k}} \mathbf{h}_{i,t}^{L})$. {Different from the state-of-the-art clustering-based hierarchical model \citep{janny2023eagle}, we replace the GRU-based aggregator with a permutation-invariant pooling operation when extracting segment features, which provides a more efficient and robust alternative for segment-level representation. It sidesteps the order bias and gradient noise of sequence models, yielding representations that remain stable across epochs and insensitive to mesh reordering \citep{vinyals2015order}. Moreover, average pooling circumvents the information‑dilution problem inherent in long RNN chains \citep{bengio1994learning}, preserving salient local features from vanishing‑gradient effects and thereby enhancing predictive accuracy. Meanwhile, its computational burden is only $O(Nd)$, compared with the $O(Nd^2)$ matrix operations and lengthy back‑propagation demanded by GRU pooling \citep{chung2014empirical}. These properties make average pooling a proper choice when accuracy and efficiency need to be balanced.}

%By preserving salient extremes without the information dilution typical of long RNN chains \citep{bengio1994learning}, max pooling retains critical local cues while scaling gracefully to large, irregular meshes. 

\textbf{Position Encoding (PE)} --
As dynamic effect propagates continuously over mesh domains, knowing relative location among segments could provide extra information for next-step macro-level information exchange and increase expressivity of the network. Mathematically, for each pair of mesh segment graph, $\{S_i, S_j\}$, their relative positional information can be obtained through segment-level adjacency matrix $A^K \in \mathbb{R}^{K \times K}$: $A^K_{S_i S_j} = \sum_{m \in \mathcal{V}_{S_i}}\sum_{n \in \mathcal{V}_{S_j}}A_{mn}$. We follow the strategy in \citep{rampavsek2022recipe} that uses random-walk structural encoding (RWSE) \citep{dwivedi2021graph} for PE calculation. Then the PE for the $k$-th segment, denoted as $\mathbf{p}_{S_k,t}$, is processed through an MLP layer ($f_{sp}$) and then added to update the SE as follows: $\mathbf{h}_{S_k,t} \leftarrow \mathbf{h}_{S_k,t} + f_{sp}(\mathbf{p}_{S_k,t})$. We can further enhance the network's expressivity by adding absolute PE to the graph nodes. We use an MLP ($f_{np}$) to process each node's PE ($\mathbf{p}_{i,t}$), calculated with a similar approach as segment level, and add it to the input node feature. Thus, Eq~(\ref{eq:MGN_encoding}) becomes $\mathbf{h}_{i,t}^{0} = f_{n}(\mathbf{x}_{i,t} + f_{np}(\mathbf{p}_{i,t}))$. By incorporating node PE directly into the input features, these features participate in the micro-level information exchange described in Section~\ref{Micro_intro}, potentially improving the continuity of the extracted mesh segment features. Table~\ref{tab:ablation_PE} presents ablation results illustrating the impact of including or excluding PE on prediction performance.
% As each segment has different shapes, each node would have different contribution to the patch segment depends on its relative location within the segment. Therefore, a random walk position embedding (RWPE) \cite{xx} is utilized to obtain a 
% \begin{itemize}
%     \item Talk about PE algorithm and how to add extra edge for obstacle to apply PE
% \end{itemize}

\subsubsection{Mesh Segment Transformer, Feature Dispatch, and Training} \label{Transformer}
We construct a fully connected mesh segment graph, where the $i$-th mesh segment feature is represented by $\mathbf{h}_{S_i}$ . 
% \YH{The edge that connect two mesh segments $S_i$ and $S_{j}$ has edge feature represented as $\mathbf{e}_{S_i S_{j}}$. Our proposed transformer layer propagates messages through edges that connect mesh segments. As the obstacle is defined as a single mesh segment, it also has edges connected to each mesh segment of the deforming object. \textcolor{red}{[Bo, can you help modify the blue-colored text above?]}} 
Note that since the transformer operates on mesh segments rather than individual mesh nodes, and the total number of mesh segments ($K$) is significantly smaller than the total number of mesh nodes ($N$), the computational cost of our transformer is substantially reduced compared to a traditional graph transformer that operates on graph nodes (i.e. $O(K^2) \ll O(N^2)$). The $l$-th block of the mesh segment transformer layer is defined as follows:
\begin{equation}
    %\mathbf{a}_{S_i S_j}^{k,l} &= \text{softmax}_{S_j}\biggl(\frac{\mathbf{Q}_h^{k,l}\text{LN}(\mathbf{h}_{S_i}^l) \cdot \mathbf{K}_h^{k,l}\text{LN}(\mathbf{h}_{S_j}^l) }{\sqrt{d_h}}\biggr), \\%\mathbf{W}_h^{k,l}\text{LN}(\mathbf{e}_{S_i S_j}^{l-1}), \\
    \Bar{\mathbf{h}}_{S_i}^l = \parallel_{k=1}^{H} \sum_{j=1}^K \mathbf{a}_{S_i S_j}^{k,l}(\mathbf{V}_h^{k,l}\text{LN}(\mathbf{h}_{S_j}^l)), \quad
    \mathbf{h}_{S_i}^{l+1} = \mathbf{h}_{S_i}^l + \mathbf{O}_h^l \Bar{\mathbf{h}}_{S_i}^l + \text{FFN}_h^l(\text{LN}(\mathbf{h}_{S_i}^l + \mathbf{O}_h^l \Bar{\mathbf{h}}_{S_i}^l)),
\end{equation}
where $\mathbf{a}_{S_i S_j}^{k,l}$ is self-attention weight between $S_i$ and $S_j$, $\mathbf{V}_h^{k,l} \in \mathbb{R}^{d_h \times d}$ is a trainable parameter matrix, and $\mathbf{O}_h^l \in \mathbb{R}^{d \times d}$ is the learned output project matrix. $k$ = 1 to $H$ denotes the number of attention heads, and $\parallel$ denotes concatenation. $d_h$ is the dimension of mesh segment feature for each head, and $d$ is the input and output dimension. We adopt a Pre-Layer Norm architecture \citep{xiong2020layer}, which is denoted as $\text{LN}(\cdot)$, and the point-wise Feed Forward Network is represented as FFN$(\cdot)$. The mesh segment transformer module facilitates information exchange among all mesh segments, updating the feature of each segment $\mathbf{h}_{S_i}$ after passing through $L_S$ mesh segment transformer blocks. 

The mesh segment feature dispatch module (as shown in Figure~\ref{fig: framework}) integrates information obtained from both macro-level and micro-level exchanges. Specifically, the final feature for node $i$ at time step $t$ is updated as $\mathbf{h}_{i,t} \leftarrow [\mathbf{h}_{i,t}, \mathbf{h}_{S_i, t}]$ where $i \in \mathcal{V}_{S_i}$. This ensures that each node incorporates information from both neighboring mesh nodes and spatially distant, yet correlated regions. Finally, we train our dynamics model by supervising on the per-node output features $\hat{\mathbf{x}}_{i, t+1}$, produced by feeding $\mathbf{h}_{i,t}$ into an MLP-based decoder, using an $L_2$ loss between  $\hat{\mathbf{x}}_{i, t+1}$ and the corresponding ground truth values  ${\mathbf{x}}_{i, t+1}$. 
% \begin{wrapfigure}{r}{0.5\textwidth}
%     \centering
%     \includegraphics[scale=0.2]{./image/main_text_rollout_horizontal_new_reduced.pdf}
%     \vskip -0.1in
%     \caption{Comparison of prediction results across different models, with each node in the plots color-coded according to its RMSE error over $t$-step rollouts. Result shows that M4GN achieves notably lower RMSE errors in areas where other methods struggle, particularly at later time steps and in regions further from the inlet or contact point.}
%     \label{fig: rollout_main_text}
% \end{wrapfigure}

% \begin{wrapfigure}{r}{0.5\textwidth}
%     \centering
%     \includegraphics[scale=0.12]{image/ICML_first_image.png}
%     \caption{Visualization of multi-scale dynamics—encompassing micro-, meso-, and macro-level interactions—during a solid mechanics simulation from time $t$ to $t+1$. In this scenario, a rigid object is interacting with a hyperelastic beam that is anchored to the ground, resulting in deformation.}
%     \label{fig: first_image}
% \end{wrapfigure}
%\subsubsection{Mesh Segment Feature Dispatch and Training}

\section{Experiment}
\subsection{Experiment Setup}
\textbf{Datasets} --
{We benchmark on six datasets that together capture the two principal application regimes for mesh–based surrogates—Eulerian incompressible flow and Lagrangian hyper-elastic solids—while also probing both short-range and long-range interactions, small and large graphs, and steady versus highly-transient behavior. \textit{CylinderFlow} and \textit{DeformingPlate} are the widely-used public datasets of \citep{pfaff2020learning}. CylinderFlow varies cylinder diameter, position and inlet velocity, while DeformingPlate changes obstacle trajectory, plate geometry, and boundary conditions. To challenge long-range coupling we introduce the \textit{DeformingBeam}, a 3-D hyper-elastic beam whose length, cross-section, and end loads vary, yielding the largest graph diameter in the suite. Moreover, \textit{DeformingBeam-Large} doubles the physical span of that beam and the total number of meshes to probe scalability. {Finally, we utilize two supplementary benchmarks: \textit{FlagSimple} \citep{pfaff2020learning} and \textit{EAGLE} \citep{janny2023eagle} to further demonstrate the robustness of M4GN across distinct physical regimes and dataset types. For these two datasets, we report results in the Appendix and do not perform full ablation analysis to keep the experimental scope tractable and to maintain clarity in interpreting results.} This mix exercises both short-range and long-range interactions, small and large graphs, and steady versus highly-transient behavior. Because geometry, loading and obstacle parameters all change from run to run, the sets jointly test a surrogate’s ability to generalize across shape, boundary condition and scale variations, giving a balanced yet concise benchmark suite.} 
 
At each time step, the network is provided with (i) nodal coordinates and edge vectors that describe the local mesh geometry, (ii) the current physical state, and (iii) categorical node masks identifying walls, inlets/outlets, or fixed/handle/obstacle regions. For the modal-decomposition stage, we further supply the material parameters and boundary conditions already defined in the finite-element setup (density, Young’s modulus, Poisson’s ratio). The network outputs the next-step state (e.g., positions for solids) which is recursively fed back as input to generate full rollouts. Comprehensive dataset specifications are provided in Appendix \ref{appendix: dataset}.

%We use two public datasets from \citep{pfaff2020learning}: \textit{CylinderFlow} (fluid flows around a cylinder) and \textit{DeformingPlate} (elastic plate deformed by an actuator). We also create a new dataset, \textit{DeformingBeam}, featuring a hyperelastic beam deformed by an actuator in a 3D mesh. Details of the datasets can be found in the appendix \ref{appendix: dataset}. We create \textit{DeformingBeam} dataset for three major reasons: (1) This dataset exhibits long-range interactions, with the largest graph diameter compared to the other two datasets (Table \ref{tab: dataset_basics}). (2) The inclusion of diverse mesh structures significantly increases the complexity of the underlying physics, making the task more challenging. Theses datasets span the two dominant practical settings for unstructured-mesh dynamics surrogates: Eulerian incompressible flow ( CylinderFlow ) and Lagrangian hyper-elastic solids ( DeformingPlate , DeformingBeam ). We also include a doubled-size DeformingBeam (Large) set to stress-test scalability. This mix exercises both short-range and long-range interactions, small and large graphs, and steady versus highly-transient behavior. Because geometry, loading and obstacle parameters all change from run to run, the sets jointly test a surrogate’s ability to generalize across shape, boundary-condition and scale variations, giving a balanced yet concise benchmark suite.

\textbf{M4GN and Baselines} --
As a default configuration for our M4GN model, we use 7 message passing steps in the mesh graph network. The mesh segment transformer adopts 4 self-attention layers with 8 heads. We compare our method to five baseline models: 1) \textit{GCN} \citep{kipf2016semi, belbute2020combining}, a basic GNN structure widely used for simulating fluid dynamics; 2) \textit{g-U-Nets} \citep{gao2019graph, alsentzer2020subgraph}, a representative method that incorporates graph pooling modules to enhance long-range interactions; 3) \textit{MeshGraphNets} (MGNs) \citep{pfaff2020learning}, a single-level GNN architecture that achieves great performance and generalizability across various dynamical systems; 4) \textit{BSMS-GNN} \citep{cao2023efficient}, a recent work featuring a multi-level hierarchical GNN architecture that aims to enhance computational efficiency in simulating physical systems; and 5) \textit{EAGLE}\citep{janny2023eagle}, a recent work presenting a clustering-based pooling method along with transformer to enhance performance on large-scale turbulent fluid dynamics. Detailed descriptions of the these models and training procedures can be found in Appendix~\ref{appendix: model_baselines}.

% \textbf{MMSGN Configuration} -- As a default configuration for our MMSGN model, we use 7 message passing steps in the mesh graph network. The mesh segment transformer adopts 4 self-attention layers with 8 heads. Other model details and hyperparameters can be found in Appendix~\ref{appendix: model_baselines}. The training details can be found in Appendix~\ref{appendix: model_baselines}.

\textbf{Metrics} --
%To evaluate prediction accuracy, we compute the Root Mean Square Error (RMSE) of the model’s output, averaging it across output dimensions, mesh nodes, time steps, and the entire test dataset. Specifically, we calculate errors of each method for a single prediction step (1-step), 50-step rollouts, and rollout of the whole trajectory.
{In addition to traditional accuracy metrics, we also report mesh-quality metrics (Table~\ref{tab: mesh_quality_metric}) for the Lagrangian cases, where the computational mesh deforms with the material and may accumulate distortion. These evaluations are not required for the Eulerian dataset we used, because its meshes remain fixed in space and therefore cannot experience element-quality degradation. Mesh‐quality metrics serve objectives that a nodal RMSE alone cannot address: (i) Many workflows feed the predicted mesh into inverse design or topology-optimization loops \citep{han2012surrogate}; if elements are inverted or highly skewed, gradient-based updates fail even when the nodal RMSE is low. (ii) Well-shaped, smoothly graded elements improve the numerical conditioning of the surrogate, yielding higher predictive accuracy and better generalization \citep{kamenski2014conditioning}. Four mesh quality metrics are used with concise descriptions in Table~\ref{tab: mesh_quality_metric}, where each metric targets a distinct failure mode, so a model may excel in one yet falter in another. For instance, a mesh can align perfectly with the true geometry while still containing many sliver elements. Evaluating all four metrics together thus offers a genuinely comprehensive view of the surrogate’s output quality. More detailed descriptions and mathematical definitions of these metrics can be found in Appendix ~\ref{appendix: additional_mesh_quality_metrics}.}

\begin{table}[ht]
\centering
\caption{{Compact summary of mesh–quality metrics for evaluating prediction results}}
\label{tab: mesh_quality_metric}
\setlength{\tabcolsep}{4pt}          %  tighter column padding
\renewcommand{\arraystretch}{1.0}    %  tighter line spacing
\footnotesize
\begin{tabularx}{\textwidth}{@{}p{0.19\textwidth}  p{0.22\textwidth}  p{0.19\textwidth}  p{0.34\textwidth}@{}}
%\begin{tabularx}{\textwidth}{@{}l X X X@{}}

\toprule
\textbf{Metric} & \textbf{Geometric intuition} & \textbf{What it catches} & \textbf{Impact on simulation outcomes} \\ \midrule
%\begin{tabular}[c]{@{}l@{}} Hausdorff  ($\text{GF}_h$)  \\ \citep{huttenlocher1993comparing} \end{tabular}  &
Hausdorff  ($\text{GF}_h$) & Max node–surface gap & Isolated large outliers & Drives worst-case error; signals local shape failures. \\[1pt]
%\begin{tabular}[c]{@{}l@{}} Chamfer  ($\text{GF}_c$)  \\ \citep{wu2021balanced} \end{tabular} &
Chamfer  ($\text{GF}_c$) & Mean node–surface gap & Uniform global drift & Lowers overall fidelity; raises global error norms. \\[1pt]

%\begin{tabular}[c]{@{}l@{}} Mesh Continuity (MC) \\ \citep{knupp2007remarks} \end{tabular}&
Mesh Continuity (MC) & Neighbour cell–volume ratio & Abrupt size “cliffs” & Introduces artificial discontinuities; noisy spatial gradients. \\[1pt]

%\begin{tabular}[c]{@{}l@{}} Aspect-Ratio error (AR) \\ \citep{zienkiewicz2005finite} \end{tabular} &
Aspect-Ratio error (AR) & Deviation from ideal element shape & Slivers / stretched cells & Injects anisotropic bias; hampers generalization. \\ \bottomrule
\end{tabularx}
\end{table}

%Maintaining mesh quality is crucial in these systems because changes in mesh elements over time can lead to numerical errors and misrepresentation of dynamic behaviors. 
\subsection{Results and Discussion}
\subsubsection{Overall Performance Evaluation Across Multiple Datasets}
The quantitative results in Table~\ref{tab: overall_results} show that M4GN outperforms all baselines across multiple evaluation metrics. Specifically, for the CylinderFlow dataset, M4GN achieves a 36\% reduction in test RMSE-all compared to the second-best performing model, EAGLE. For the DeformingPlate dataset, M4GN reduces the test RMSE-all by 32\%. This improvement is even more pronounced for DeformingBeam dataset, where M4GN demonstrates a 56\% reduction in test RMSE-all. Such performance in 50-step and longer-step predictions underscores its enhanced capability for long-term predictions. In addition to its high prediction accuracy, M4GN demonstrates strong mesh quality, with up to a 48\% reduction in GF and a 14\% reduction in MC compared to the second-best model across both Lagrangian system datasets.

\subsubsection{Segmentation Quality and Its Impact on Performance Metrics}
{To rigorously assess the quality of our hybrid segmentation strategy and its influence on dynamics prediction, we employ three metrics—\textit{Conductance}, \textit{Edge-Cut Ratio}, and \textit{Silhouette Score}—that quantify intra-segment cohesion and inter-segment separation. These metrics assess segment isolation and intra-segment homogeneity properties that prevent feature dilution and lower prediction error. Formal definitions appear in Appendix \ref{appendix: segment_quality_metrics}. Figure \ref{fig: segmentation_analysis_main}(a–c) correlates segmentation scores with mesh quality and prediction error for EAGLE’s node‐based partitions and three variants of our hybrid scheme (see Appendix \ref{appendix: mesh_segmentation}). The plots confirm that segmentation choice matters: our hybrid method, which better aligns partitions with underlying dynamic behaviors, consistently reduces error and improves mesh quality. Figure \ref{fig: segmentation_analysis_main}(d–f) color nodes by each segment’s mean prediction error at successive time steps. Our segments stay nearly uniform in color across all datasets, indicating coherent intra-segment dynamics. For example, in
(d), segmentation follows periodic wave patterns in fluid dynamics, while in (f), it reflects symmetrical system dynamics with symmetric segment coloring. These visualizations demonstrate that our segmentation effectively captures the temporal and spatial dynamics of the system, outperforming state-of-the-art method. Additional in-depth analysis can be found in Appendix \ref{appendix: ablation_meso_1} and \ref{appendix: ablation_meso_2}.} 

\begin{table*}[t]
\caption{Comparison of results with state-of-the-art methods across three datasets, where each model is trained independently for each dataset. Prediction accuracy is evaluated using Root Mean Square Error (RMSE), with the output being the next-step positions for DeformingPlate/Beam and velocity/pressure fields for CylinderFlow. Errors are reported for 1-step rollout, 50-step rollouts, and the entire trajectory. {Each mesh quality metric is evaluated at every time step then averaged over time—by comparing the predicted mesh to the ground‑truth configuration.} Results are averaged over three experiments with different random seeds and presented as mean and standard deviation.}
\label{tab: overall_results}
\begin{adjustbox}{width=\textwidth}
%\vskip 0.15in
\begin{scriptsize}
\begin{sc}
{
\begin{tabular}{llccccccc} 
\toprule
  &  & \multicolumn{4}{c}{\textnormal{Mesh Quality Metrics $\downarrow$} }  &\multicolumn{3}{c}{\textnormal{Prediction Error Metrics $\downarrow$}}   \\
 \cmidrule(lr){3-6} \cmidrule(lr){7-9} 
Dataset  & Model & \begin{tabular}[c]{@{}c@{}} $\text{GF}_h$ \\ ($\times 10^{-3}$) \end{tabular} & \begin{tabular}[c]{@{}c@{}} $\text{GF}_c$ \\ ($\times 10^{-6}$) \end{tabular}& \begin{tabular}[c]{@{}c@{}} MC \\ ($\times 10^{-3}$) \end{tabular} & \begin{tabular}[c]{@{}c@{}} AR \\ ($\times 10^{-3}$) \end{tabular} & \begin{tabular}[c]{@{}c@{}} RMSE-1 \\ ($\times 10^{-5}$) \end{tabular} &  \begin{tabular}[c]{@{}c@{}} RMSE-50 \\ ($\times 10^{-4}$) \end{tabular}  &  \begin{tabular}[c]{@{}c@{}} RMSE-\textnormal{all} \\ ($\times 10^{-4}$) \end{tabular} \\
\midrule
\midrule
\multirow{6}{*}{\begin{tabular}[c]{@{}l@{}} Cylinder\\Flow \end{tabular}} 
& GCN        & -  & - & -  & - & 675 \tiny$\pm$ 28 & 382 \tiny$\pm$ 69 & 1702 \tiny$\pm$ 310\\
& $g$-U-Net    & -  & -  & -  & - & 401 \tiny$\pm$ 3.7 & 179 \tiny$\pm$ 32 & 758 \tiny$\pm$ 88\\
& MGN     & -  & - & - & - & 246 \tiny$\pm$ 13 & 62.8 \tiny$\pm$ 2.8 & 412 \tiny$\pm$ 48\\
& BSMS-GNN   & -  & -    & - & - &  \textbf{181 \tiny$\pm$ 22} & 252 \tiny$\pm$ 8.2 & 1218 \tiny$\pm$ 83\\
& EAGLE       & -  & -   & - & - &  456 \tiny$\pm$ 24 & 69.6 \tiny$\pm$ 3.0 & 525\tiny$\pm$ 24\\
& M4GN (Ours) & -  & -  & - & -&  288 \tiny$\pm$ 19 & \textbf{60.2 \tiny$\pm$ 2.4} & \textbf{337 \tiny$\pm$ 21} \\

% & GCN        & -  & - & -  & - & 764 \tiny$\pm$ 32 & 425 \tiny$\pm$ 82 & 1887 \tiny$\pm$ 358\\
% & $g$-U-Net    & -  & -  & -  & - & 423 \tiny$\pm$ 4 & 199 \tiny$\pm$ 37 & 843 \tiny$\pm$ 141\\
% & MGN     & -  & - & - & - & 274 \tiny$\pm$ 15 & 64.4 \tiny$\pm$ 3.4 & 481 \tiny$\pm$ 53\\
% & BSMS-GNN   & -  & -    & - & - &  \textbf{202 \tiny$\pm$ 24} & 280 \tiny$\pm$ 9 & 1373 \tiny$\pm$ 90\\
% & EAGLE       & -  & -   & - & - &  507 \tiny$\pm$ 25 & 71.5 \tiny$\pm$ 3.2& 583\tiny$\pm$ 29\\
% & M4GN (Ours) & -  & -  & - & -&  320 \tiny$\pm$ 29 & \textbf{63.6 \tiny$\pm$ 2.6} & \textbf{372 \tiny$\pm$ 27} \\
\midrule
\multirow{6}{*}{\begin{tabular}[c]{@{}l@{}} Deforming\\Plate \end{tabular}}
& GCN            & 24.0 \tiny$\pm$ 0.6 & {323 \tiny$\pm$ 4} & 11.0 \tiny$\pm$ 0.3 & {9.33 \tiny$\pm$ 0.57} &  34.8 \tiny$\pm$ 0.6 & 26.1 \tiny$\pm$ 0.1 & 169 \tiny$\pm$ 1\\
& $g$-U-Net     &  36.1 \tiny$\pm$ 8.5 &  {452\tiny$\pm$ 125} & 20.1 \tiny$\pm$ 0.5 & {12.4 \tiny$\pm$ 4.3} &  41.2 \tiny$\pm$ 0.2 & 30.4 \tiny$\pm$ 0.8 & 179 \tiny$\pm$ 7\\
& MGN     & 12.7 \tiny$\pm$ 0.9 & {248 \tiny$\pm$ 12} & 9.25 \tiny$\pm$ 0.39 & {5.34 \tiny$\pm$0.26} &  \textbf{22.8 \tiny$\pm$ 0.2} & 20.0 \tiny$\pm$ 0.4 & 147 \tiny$\pm$ 3\\
& BSMS-GNN      & 23.8 \tiny$\pm$ 2.6 & {170 \tiny$\pm$ 13} & 18.3 \tiny$\pm$ 4.4 & {15.4 \tiny$\pm$ 5.9} &  30.3 \tiny$\pm$ 5.6 & 23.7 \tiny$\pm$ 3.5 & 118 \tiny$\pm$ 4 \\
& EAGLE         & 6.75 \tiny$\pm$ 0.8 & {41.1 \tiny$\pm$ 2.6} & 5.56 \tiny$\pm$ 0.12  & {3.31 \tiny$\pm$ 0.04} & 36.4 \tiny$\pm$ 5.2 & 5.63 \tiny$\pm$ 1.7 &  38.7 \tiny$\pm$ 1.8  \\
& M4GN (Ours)  & \textbf{4.29 \tiny$\pm$ 0.07}& {\textbf{7.05 \tiny$\pm$ 1.05}} & \textbf{4.82 \tiny$\pm$ 0.06} & {\textbf{2.67 \tiny$\pm$ 0.06}} &  {26.7 \tiny$\pm$ 0.5} & {\textbf{3.03 \tiny$\pm$ 0.16}} & {\textbf{26.5 \tiny$\pm$ 2.4}} \\
\midrule
\multirow{6}{*}{\begin{tabular}[c]{@{}l@{}} Deforming\\Beam \end{tabular}}
& GCN          & 4.91 \tiny$\pm$ 0.36 & {3.53 \tiny$\pm$ 0.51} & 54.8 \tiny$\pm$ 8.2 & {69.5 \tiny$\pm$ 3.8} &  7.25 \tiny$\pm$ 0.12 & 5.08 \tiny$\pm$ 0.11 & 30.7 \tiny$\pm$ 4.1 \\
& $g$-U-Net    & 4.91 \tiny$\pm$ 0.50 & {3.55 \tiny$\pm$ 0.73} & 34.7 \tiny$\pm$ 1.8 & {31.5 \tiny$\pm$ 1.2} &  7.28 \tiny$\pm$ 0.39 & 5.09 \tiny$\pm$ 0.23 & 31.7 \tiny$\pm$ 4.0 \\
& MGN    & 0.82 \tiny$\pm$ 0.04 & {0.12 \tiny$\pm$ 0.01} & 16.9 \tiny$\pm$ 0.1& {7.43 \tiny$\pm$ 0.10} &  4.43\tiny$\pm$ 0.08 & 2.41 \tiny$\pm$ 0.16& 4.72 \tiny$\pm$ 0.27\\
& BSMS-GNN      & 0.99 \tiny$\pm$ 0.03 & {0.21 \tiny$\pm$ 0.04}  & 32.5 \tiny$\pm$ 0.5 & {16.1 \tiny$\pm$ 0.3} & 6.86 \tiny$\pm$ 0.09 & 1.95 \tiny$\pm$ 0.22 & 4.98 \tiny$\pm$ 0.71 \\
& EAGLE       & 0.64 \tiny$\pm$ 0.04 & {0.17 \tiny$\pm$ 0.01} & 5.98 \tiny$\pm$ 0.43 & {5.17 \tiny$\pm$ 0.37} &   1.51 \tiny$\pm$ 0.04 & 0.67 \tiny$\pm$ 0.12 & 4.22 \tiny$\pm$ 0.30\\
& M4GN (Ours)         & \textbf{0.31 \tiny$\pm$ 0.01} & {\textbf{0.05 \tiny$\pm$ 0.00}} & \textbf{5.26 \tiny$\pm$ 0.04} & {\textbf{3.08 \tiny$\pm$ 0.06 }} & {\textbf{1.17 \tiny$\pm$ 0.01}} & {\textbf{0.34 \tiny$\pm$ 0.02}} & {\textbf{1.87 \tiny$\pm$ 0.12}}\\

\bottomrule
\end{tabular}}
\end{sc}
\end{scriptsize}
\end{adjustbox}
%\vskip -0.2in
\end{table*}

\begin{figure*}[htbp]
    \centering
    \includegraphics[width=1\textwidth]{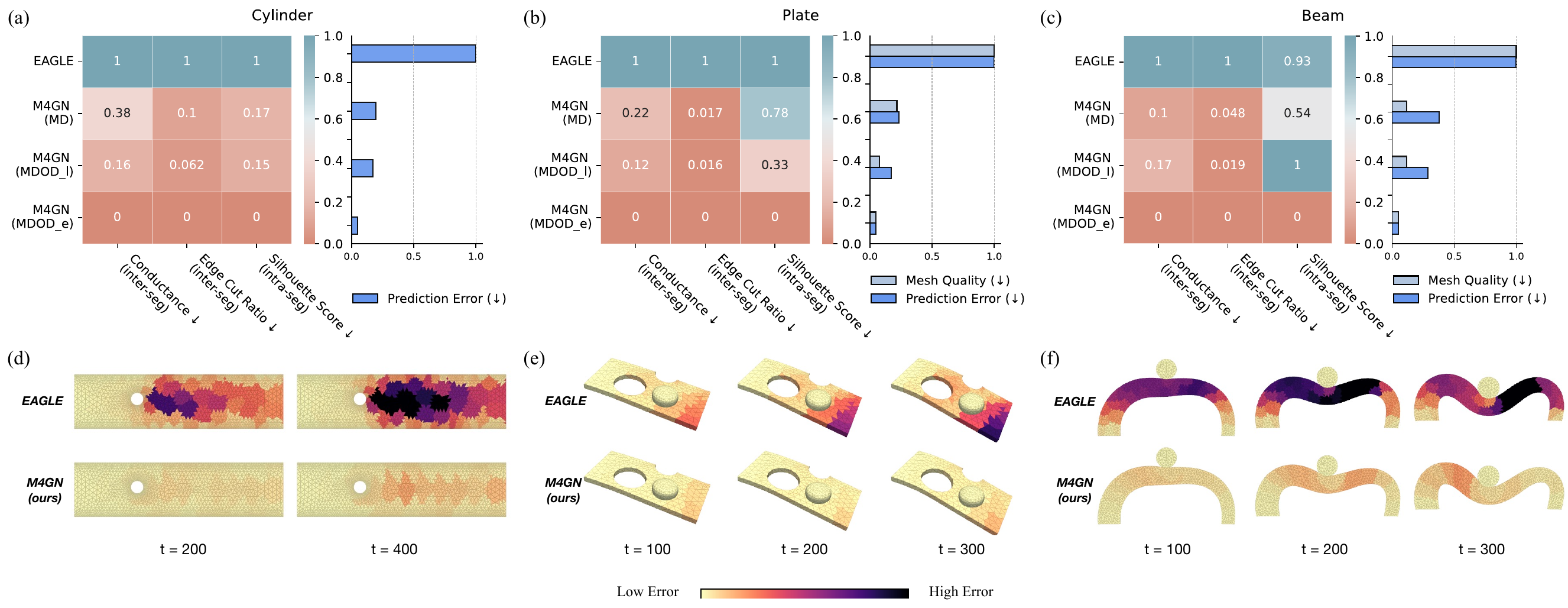}
    %\vskip -0.1in
    \caption{{(a-c) Evaluation of different segmentation methods under three datasets. The heatmap (left) presents normalized Conductance, Edge Cut Ratio, and reversed Silhouette Score for EAGLE and three M4GN variants. Metrics are scaled between 0 and 1, where lower values indicate better segmentation quality. The sidebar plot (right) depicts normalized Prediction Error and Mesh Quality, with a minimum value of 0.05 applied to avoid invisible bars. (d-f) Visualization of simulation rollouts over time for three datasets, comparing our segmentation method with EAGLE. Nodes are colored based on the average prediction error within their segments.}}
    \label{fig: segmentation_analysis_main}
    \vskip -0.2in
\end{figure*}

\subsubsection{Accuracy–Efficiency Trade‑off Analysis}
\begin{figure*}[htbp]
    \centering
    \includegraphics[width=0.87\textwidth]{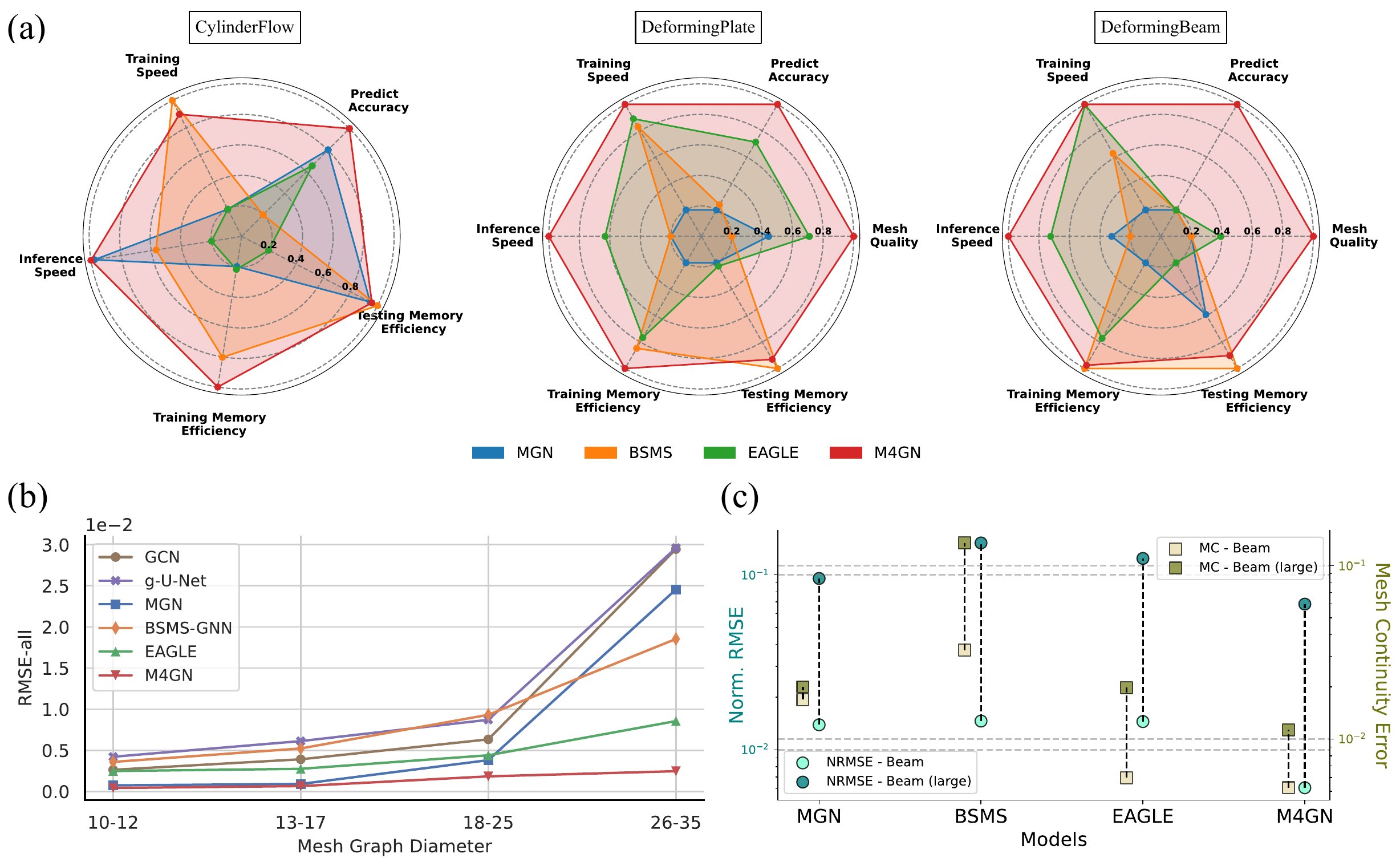}
    \vskip -0.1in
    \caption{(a) {Radar charts summarizing model performance on three datasets. All metrics are min‑max normalized to the range $[0.2, 1.0]$; "lower‑is‑better" metrics are first inverted, and values below 0.2 are clipped to avoid visual collapse; larger filled areas reflect better overall performance.} (b) RMSE-all  accuracy versus graph diameter on the DeformingPlate dataset. (c) Performance comparison of each model when generalizing from DeformingBeam to DeformingBeam (large). {Circles denote normalized RMSE (left axis), and squares denote mesh continuity error (right axis). Dashed lines highlight performance shifts. Lower values indicate better accuracy and mesh quality.}}
    \label{fig: combined_results_new}
    \vskip -0.2in
\end{figure*}

Figure~\ref{fig: combined_results_new}(a) shows distinct trade‑offs: MGN excels on the small‑diameter CylinderFlow but loses accuracy and speed on DeformingPlate and DeformingBeam, where deep message passing leads to over-smoothing and higher latency; BSMS is memory‑efficient thanks to bi‑stride pooling, yet that pooling lowers mesh fidelity and accuracy and slows inference when long‑range details must be 
reconstructed; and EAGLE offers moderate performance overall—its physics‑agnostic clustering limits long‑range capture, remaining efficient on the some datasets but slowing sharply on CylinderFlow as dense meshes inflate the number of required clusters. {By contrast, M4GN maintains high accuracy and efficiency across all cases because its hybrid segmentation groups physically coherent nodes and its three‑level hierarchy offloads long‑range reasoning to a lightweight segment‑level transformer, reducing both message‑passing depth and token count. This design lets M4GN preserve local fidelity while controlling computational cost, yielding the balanced performance seen in the plots. Additional qualitative results are provided in Table~\ref{tab: ablation_complexity}.}

%In contrast, across all three datasets, M4GN covers the greatest area, indicating an optimal trade‑off between accuracy and computational cost while maintaining high mesh quality. 

%M4GN achieves the largest filled areas across all three datasets, demonstrating high prediction accuracy, superior mesh quality, and strong computational efficiency. This is due to its physically aligned hierarchical structure, which integrates micro-level local interactions with macro-level global exchanges to capture both short- and long-range dynamics. 

%The physics-guided segmentation ensures nodes within segments exhibit similar behaviors, while macro-level exchanges efficiently handle long-range dependencies. This alignment with physical dynamics enables M4GN to maintain high accuracy without compromising efficiency, making it highly effective for dynamic system simulation. 

%The mesh segment transformer’s effectiveness in facilitating efficient macro-level information exchange demonstrates its potential for large-scale 3D dynamics modeling, where memory and computational efficiency are critical concerns \cite{bartoldson2023scientific}.

\subsubsection{Performance Analysis Across Graph Diameter and Scale}
{One of the main goals of hierarchical GNNs is to alleviate over‑smoothing and capture long‑range interactions. Exploiting the wide diameter range in our DeformingBeam benchmark, we group test cases by graph diameter and plot RMSE‑all versus diameter in Figure~\ref{fig: combined_results_new}(b). Errors for the baseline models escalate rapidly with increasing diameter, whereas M4GN’s error increases only slightly, showing sustained accuracy on wide meshes with long‑range interactions. This robustness comes from M4GN’s hybrid segmentation, which clusters nodes that share modal behavior, and its segment‑level transformer, which achieves global information exchange in one hop among segments; together, these mechanisms mitigate over-smoothing while preserving local detail, sustaining performance as graph diameter grows. Moreover, we perform generalization tests to show whether a surrogate trained on modest meshes stays reliable on larger domains. According to Figure~\ref{fig: combined_results_new}(c), all four surrogates deteriorate when directly generalize to larger mesh domain. The modest degradation of MGN confirms that a flat model can generalize spatially if the growth factor is moderate, but only at the expense of longer inference times and an elevated risk of over‑smoothing as graph diameter continues to rise. Hierarchical methods trim that cost, yet their performance hinges on how they coarsen the mesh and whether cross‑level communication remains efficient. Among them, M4GN’s smaller generalization gap indicates that its hybrid segmentation and permutation‑invariant aggregation alleviate—but do not eliminate—this sensitivity, suggesting future work on adaptive segment counts or depth‑aware micro-level passes when extrapolating to substantially larger meshes.}

\subsection{Additional Studies}
We conducted additional studies to comprehensively evaluate model performance, hyperparameter selection, and the impact of key architectural designs, with detailed results and discussions provided in the appendices. {Results on additional datasets are reported in Appendix~\ref{appendix: supplementary_dataset_results}, and the segmentation‑effectiveness study appears in Appendix~\ref{appendix: segmentation_transfer}. A systematic sensitivity study, detailing how each hyperparameter affects performance and offering practical tuning guidelines, can be found in Appendix~\ref{appendix: ablation}.} Additionally, a comprehensive analysis of generalization performance is provided in Appendix~\ref{appendix: generalization}, and further insights into computational efficiency are included in Appendix~\ref{appendix: complexity}.

\section{Related Works}
%\subsection{GNNs for Dynamic System Simulation}
%The application of Graph Neural Networks (GNN) for dynamic system prediction is an emerging research area in scientific machine learning due to their versatility and effectiveness \citep{mrowca2018flexible, belbute2020combining, rubanova2021constraint}. Unlike image-based learning methods such as Convolutional Neural Networks (CNNs) \citep{um2018liquid, ummenhofer2019lagrangian}, GNNs can directly handle unstructured simulation meshes, making them well-suited for simulating systems with complex domain boundaries while ensuring spatial invariance and locality \citep{battaglia2018relational, wu2020comprehensive}. The initial application of GNNs to physics-based simulations focused on deformable solids and fluids, with MeshGraphNets (MGN) being a pioneering work in this area \citep{pfaff2020learning}. Building on this foundation, various MGN variants have been proposed: integrating GNNs with Physics-Informed Neural Networks (PINNs) \citep{gao2022physics}, enabling long-term predictions by combining GraphAutoEncoder (GAE) and Transformer models \citep{han2022predicting}, and directly predicting steady states through multi-layer readouts \citep{harsch2021direct}.

\subsection{Hierarchical GNN Models for Dynamical System Simulation}
The application of Graph Neural Networks (GNN) for dynamic system prediction is an emerging research area in scientific machine learning due to their versatility and effectiveness \citep{chang2016compositional, li2018learning, belbute2020combining, hu2023graph}. Unlike image-based learning methods such as Convolutional Neural Networks (CNNs) \citep{um2018liquid, ummenhofer2019lagrangian}, GNNs can directly handle unstructured simulation meshes, making them well-suited for simulating systems with complex domain boundaries while ensuring spatial invariance and locality \citep{wu2020comprehensive}. A notable milestone in this field is MeshGraphNets \citep{pfaff2020learning}, which enables the general scheme for learning mesh-based dynamical simulations. To mitigate the over-smoothing issue \citep{li2018deeper} that typically occurs in GNNs when applied to large or complex datasets with long-range interactions, several hierarchical models have been introduced recently. For instance, GMR-GMUS \citep{han2022predicting} utilizes a pooling method to select pivotal nodes through uniform sampling. Similarly, the EAGLE \citep{janny2023eagle} employs a clustering-based pooling method along with a transformer mechanism, showing promising performance in fluid dynamics. MS-MGN \citep{fortunato2022multiscale} proposes a dual-layer framework that passes messages at both fine and coarse resolutions for mesh-based simulation learning. BSMS-GNN \citep{cao2023efficient} analyzes limitations of existing pooling strategies and introduces a bi-stride pooling method using breadth-first search (BFS) to select nodes. \citep{yu2023learning} proposes a similar hierarchical structure as \citep{cao2023efficient} but with two different transformers to enable long-range interactions. 
%Subsequent work has extended this line in several directions: coupling GNNs with physics-informed neural networks (PINNs) \citep{gao2022physics}, achieving long-horizon forecasts by pairing Graph Autoencoders (GAEs) with Transformers \citep{han2022predicting}, and predicting steady states via multi-layer readout schemes \citep{harsch2021direct}. 
\subsection{Datasets for Unstructured Mesh-based Simulations}
{As a cornerstone in the field, MeshGraphNets \citep{pfaff2020learning} introduces a collection of datasets, showcasing the versatility of graph-based surrogates in various problems involving unstructured mesh simulations. Moreover, EAGLE \citep{janny2023eagle} presents a fluid dynamics dataset capturing unsteady and turbulent airflows; BSMS-GNN \citep{cao2023efficient} provides the InflatingFont dataset, which focuses on the quasi-static inflation of enclosed elastic surfaces. Despite recent progress, existing public datasets still leave two critical gaps for assessing hierarchical surrogates: (1) Many benchmarks contain thousands of vertices, yet their graph diameter—the maximum shortest-path length between any two mesh nodes—remains small, which under-exercises long-range message passing. (2) Most datasets expose only one mesh density or size band, preventing systematic studies of how a surrogate trained on small problems generalizes to larger ones. To bridge these gaps, we contribute \textit{DeformingBeam} together with its enlarged variant \textit{DeformingBeam-Large}. This pair constitutes the first public 3-D Lagrangian contact-deformation benchmark that also provides a scale-up case. The beam’s slender geometry yields graph diameters several times longer than existing solid mechanics datasets, exposing long-range interactions. Such a dataset enables rigorous testing of hierarchical surrogates and serves as a benchmark for probing model scalability and cross-scale generalization.}

\section{Conclusion}\label{conclusion}
{In this work we addressed two challenges in hierarchical mesh‑GNN surrogates—physics‑faithful graph coarsening and the accuracy–efficiency imbalance—by introducing M4GN, a three‑tier, segment‑centric framework built on two key innovations: a physics‑aware, hybrid segmentation strategy that yields contiguous, dynamically coherent segments, and a permutation‑invariant and computational efficient mesh segment aggregator. We systematically compared M4GN with leading baselines across multiple benchmarks using both traditional error metrics and proposed mesh‑quality metrics, and we quantified segmentation quality through three different intra- and inter-segment scores to illuminate how our hybrid segmentation shapes downstream accuracy. Additional studies assessed the accuracy‑versus‑speed trade‑off and robustness across graph diameter and scale. Across all tests, M4GN achieved the strongest overall balance of accuracy, efficiency, and mesh fidelity, exhibited the smallest error growth on wide‑diameter graphs, and maintained the smallest generalization gap among all hierarchical models. Despite these advances, M4GN still exhibits limitations that need further investigation. For example, the method currently imposes no hard constraints on contact boundaries and it offers no formal guarantees of physical consistency across segment interfaces. In addition, selecting segmentation hyper‑parameters still requires modest empirical tuning despite the provided guidelines. In summary, the findings in this paper underscore the value of coupling principled segmentation with balanced hierarchical reasoning for scalable, high‑fidelity mesh simulation. Future research will focus on incorporating constraint enforcement, automating hyper‑parameter selection, and extending the framework to coupled multi‑physics scenarios.}
%

%In this paper, we introduced M4GN, a novel approach that enhances dynamic system simulations through a hierarchical pipeline. Our extensive evaluations demonstrate that M4GN outperforms traditional models, offering significant improvements in accuracy and computational efficiency, particularly in scenarios involving long-range dynamics and larger physical domains. The adaptability of M4GN to large-scale graphs underscores its potential for real-world applications in complex physical systems. However, the method has limitations, including the absence of hard constraints on contact meshes, which can result in overlapping meshes, and it has no guarantees on physical consistency at segmentation interfaces. Also, although our approach is effective, the added segmentation hyperparameters require some experimentation to choose suitable values, introducing extra setup effort even with our quick‑start guidelines. These are important areas for future work to improve the robustness and applicability of the model. 

\subsubsection*{Broader Impact Statement}
This paper presents work whose goal is to advance the field of 
Machine Learning for Physics, Surrogate Modeling, and Dynamical System Simulation. There are many potential societal consequences of our work, none of which we feel must be specifically highlighted here.

\subsubsection*{Acknowledgments}
This work was performed under the auspices of the U.S. Department of Energy by the Lawrence Livermore National Laboratory under Contract DE-AC52-07NA27344. LLNL release number: LLNL-JRNL-2006212.

\bibliography{main}

\begin{thebibliography}{63}
\providecommand{\natexlab}[1]{#1}
\providecommand{\url}[1]{\texttt{#1}}
\expandafter\ifx\csname urlstyle\endcsname\relax
  \providecommand{\doi}[1]{doi: #1}\else
  \providecommand{\doi}{doi: \begingroup \urlstyle{rm}\Url}\fi

\bibitem[Achanta et~al.(2012)Achanta, Shaji, Smith, Lucchi, Fua, and S{\"u}sstrunk]{achanta2012slic}
Radhakrishna Achanta, Appu Shaji, Kevin Smith, Aurelien Lucchi, Pascal Fua, and Sabine S{\"u}sstrunk.
\newblock Slic superpixels compared to state-of-the-art superpixel methods.
\newblock \emph{IEEE transactions on pattern analysis and machine intelligence}, 34\penalty0 (11):\penalty0 2274--2282, 2012.

\bibitem[Alpert \& Yao(1995)Alpert and Yao]{alpert1995spectral}
Charles~J Alpert and So-Zen Yao.
\newblock Spectral partitioning: The more eigenvectors, the better.
\newblock In \emph{Proceedings of the 32nd annual ACM/IEEE design automation conference}, pp.\  195--200, 1995.

\bibitem[Alsentzer et~al.(2020)Alsentzer, Finlayson, Li, and Zitnik]{alsentzer2020subgraph}
Emily Alsentzer, Samuel Finlayson, Michelle Li, and Marinka Zitnik.
\newblock Subgraph neural networks.
\newblock \emph{Advances in Neural Information Processing Systems}, 33:\penalty0 8017--8029, 2020.

\bibitem[Andersen(2006)]{andersen2006linear}
Lars Andersen.
\newblock Linear elastodynamic analysis.
\newblock 2006.

\bibitem[Bathe(2001)]{bathe2001computational}
Klaus-J{\"u}rgen Bathe.
\newblock \emph{Computational fluid and solid mechanics}.
\newblock Elsevier, 2001.

\bibitem[Belbute-Peres et~al.(2020)Belbute-Peres, Economon, and Kolter]{belbute2020combining}
Filipe De~Avila Belbute-Peres, Thomas Economon, and Zico Kolter.
\newblock Combining differentiable pde solvers and graph neural networks for fluid flow prediction.
\newblock In \emph{international conference on machine learning}, pp.\  2402--2411. PMLR, 2020.

\bibitem[Bengio et~al.(1994)Bengio, Simard, and Frasconi]{bengio1994learning}
Yoshua Bengio, Patrice Simard, and Paolo Frasconi.
\newblock Learning long-term dependencies with gradient descent is difficult.
\newblock \emph{IEEE transactions on neural networks}, 5\penalty0 (2):\penalty0 157--166, 1994.

\bibitem[Bontempi \& Faravelli(1998)Bontempi and Faravelli]{bontempi1998lagrangian}
Franco Bontempi and Lucia Faravelli.
\newblock Lagrangian/eulerian description of dynamic system.
\newblock \emph{Journal of Engineering Mechanics}, 124\penalty0 (8):\penalty0 901--911, 1998.

\bibitem[Booij \& Holthuijsen(1987)Booij and Holthuijsen]{booij1987propagation}
Nico Booij and Leo~H Holthuijsen.
\newblock Propagation of ocean waves in discrete spectral wave models.
\newblock \emph{Journal of Computational Physics}, 68\penalty0 (2):\penalty0 307--326, 1987.

\bibitem[Cao et~al.(2023)Cao, Chai, Li, and Jiang]{cao2023efficient}
Yadi Cao, Menglei Chai, Minchen Li, and Chenfanfu Jiang.
\newblock Efficient learning of mesh-based physical simulation with bi-stride multi-scale graph neural network.
\newblock In \emph{International Conference on Machine Learning}, pp.\  3541--3558. PMLR, 2023.

\bibitem[Chang et~al.(2016)Chang, Ullman, Torralba, and Tenenbaum]{chang2016compositional}
Michael~B Chang, Tomer Ullman, Antonio Torralba, and Joshua~B Tenenbaum.
\newblock A compositional object-based approach to learning physical dynamics.
\newblock \emph{arXiv preprint arXiv:1612.00341}, 2016.

\bibitem[Chen et~al.(2020)Chen, Lin, Li, Li, Zhou, and Sun]{chen2020measuring}
Deli Chen, Yankai Lin, Wei Li, Peng Li, Jie Zhou, and Xu~Sun.
\newblock Measuring and relieving the over-smoothing problem for graph neural networks from the topological view.
\newblock In \emph{Proceedings of the AAAI conference on artificial intelligence}, volume~34, pp.\  3438--3445, 2020.

\bibitem[Cho et~al.(2014)Cho, Van~Merri{\"e}nboer, Gulcehre, Bahdanau, Bougares, Schwenk, and Bengio]{cho2014learning}
Kyunghyun Cho, Bart Van~Merri{\"e}nboer, Caglar Gulcehre, Dzmitry Bahdanau, Fethi Bougares, Holger Schwenk, and Yoshua Bengio.
\newblock Learning phrase representations using rnn encoder-decoder for statistical machine translation.
\newblock \emph{arXiv preprint arXiv:1406.1078}, 2014.

\bibitem[Chung et~al.(2014)Chung, Gulcehre, Cho, and Bengio]{chung2014empirical}
Junyoung Chung, Caglar Gulcehre, KyungHyun Cho, and Yoshua Bengio.
\newblock Empirical evaluation of gated recurrent neural networks on sequence modeling.
\newblock \emph{arXiv preprint arXiv:1412.3555}, 2014.

\bibitem[De~Witt et~al.(2012)De~Witt, Lessig, and Fiume]{de2012fluid}
Tyler De~Witt, Christian Lessig, and Eugene Fiume.
\newblock Fluid simulation using laplacian eigenfunctions.
\newblock \emph{ACM Transactions on Graphics (TOG)}, 31\penalty0 (1):\penalty0 1--11, 2012.

\bibitem[Delingette(1999)]{delingette1999general}
Herv{\'e} Delingette.
\newblock General object reconstruction based on simplex meshes.
\newblock \emph{International journal of computer vision}, 32:\penalty0 111--146, 1999.

\bibitem[Diao et~al.(2023)Diao, Yang, Zhang, Zhang, and Du]{diao2023solving}
Yu~Diao, Jianchuan Yang, Ying Zhang, Dawei Zhang, and Yiming Du.
\newblock Solving multi-material problems in solid mechanics using physics-informed neural networks based on domain decomposition technology.
\newblock \emph{Computer Methods in Applied Mechanics and Engineering}, 413:\penalty0 116120, 2023.

\bibitem[Dolean et~al.(2015)Dolean, Jolivet, and Nataf]{dolean2015introduction}
Victorita Dolean, Pierre Jolivet, and Fr{\'e}d{\'e}ric Nataf.
\newblock \emph{An introduction to domain decomposition methods: algorithms, theory, and parallel implementation}.
\newblock SIAM, 2015.

\bibitem[Dwivedi et~al.(2021)Dwivedi, Luu, Laurent, Bengio, and Bresson]{dwivedi2021graph}
Vijay~Prakash Dwivedi, Anh~Tuan Luu, Thomas Laurent, Yoshua Bengio, and Xavier Bresson.
\newblock Graph neural networks with learnable structural and positional representations.
\newblock \emph{arXiv preprint arXiv:2110.07875}, 2021.

\bibitem[Emanuel(1994)]{emanuel1994atmospheric}
Kerry~A Emanuel.
\newblock \emph{Atmospheric convection}.
\newblock Oxford University Press, USA, 1994.

\bibitem[Fahy(2007)]{fahy2007sound}
Frank~J Fahy.
\newblock \emph{Sound and structural vibration: radiation, transmission and response}.
\newblock Elsevier, 2007.

\bibitem[Fortunato et~al.(2022)Fortunato, Pfaff, Wirnsberger, Pritzel, and Battaglia]{fortunato2022multiscale}
Meire Fortunato, Tobias Pfaff, Peter Wirnsberger, Alexander Pritzel, and Peter Battaglia.
\newblock Multiscale meshgraphnets.
\newblock In \emph{ICML 2022 2nd AI for Science Workshop}, 2022.

\bibitem[Fu \& He(2001)Fu and He]{fu2001modal}
Zhi-Fang Fu and Jimin He.
\newblock \emph{Modal analysis}.
\newblock Elsevier, 2001.

\bibitem[Gao et~al.(2022)Gao, Zahr, and Wang]{gao2022physics}
Han Gao, Matthew~J Zahr, and Jian-Xun Wang.
\newblock Physics-informed graph neural galerkin networks: A unified framework for solving pde-governed forward and inverse problems.
\newblock \emph{Computer Methods in Applied Mechanics and Engineering}, 390:\penalty0 114502, 2022.

\bibitem[Gao \& Ji(2019)Gao and Ji]{gao2019graph}
Hongyang Gao and Shuiwang Ji.
\newblock Graph u-nets.
\newblock In \emph{international conference on machine learning}, pp.\  2083--2092. PMLR, 2019.

\bibitem[Godwin et~al.(2021)Godwin, Schaarschmidt, Gaunt, Sanchez-Gonzalez, Rubanova, Veli{\v{c}}kovi{\'c}, Kirkpatrick, and Battaglia]{godwin2021simple}
Jonathan Godwin, Michael Schaarschmidt, Alexander Gaunt, Alvaro Sanchez-Gonzalez, Yulia Rubanova, Petar Veli{\v{c}}kovi{\'c}, James Kirkpatrick, and Peter Battaglia.
\newblock Simple gnn regularisation for 3d molecular property prediction \& beyond.
\newblock \emph{arXiv preprint arXiv:2106.07971}, 2021.

\bibitem[Grebenkov \& Nguyen(2013)Grebenkov and Nguyen]{grebenkov2013geometrical}
Denis~S Grebenkov and B-T Nguyen.
\newblock Geometrical structure of laplacian eigenfunctions.
\newblock \emph{siam REVIEW}, 55\penalty0 (4):\penalty0 601--667, 2013.

\bibitem[Han et~al.(2022)Han, Gao, Pfaff, Wang, and Liu]{han2022predicting}
Xu~Han, Han Gao, Tobias Pfaff, Jian-Xun Wang, and Li-Ping Liu.
\newblock Predicting physics in mesh-reduced space with temporal attention.
\newblock \emph{arXiv preprint arXiv:2201.09113}, 2022.

\bibitem[Han et~al.(2012)Han, Zhang, et~al.]{han2012surrogate}
Zhong-Hua Han, Ke-Shi Zhang, et~al.
\newblock Surrogate-based optimization.
\newblock \emph{Real-world applications of genetic algorithms}, 343:\penalty0 343--362, 2012.

\bibitem[Hu et~al.(2023)Hu, Lei, and Castillo]{hu2023graph}
Yeping Hu, Bo~Lei, and Victor~M Castillo.
\newblock Graph learning in physical-informed mesh-reduced space for real-world dynamic systems.
\newblock In \emph{Proceedings of the 29th ACM SIGKDD Conference on Knowledge Discovery and Data Mining}, pp.\  4166--4174, 2023.

\bibitem[Huang et~al.(2009)Huang, Wicke, Adams, and Guibas]{huang2009shape}
Qi-Xing Huang, Martin Wicke, Bart Adams, and Leonidas Guibas.
\newblock Shape decomposition using modal analysis.
\newblock In \emph{Computer Graphics Forum}, volume~28, pp.\  407--416. Wiley Online Library, 2009.

\bibitem[Huttenlocher et~al.(1993)Huttenlocher, Klanderman, and Rucklidge]{huttenlocher1993comparing}
Daniel~P Huttenlocher, Gregory~A. Klanderman, and William~J Rucklidge.
\newblock Comparing images using the hausdorff distance.
\newblock \emph{IEEE Transactions on pattern analysis and machine intelligence}, 15\penalty0 (9):\penalty0 850--863, 1993.

\bibitem[Janny et~al.(2023)Janny, Beneteau, Nadri, Digne, Thome, and Wolf]{janny2023eagle}
Steeven Janny, Aur{\'e}lien Beneteau, Madiha Nadri, Julie Digne, Nicolas Thome, and Christian Wolf.
\newblock Eagle: Large-scale learning of turbulent fluid dynamics with mesh transformers.
\newblock \emph{arXiv preprint arXiv:2302.10803}, 2023.

\bibitem[Kamenski et~al.(2014)Kamenski, Huang, and Xu]{kamenski2014conditioning}
Lennard Kamenski, Weizhang Huang, and Hongguo Xu.
\newblock Conditioning of finite element equations with arbitrary anisotropic meshes.
\newblock \emph{Mathematics of computation}, 83\penalty0 (289):\penalty0 2187--2211, 2014.

\bibitem[Karypis \& Kumar(1998)Karypis and Kumar]{karypis1998fast}
George Karypis and Vipin Kumar.
\newblock A fast and high quality multilevel scheme for partitioning irregular graphs.
\newblock \emph{SIAM Journal on scientific Computing}, 20\penalty0 (1):\penalty0 359--392, 1998.

\bibitem[Kennett(2009)]{kennett2009seismic}
Brian Kennett.
\newblock \emph{Seismic wave propagation in stratified media}.
\newblock ANU Press, 2009.

\bibitem[Keriven(2022)]{keriven2022not}
Nicolas Keriven.
\newblock Not too little, not too much: a theoretical analysis of graph (over) smoothing.
\newblock \emph{Advances in Neural Information Processing Systems}, 35:\penalty0 2268--2281, 2022.

\bibitem[Kipf \& Welling(2016)Kipf and Welling]{kipf2016semi}
Thomas~N Kipf and Max Welling.
\newblock Semi-supervised classification with graph convolutional networks.
\newblock \emph{arXiv preprint arXiv:1609.02907}, 2016.

\bibitem[Knupp(2007)]{knupp2007remarks}
Patrick Knupp.
\newblock Remarks on mesh quality.
\newblock Technical report, Sandia National Lab.(SNL-NM), Albuquerque, NM (United States), 2007.

\bibitem[Lai et~al.(2009)Lai, Rubin, and Krempl]{lai2009introduction}
W~Michael Lai, David Rubin, and Erhard Krempl.
\newblock \emph{Introduction to continuum mechanics}.
\newblock Butterworth-Heinemann, 2009.

\bibitem[Li et~al.(2018{\natexlab{a}})Li, Han, and Wu]{li2018deeper}
Qimai Li, Zhichao Han, and Xiao-Ming Wu.
\newblock Deeper insights into graph convolutional networks for semi-supervised learning.
\newblock In \emph{Proceedings of the AAAI conference on artificial intelligence}, volume~32, 2018{\natexlab{a}}.

\bibitem[Li et~al.(2018{\natexlab{b}})Li, Wu, Tedrake, Tenenbaum, and Torralba]{li2018learning}
Yunzhu Li, Jiajun Wu, Russ Tedrake, Joshua~B Tenenbaum, and Antonio Torralba.
\newblock Learning particle dynamics for manipulating rigid bodies, deformable objects, and fluids.
\newblock \emph{arXiv preprint arXiv:1810.01566}, 2018{\natexlab{b}}.

\bibitem[Li et~al.(2020)Li, Kovachki, Azizzadenesheli, Liu, Stuart, Bhattacharya, and Anandkumar]{li2020multipole}
Zongyi Li, Nikola Kovachki, Kamyar Azizzadenesheli, Burigede Liu, Andrew Stuart, Kaushik Bhattacharya, and Anima Anandkumar.
\newblock Multipole graph neural operator for parametric partial differential equations.
\newblock \emph{Advances in Neural Information Processing Systems}, 33:\penalty0 6755--6766, 2020.

\bibitem[Lino et~al.(2022)Lino, Fotiadis, Bharath, and Cantwell]{lino2022towards}
Mario Lino, Stathi Fotiadis, Anil~A Bharath, and Chris Cantwell.
\newblock Towards fast simulation of environmental fluid mechanics with multi-scale graph neural networks.
\newblock \emph{arXiv preprint arXiv:2205.02637}, 2022.

\bibitem[Pfaff et~al.(2020)Pfaff, Fortunato, Sanchez-Gonzalez, and Battaglia]{pfaff2020learning}
Tobias Pfaff, Meire Fortunato, Alvaro Sanchez-Gonzalez, and Peter~W Battaglia.
\newblock Learning mesh-based simulation with graph networks.
\newblock \emph{arXiv preprint arXiv:2010.03409}, 2020.

\bibitem[Ramp{\'a}{\v{s}}ek et~al.(2022)Ramp{\'a}{\v{s}}ek, Galkin, Dwivedi, Luu, Wolf, and Beaini]{rampavsek2022recipe}
Ladislav Ramp{\'a}{\v{s}}ek, Michael Galkin, Vijay~Prakash Dwivedi, Anh~Tuan Luu, Guy Wolf, and Dominique Beaini.
\newblock Recipe for a general, powerful, scalable graph transformer.
\newblock \emph{Advances in Neural Information Processing Systems}, 35:\penalty0 14501--14515, 2022.

\bibitem[Ronneberger et~al.(2015)Ronneberger, Fischer, and Brox]{ronneberger2015u}
Olaf Ronneberger, Philipp Fischer, and Thomas Brox.
\newblock U-net: Convolutional networks for biomedical image segmentation.
\newblock In \emph{Medical image computing and computer-assisted intervention--MICCAI 2015: 18th international conference, Munich, Germany, October 5-9, 2015, proceedings, part III 18}, pp.\  234--241. Springer, 2015.

\bibitem[Schmid et~al.(2011)Schmid, Li, Juniper, and Pust]{schmid2011applications}
Peter~J Schmid, Larry Li, Matthew~P Juniper, and Oliver Pust.
\newblock Applications of the dynamic mode decomposition.
\newblock \emph{Theoretical and computational fluid dynamics}, 25:\penalty0 249--259, 2011.

\bibitem[Sun et~al.(2020)Sun, Gao, Pan, and Wang]{sun2020surrogate}
Luning Sun, Han Gao, Shaowu Pan, and Jian-Xun Wang.
\newblock Surrogate modeling for fluid flows based on physics-constrained deep learning without simulation data.
\newblock \emph{Computer Methods in Applied Mechanics and Engineering}, 361:\penalty0 112732, 2020.

\bibitem[Taira et~al.(2017)Taira, Brunton, Dawson, Rowley, Colonius, McKeon, Schmidt, Gordeyev, Theofilis, and Ukeiley]{taira2017modal}
Kunihiko Taira, Steven~L Brunton, Scott~TM Dawson, Clarence~W Rowley, Tim Colonius, Beverley~J McKeon, Oliver~T Schmidt, Stanislav Gordeyev, Vassilios Theofilis, and Lawrence~S Ukeiley.
\newblock Modal analysis of fluid flows: An overview.
\newblock \emph{Aiaa Journal}, 55\penalty0 (12):\penalty0 4013--4041, 2017.

\bibitem[Um et~al.(2018)Um, Hu, and Thuerey]{um2018liquid}
Kiwon Um, Xiangyu Hu, and Nils Thuerey.
\newblock Liquid splash modeling with neural networks.
\newblock In \emph{Computer Graphics Forum}, volume~37, pp.\  171--182. Wiley Online Library, 2018.

\bibitem[Ummenhofer et~al.(2019)Ummenhofer, Prantl, Thuerey, and Koltun]{ummenhofer2019lagrangian}
Benjamin Ummenhofer, Lukas Prantl, Nils Thuerey, and Vladlen Koltun.
\newblock Lagrangian fluid simulation with continuous convolutions.
\newblock In \emph{International Conference on Learning Representations}, 2019.

\bibitem[Veksler et~al.(2010)Veksler, Boykov, and Mehrani]{veksler2010superpixels}
Olga Veksler, Yuri Boykov, and Paria Mehrani.
\newblock Superpixels and supervoxels in an energy optimization framework.
\newblock In \emph{Computer Vision--ECCV 2010: 11th European Conference on Computer Vision, Heraklion, Crete, Greece, September 5-11, 2010, Proceedings, Part V 11}, pp.\  211--224. Springer, 2010.

\bibitem[Vinyals et~al.(2015)Vinyals, Bengio, and Kudlur]{vinyals2015order}
Oriol Vinyals, Samy Bengio, and Manjunath Kudlur.
\newblock Order matters: Sequence to sequence for sets.
\newblock \emph{arXiv preprint arXiv:1511.06391}, 2015.

\bibitem[Wang et~al.(2024)Wang, Cao, Huang, Liu, Hu, Luo, Song, Zhao, Liu, Sun, et~al.]{wang2024recent}
Haixin Wang, Yadi Cao, Zijie Huang, Yuxuan Liu, Peiyan Hu, Xiao Luo, Zezheng Song, Wanjia Zhao, Jilin Liu, Jinan Sun, et~al.
\newblock Recent advances on machine learning for computational fluid dynamics: A survey.
\newblock \emph{arXiv preprint arXiv:2408.12171}, 2024.

\bibitem[Wilson(2002)]{wilson2002three}
Edward~L Wilson.
\newblock Three-dimensional static and dynamic analysis of structures.
\newblock \emph{Computers and structures, Inc}, 1, 2002.

\bibitem[Wu et~al.(2021)Wu, Pan, Zhang, Wang, Liu, and Lin]{wu2021balanced}
Tong Wu, Liang Pan, Junzhe Zhang, Tai Wang, Ziwei Liu, and Dahua Lin.
\newblock Balanced chamfer distance as a comprehensive metric for point cloud completion.
\newblock \emph{Advances in Neural Information Processing Systems}, 34:\penalty0 29088--29100, 2021.

\bibitem[Wu et~al.(2020)Wu, Pan, Chen, Long, Zhang, and Philip]{wu2020comprehensive}
Zonghan Wu, Shirui Pan, Fengwen Chen, Guodong Long, Chengqi Zhang, and S~Yu Philip.
\newblock A comprehensive survey on graph neural networks.
\newblock \emph{IEEE transactions on neural networks and learning systems}, 32\penalty0 (1):\penalty0 4--24, 2020.

\bibitem[Xiong et~al.(2020)Xiong, Yang, He, Zheng, Zheng, Xing, Zhang, Lan, Wang, and Liu]{xiong2020layer}
Ruibin Xiong, Yunchang Yang, Di~He, Kai Zheng, Shuxin Zheng, Chen Xing, Huishuai Zhang, Yanyan Lan, Liwei Wang, and Tieyan Liu.
\newblock On layer normalization in the transformer architecture.
\newblock In \emph{International Conference on Machine Learning}, pp.\  10524--10533. PMLR, 2020.

\bibitem[Yang et~al.(2020)Yang, Wang, Yao, Liu, and Abdelzaher]{yang2020revisiting}
Chaoqi Yang, Ruijie Wang, Shuochao Yao, Shengzhong Liu, and Tarek Abdelzaher.
\newblock Revisiting over-smoothing in deep gcns.
\newblock \emph{arXiv preprint arXiv:2003.13663}, 2020.

\bibitem[Yang et~al.(2016)Yang, Li, Lan, Wang, Hao, and Qin]{yang2016coupling}
Chen Yang, Shuai Li, Yu~Lan, Lili Wang, Aimin Hao, and Hong Qin.
\newblock Coupling time-varying modal analysis and fem for real-time cutting simulation of objects with multi-material sub-domains.
\newblock \emph{Computer Aided Geometric Design}, 43:\penalty0 53--67, 2016.

\bibitem[Yu et~al.(2023)Yu, Choi, Cho, Lee, Kim, Chang, Woo, Kim, Lee, Yang, et~al.]{yu2023learning}
Youn-Yeol Yu, Jeongwhan Choi, Woojin Cho, Kookjin Lee, Nayong Kim, Kiseok Chang, ChangSeung Woo, Ilho Kim, SeokWoo Lee, Joon~Young Yang, et~al.
\newblock Learning flexible body collision dynamics with hierarchical contact mesh transformer.
\newblock \emph{arXiv preprint arXiv:2312.12467}, 2023.

\bibitem[Zienkiewicz \& Taylor(2005)Zienkiewicz and Taylor]{zienkiewicz2005finite}
Olek~C Zienkiewicz and Robert~L Taylor.
\newblock \emph{The finite element method set}.
\newblock Elsevier, 2005.

\end{thebibliography}
\bibliographystyle{tmlr}

\restoremaintoctoc             

\clearpage
\renewcommand{\contentsname}{Appendix: Table of Contents}
\tableofcontents               

\clearpage
\appendix

\section{Datasets} \label{appendix: dataset}

\subsection{Datasets Details}
\textbf{CylinderFlow} -- This public dataset includes simulations of transient incompressible flow around a cylinder, with varying diameters and locations, on a fixed 2D Eulerian mesh. In all fluid domains, the node type distinguishes fluid nodes, wall nodes, and inflow/outflow boundary nodes. The inlet boundary conditions are given by a prescribed parabolic profile, $u_{in} = u_0[1-4(y/H)]$ where $u_0$ and H are the centerline velocity and the distance between the sidewalls, respectively. The dataset contains 1000 training simulations, 100 validation simulations, and 100 test simulations.

{\textbf{EAGLE}\footnotemark[1] --  This public dataset contains simulations of complex airflow generated by a 2D unmanned aerial vehicle maneuvering in 2D scenes with varying floor profiles. While the scene geometry varies, the UAV trajectory is constant: the UAV starts in the center of the scene and navigates, hovering near the floor surface. The node types are fluid nodes, wall nodes, and aerial vehicle nodes. The dataset contains 948 training simulations, 118 validation simulations, and 118 test simulations.}

{\textbf{FlagSimple}\footnotemark[1] -- This public dataset includes simulations of a flag blowing in the wind and flag direction, with variation in wind speed. The mesh is static and remains the same for all simulations. Node types are flag nodes and handle nodes that are fixed. This dataset contains 1000 training simulations, 100 validation simulations, and 100 test simulations.}

\footnotetext[1]{EAGLE and FlagSimple are included only as supplementary benchmarks to show robustness across different dataset types and support additional analysis; we exclude ablation studies on these datasets to maintain a focused experimental scope and ensure clarity in result interpretation.}
\textbf{DeformingPlate} -- This public dataset includes simulations of hyperelastic plates deformed by a moving obstacle, with variations in plate design and obstacle design. The node types are plate nodes, handle nodes that are fixed, and obstacle nodes. This dataset contains 1000 training simulations, 100 validation simulations, and 100 test simulations.

\textbf{DeformingBeam} -- This dataset is generated using \textit{solids4foam} which is a toolbox for performing solid mechanics and fluid-solid interaction simulations in OpenFOAM and foam-extend. A nearly incompressible neo-Hookean model is used where the material properties are density $\rho_0$ = 1000 kg/$\text{m}^3$, Young’s modulus $E$ = 1 MPa, and Poisson’s ratio $\nu$ = 0.4. The beam comes in different geometries with various initial conditions and boundary conditions. The node types are plate nodes, handle nodes that are fixed, and obstacle nodes. This dataset contains 355 training simulations, 40 validation simulations, and 60 test simulations.
% \YH{Add more descriptions to this dataset description, such as node types, constant displacement for contact object, etc.}

\textbf{DeformingBeam (large)} -- A large domain DeformingBeam dataset is created for generalization studies. The physical domain size is doubled. The size of the mesh cell is kept consistent with the regular DeformingBeam dataset. This generalization dataset contains 112 simulations.

\begin{table}[h]
\caption{Detailed information for each dataset.}
\label{tab: dataset_basics}
\begin{sc}
\centering
\begin{scriptsize}
\centering
\resizebox{\linewidth}{!}{
\begin{tabular}{c|c|c|c|c|c|c|c|c}
\toprule
   System & Dataset  &  \begin{tabular}[c]{@{}c@{}} Avg.\\ \# Nodes\end{tabular} & \# Steps & Mesh Type & \begin{tabular}[c]{@{}c@{}} Graph\\ Diameter\end{tabular} & Node Feature & Edge Feature & Output\\
\midrule
\midrule
   \multirow{2}{*}{\begin{tabular}[c]{@{}l@{}} Fluid \end{tabular}}
   & CylinderFlow  & 1885 & 600 & Triangle, Eulerian, 2D & 11 & $\mathbf{v}_i, \mathbf{n}_i$ & $\mathbf{m}_{ij}, \vert\mathbf{m}_{ij}\vert$ & $\dot{\mathbf{v}}_i, p_i$ \\
   & EAGLE\footnotemark[1]   & 3390 & 990 & Triangle, Eulerian, 2D & 29.5 $\pm$ 1.7 & $\mathbf{v}_i, \mathbf{p}_i, \mathbf{n}_i$ & $\mathbf{m}_{ij}, \vert\mathbf{m}_{ij}\vert$ & $\dot{\mathbf{v}}_i, \dot{\mathbf{p}}_i$ \\
   \midrule
   \multirow{1}{*}{\begin{tabular}[c]{@{}l@{}} Flow-Structure \end{tabular}}
    & FlagSimple\footnotemark[1]  & 1579 & 400 & Triangle, Lagrangian, 3D & 41 & $\mathbf{x}_i, \dot{\mathbf{x}}_i, \mathbf{n}_i$ & $\mathbf{m}_{ij}, \vert\mathbf{m}_{ij}\vert$ & $\ddot{\mathbf{x}}_i$ \\
   \midrule
   \multirow{3}{*}{\begin{tabular}[c]{@{}l@{}} Solid \end{tabular}}  
   & DeformingPlate  & 1271 & 400 & Tetrahedron, Lagrangian, 3D & 16.9 $\pm$ 5.8 &  $\mathbf{x}_i, \dot{\mathbf{x}}_{\text{obs}}, \mathbf{n}_i$ & $\mathbf{x}_{ij}, \vert\mathbf{x}_{ij}\vert$, $\mathbf{m}_{ij}, \vert\mathbf{m}_{ij}\vert$ & $\dot{\mathbf{x}}_i$ \\
   & DeformingBeam & 1542 & 400 & Prism, Lagrangian, 3D & 41.3 $\pm$ 11.8 &  $\mathbf{x}_i, \dot{\mathbf{x}}_{\text{obs}}, \mathbf{n}_i$ & $\mathbf{x}_{ij}, \vert\mathbf{x}_{ij}\vert$, $\mathbf{m}_{ij}, \vert\mathbf{m}_{ij}\vert$ & $\dot{\mathbf{x}}_i$\\
   & DeformingBeam (large) & 4540 & 400 & Prism, Lagrangian, 3D & 82.1 $\pm$ 23.0 &  $\mathbf{x}_i, \dot{\mathbf{x}}_{\text{obs}}, \mathbf{n}_i$ & $\mathbf{x}_{ij}, \vert\mathbf{x}_{ij}\vert$, $\mathbf{m}_{ij}, \vert\mathbf{m}_{ij}\vert$ & $\dot{\mathbf{x}}_i$\\
\bottomrule
\end{tabular}}
\end{scriptsize}
\end{sc}
\end{table}

\section{Model Details}\label{appendix: model_baselines}
\subsection{M4GN Configurations}
The GNN part of M4GN adopts the encoder and graph processor in the MGN model \citep{pfaff2020learning}. The basic building block is Multi-Layer Perceptron (MLP). The MLP has 3 layers, a hidden dimension of 128, ReLU activation, and a single layer of Layer Normalization at the end. The node encoder and edge encoder(s) are 3-layer MLPs. By default, the M4GN has 7 message passing steps in the GNN. The mesh segment transformer consists of 4 self-attention layers, each with 8 heads. The output decoder is a 3-layer MLP without Layer Normalization. For DeformingPlate and DeformingBeam, M4GN only considers world edges between contacting mesh objects. The world edge radius is set to 0.01 for DeformingPlate and 0.002 for DeformingBeam. As both the DeformingBeam and DeformingPlate datasets feature a rigid object with quasi-static motion, we use only the first mode from our modal decomposition, which sufficiently captures the largest-scale deformation pattern. For the CylinderFlow dataset, we employ 6 modes, determined by an energy threshold criterion, ensuring a more comprehensive representation of the flow’s multi-scale dynamics. For Table~\ref{tab: overall_results}, every dataset uses SLIC‑MDODe segmentation, but the hyper‑parameters differ: For the CylinderFlow dataset, 36 segments are used with compactness value $\tau$ = 1.0, segment overlap, and positional encoding (PE) are both enabled. For DeformingPlate and DeformingBeam, 19 segments are used with the former using $\tau$ = 1.0 with no overlap or PE, while the latter uses $\tau$ = 0.5 with overlap but without PE. Ablation studies are also conducted based on these configurations.

%For solid mechanics datasets, world edges are created and the information can be found in Table~\ref{tab: world_edge}. MMSGN only creates contact world edges between the deformed object and the obstacle driver with a small $r_W$. The number of world edges is significantly lower than MGN, which creates world edges between all mesh nodes using larger $r_W$.

% \begin{table}[h]
% \caption{Hyperparameters for default M4GN configurations}
% \label{tab:hyperparams_M4GN}
% \centering
% \begin{footnotesize}
% \begin{sc}
% \centering
% \begin{tabular}{lcccccc}
% \toprule
% Dataset   & Seg Method & $N_{\text{SEG}}$ & $\tau$ & Seg Overlap & PE \\
% \midrule
% Cylinder  & {SLIC-MDOD$_{e}$}       & 36             & 1.0      & \cmark  & \cmark \\
% Plate     & {SLIC-MDOD$_{e}$}       & 19             & 1.0      & \xmark  & \xmark \\
% Beam      & {SLIC-MDOD$_{e}$}       & 19             & 0.5      & \cmark  & \xmark \\
% \bottomrule
% \end{tabular}
% \end{sc}
% \end{footnotesize}
% \end{table}

\subsection{Baselines}
\textbf{GCN} -- The GCN model consists of 15 GCN layers with a hidden dimension of 128. The GCN model does not have edge input. Node input includes mesh position $\mathbf{x}_i$ for CylinderFlow. The implementation is from PyTorch Geometric.

\textbf{g-U-Net} -- The g-U-Net model is a modified version from PyTorch Geometric. Instead of GCN layers, it is built using the GNN layers similar to MGN. The level of scale is 7 for CylinderFlow, 6 for DeformingPlate, and 4 for DeformingBeam. 

\textbf{MGN} -- Our implementation of MGN follows the one described in~\cite{pfaff2020learning}. The processor of MGN contains 15 MP steps. World edges are constructed as specified in the paper, with a world edge radius of 0.03 for DeformingPlate and 0.003 for DeformingBeam.

\textbf{BSMS-GNN} -- We followed the BSMS-GNN implementation \cite{cao2023efficient} from \url{https://github.com/Eydcao/BSMS-GNN}. We introduced a modification to the original code by incorporating output normalization, which we observed to enhance the model's performance. For CylinderFlow and DeformingPlate, we used the same number of multi-scale levels as specified in the BSMS-GNN paper, at 7 and 6 levels, respectively. The number of multi-scale levels for DeformingBeam is set at 4 as an optimal configuration. 
% \YH{(If this scale follows the paper's setting, need to mention it.)}

\textbf{EAGLE} -- The implementation of EAGLE follows the paper \cite{janny2023eagle} and the code repository \url{https://github.com/eagle-dataset/EagleMeshTransformer}. We set the number of nodes per cluster at 20, which offers a balanced performance and efficiency according to the paper. This results in 94 clusters for CylinderFlow, 64 for DeformingPlate, and 38 for DeformingBeam. In addition, we add contacting world edges in the EAGLE implementation for DeformingPlate and DeformingBeam to improve performance. The world edges are added the same as in M4GN.

\subsection{Training Details}\label{appendix: training}
During training, random Gaussian noise is added to the spatial node inputs, as described in~\cite{pfaff2020learning}. For CylinderFlow, all models use a noise scale of 0.02. For DeformingPlate, all models use a noise scale of 0.003. For DeformingBeam, EAGLE and M4GN use a noise scale of 1e-4 and other models use a noise scale of 1e-3. 

For GCN, g-U-Net, MGN, EAGLE and M4GN, we adopt the same training scheme: For CylinderFlow and DeformingPlate, we trained the model for 2M steps. The learning rate starts at 1e-4 and exponentially decays to 1e-6 from 1M to 2M steps. For DeformingBeam, we trained the model for 1M steps. The learning rate starts at 1e-4 and exponentially decays to 1e-6 from 500K to 1M steps.

For BSMS-GNN, we adopt the training scheme from the original implementation. Models for CylinderFlow and DeformingPlate were trained for 50 epochs, corresponding to 3.75M and 3M training steps, respectively. DeformingBeam model was trained for 100 epochs, corresponding to 1.775M training steps.

Across all models and datasets, we use a batch size of 8. Experiments were conducted using PyTorch distributed training over two Nvidia Tesla P100 GPUs.

%\subsection{Modal Decomposition Details}

\subsection{Hybrid Mesh Segmentation Details} \label{appendix: mesh_segmentation}

\begin{figure}[htbp]
\includegraphics[width=\textwidth]{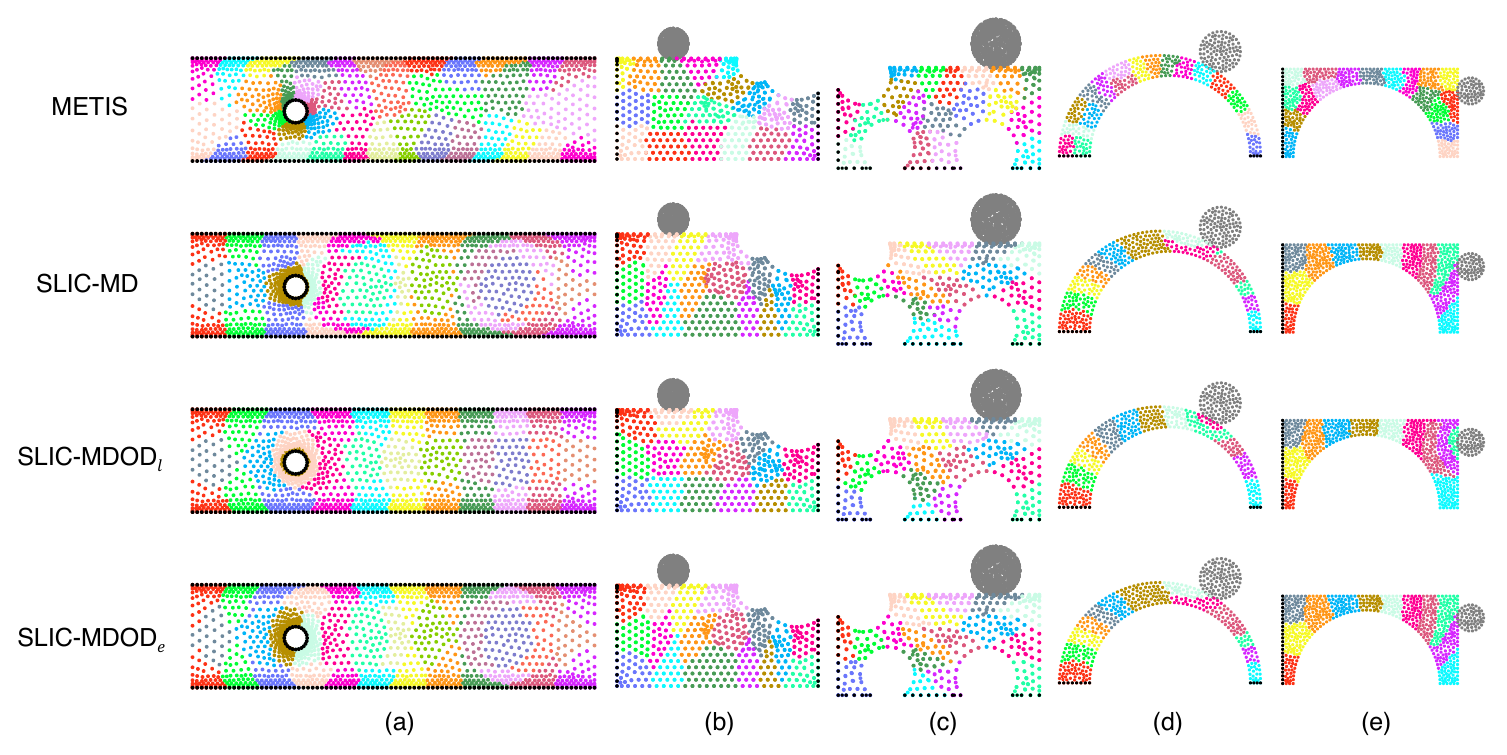}
\caption{Illustration of different segmentation methods under various cases: (a):CylinderFlow; (b)(c): DeformingPlate; (d)(e): DeformingBeam. Mesh nodes are colored based on segment id and all boundary nodes are colored in black.}
\label{fig: patch_compare}
\end{figure}

In Figure~\ref{fig: patch_compare}, several cases are selected from each dataset to illustrate the difference of each mesh graph segmentation method. It's worth noting that the graph will be partitioned only once during the training and testing phase for each simulation, and this partitioning will remain consistent across all time steps. This is because the segmentation is based solely on the system's properties and initial conditions prior to the start of the simulation. 

% \YH{(Need to mention that we assume non-deformable obstacle in this paper and assume known dynamics of the obstacle. Therefore, we treat the entire obstacle as a mesh segment.)}

%\YH{We can also creating overlapping segments \citep{xx}. Overlapping regions allow for smoother transitions and reduce the discontinuity at the boundaries.This exchange ensures that the displacement and stress fields are continuous across the boundary between A and B. Ablations results on the effect of adding overlapping segments can be found in \ref{xx}}

\begin{small}
\begin{algorithm}
\caption{Modal Decomposition}
\label{alg: modal_decomp}
\begin{algorithmic}[1]
\STATE \text{Case Type:} \texttt{solid} or \texttt{fluid}
\STATE \textbf{Input:} Finite element mesh, boundary conditions, material properties for solid (e.g. $E,\nu,\rho$), number of modes $m$ \\
\STATE \textbf{Build Finite Element Basis:}
\STATE \quad Define shape functions on each element using the node connectivity
\STATE \quad Enumerate degrees of freedom (DOFs) for each node/component
%\item Map boundary nodes to fixed DOFs or Dirichlet DOFs
\IF{\text{Case Type} = \texttt{solid}}
    \STATE \textbf{Structural Modal Analysis}
    \STATE Assemble stiffness matrix $\mathbf{K}$ (using elasticity)
    \STATE Assemble mass matrix $\mathbf{M}$ (using density)
    \STATE Apply boundary conditions to eliminate fixed DOFs
    \STATE Solve \(\mathbf{K} \boldsymbol{\phi} = \lambda\,\mathbf{M} \boldsymbol{\phi}\) for the first $m$ modes
    \STATE \textbf{Output:} Eigenpairs \(\{(\lambda_i,\,\bm{\phi}_i)\}_{i=1}^m\) (\emph{structural modes})
\ELSIF{\text{Case Type} = \texttt{fluid}}
    \STATE \textbf{Laplacian Eigenfunctions}
    \STATE Assemble Laplacian matrix
    \STATE Assemble $L^2$-type matrix
    \STATE Apply Dirichlet constraints on boundary nodes %$\phi\big|_{\partial\Omega} = 0$
    \STATE Solve $-\,\nabla^2 \phi = \lambda\,\phi$ for the first $m$ modes 
    \STATE \textbf{Output:} Eigenpairs \(\{(\lambda_i,\, {\phi}_i)\}_{i=1}^m\) (\emph{harmonic modes})
\ENDIF
\STATE \textbf{Return:} $m$-dimensional feature vector $f^{md}_i$ = $(\phi_1(i), \phi_1(i), \dots \phi_m(i))$ at each mesh node $i$

%Eigenpairs $(\lambda_i, \text{mode}_i)$ for either solid or fluid case
\end{algorithmic}    
\end{algorithm}
\end{small}

\begin{small}
\begin{algorithm}
\caption{Hybrid Mesh Segmentation}
\label{alg: slic}
\begin{algorithmic}[1]
\STATE \textbf{Input:} 
\STATE \quad Initial mesh graph $G = (\mathcal{V}, \mathcal{E})$
\STATE \quad Perform modal decomposition and compute mesh node feature $f_i$ 
\STATE \quad Number of segments $K$, compactness parameter $\tau$, average cluster size $S$
\STATE \textbf{Output:} Mesh node segmentation $\{\mathcal{V}_{S_k}\}_{k=1}^K$

\\\hrulefill

\noindent\textcolor{black}{
{\textit{$\Rightarrow$ \textbf{Graph-based Mesh Segment Initialization}}}
\STATE {\textbf{Coarsening Phase:}}
\STATE {$G_{\text{coarse}} \leftarrow G$}
\WHILE{size of $G_{\text{coarse}}$ is larger than threshold}
    \STATE Combine pairs of connected nodes in $G_{\text{coarse}}$ to form a coarser graph
    \STATE $G_{\text{coarse}} \leftarrow$ coarsened graph
\ENDWHILE
\STATE \textbf{Initial Partitioning:}
\STATE Partition $G_{\text{coarse}}$ into $K$ segments using a standard partitioning method (e.g., spectral partitioning)
\STATE \textbf{Uncoarsening and Refinement Phase:}
\WHILE{$G_{\text{coarse}} \neq G$}
    \STATE Expand $G_{\text{coarse}}$ to the next finer graph $G_{\text{fine}}$
    \STATE Project partitions onto $G_{\text{fine}}$
    \STATE Refine the partitioning on $G_{\text{fine}}$ by iteratively
           moving a vertex to neighboring segments \textbf{iff} the move lowers the total number (or weight) of edges that cross between segments, provided each segment remains roughly balanced in size.
    \STATE $G_{\text{coarse}} \leftarrow G_{\text{fine}}$
\ENDWHILE
\STATE Obtain initial clusters $\{\mathcal{V}_{S_k}\}_{k=1}^K$ from the final partitioning, which will be updated next
}
\\\hrulefill

\noindent{\textit{$\Rightarrow$ \textbf{Superpixel-based Mesh Segment Refinement}}}
% End METIS initialization
\REPEAT

    \FOR{each mesh segment centroid $C_k$}
        \STATE Update $C_k$ by averaging over all mesh nodes assigned to it:
        \[
        C_k = [x_{C_k}, f_{C_k}]^T = \frac{1}{|\mathcal{V}_{S_k}|}\sum_{i \in \mathcal{V}_{S_k}}[x_i, f_i]^T
        \]
        where $\mathcal{V}_{S_k}$ is the set of mesh nodes assigned to segment ${S_k}$
    \ENDFOR
    
    \FOR{each mesh node $i \in V$} 
        \STATE Compute the distance measure $d(i, C_k)$ to each cluster center $C_k$ using:
        \[
        d(i, C_k) = \|f_i - f_{C_k}\| + \tau \|x_i - x_{C_k}\|
        \]
        where $x_i$ and $x_{C_k}$ are the spatial coordinates, $f_i$ and $f_{C_k}$ are the physics-guided features.
        \STATE Assign mesh node $i$ to the nearest segment centroid $C_k$ if $d(i, C_k) \leq S$
    \ENDFOR
\UNTIL{convergence or a maximum number of iterations is reached}

% \STATE Refine boundaries by enforcing connectivity using graph traversal algorithms such as BFS or DFS
\end{algorithmic}
\end{algorithm}
\end{small}

The pseudo code of the hybrid segmentation module proposed in this work can be found in Algorithm~\ref{alg: slic}. Here, METIS \citep{karypis1998fast} is a graph partitioning technique that efficiently divides meshes into approximately equal-sized partitions. It leverages multilevel partitioning algorithms to minimize the edge-cut or communication costs between the resulting partitions. We employ METIS due to its versatility in creating a user-specified number of equal-sized mesh segments. SLIC \citep{achanta2012slic} is a clustering algorithm employed for partitioning data. In our approach, we adapt SLIC to segment the mesh based on physics-informed features. These features could guide SLIC to create a segmentation that captures the underlying physics of the system. The consequent mesh segments can potentially enable efficient macro-level information exchange tailored to the system's dynamics.
% \textcolor{red}{Need to add descriptions of SLIC variations}
Concretely, for each node $i$, we incorporate physics-aware feature $f_i^{md}$ derived from modal decomposition. Additionally, we augment these features by concatenating a measure of the shortest distance to obstacle nodes $d_i^{obs}$. To ensure that this measure dominates when $d_i^{{obs}}$ is small, we apply either an exponential or logarithmic transformation, defined as:
\begin{equation}
    f_{\exp}(d) = \exp(-d), \quad f_{\log}(d) = \log(d).
\end{equation}
Depending on the selection of features and the transformation function, we design 6 variants of SLIC:
\begin{itemize}
    \item SLIC-OD: $f_i = d_i^{{obs}}$
    \item SLIC-OD$_l$: $f_i = f_{\log}(d_i^{{obs}})$
    \item SLIC-OD$_e$: $f_i = f_{\exp}(d_i^{{obs}})$
    \item SLIC-MD: $f_i = f_i^{md}$
    \item SLIC-MDOD$_l$: $f_i = \left[f_{\log}(d_i^{{obs}}), f_i^{md}\right]^T$
    \item SLIC-MDOD$_e$: $f_i = \left[f_{\exp}(d_i^{{obs}}), f_i^{md}\right]^T$

\end{itemize}
After we have the physics-aware feature, we can apply the SLIC algorithm to get the mesh node segments. 

\subsection{Mesh Segment Hyperparameter Selection}
The compactness parameter $\tau$ in the SLIC algorithm controls the trade-off between physics-guided feature similarity and spatial proximity. Our goal is to choose $\tau$ such that the resulting segmentation captures both underlying physical patterns and spatial coherence (i.e., grouping nodes that are close to each other). For CylinderFlow and DeformingPlate, we set $\tau=1.0$, which provides a balanced segmentation. For DeformingBeam, we set a lower $\tau$ at 0.5 to promote a better alignment with physical features. The cluster size $S$ is determined such that the domain area satisfies: $\text{Domain Area} = KS^2$, where $K$ is the number of segments and the domain area is given by $(x_{\max} - x_{\min} )(y_{\max} - y_{\min})$. The average cluster size $S$ for CylinderFlow, DeformingPlate and DeformingBeam is set to be $\sqrt{0.656/K}$, $\sqrt{0.125/K}$ and $\sqrt{0.005/K}$, respectively.

Systematically choosing the optimal number of segments $K$ requires both domain insight and practical experimentation. In our experience, two main factors drive the choice of $K$: (1) the total mesh size ($N$) and (2) local variations in mesh density. For instance, CylinderFlow is particularly dense near boundaries, which benefits from a larger $K$, whereas DeformingBeam/DeformingPlate have more uniformly distributed nodes, so a smaller $K$ can suffice.

To make this selection concrete, we typically perform a short hyperparameter sweep over a small set of candidate values for $K$. A simple heuristic is to pick $K$ values on a roughly geometric or linear scale, for instance: $K \in \{\sqrt{N}/2, \sqrt{N}, 2\sqrt{N},....\}$ up to a point where adding more segments no longer improves validation metrics (e.g., prediction accuracy, mesh quality). In practice, testing each candidate $K$ on a subset (e.g., 10\%) of the training data is typically enough to identify a near-optimal configuration, and then we finalize training with that $K$ on the full dataset. This strategy is computationally manageable and provides a principled way to tailor $K$ to new domains.

% \subsection{Potential}
% \YH{Add Pseudo Code}

% \begin{figure}[htbp]
%     \centering
%     \includegraphics[width=0.6\textwidth]{image/deformingBeam_d_vs_rmse.pdf}
%     \vskip -0.1in
%     \caption{Relationship between graph diameter and prediction accuracy for different models under DeformingBeam dataset.}
%     \label{fig:beam_radius_vs_accuracy}
% \end{figure}

\section{{Additional Evaluation Metrics and Results}} \label{appendix: additional_results}
\subsection{{Metrics for Mesh Quality Measure}} \label{appendix: additional_mesh_quality_metrics}

{\textbf{Hausdorff Distance}} -- The Hausdorff Distance measures how well the mesh with the predicted node positions conforms to the system's true geometry. It is defined as:
\begin{equation}
    \text{GF}_h(\mathcal{V}, \hat{\mathcal{V}}) = \max\Bigl\{h(\mathcal{V}, \hat{\mathcal{V}}), h(\hat{\mathcal{V}}, \mathcal{V})\Bigr\}, \label{eq:GF}
\end{equation}
where $h(\mathcal{V}, \hat{\mathcal{V}}) = \sup_{\mathbf{x} \in \mathcal{V}} \inf_{\hat{\mathbf{x}} \in \hat{\mathcal{V}}} \| \mathbf{x} - \hat{\mathbf{x}} \|$ is the directed Hausdorff distance \citep{huttenlocher1993comparing} from the ground-truth node set $\mathcal{V}$ to the predicted node set $\hat{\mathcal{V}}$.

{\textbf{Chamfer Distance}} -- The Chamfer Distance \citep{wu2021balanced} measures the average distance between points on the predicted mesh and the true mesh, providing a balanced assessment of $\text{GF}_h$. Unlike the Hausdorff Distance, which focuses on the maximum deviation, the Chamfer Distance is sensitive to the overall distribution of errors across the mesh surfaces. As both Chamfer and Hausdorff distance are measures for GF, we name them as GF$_c$ and GF$_h$ for simplicity, respectively. The Chamfer distance is mathematically defined as: 
\begin{equation}
    \text{GF}_c(\mathcal{V}, \hat{\mathcal{V}}) = \frac{1}{|\mathcal{V}|}\sum_{\mathbf{x} \in \mathcal{V}}\text{min}_{\hat{\mathbf{x}} \in \hat{\mathcal{V}}} \| \mathbf{x} - \hat{\mathbf{x}} \|^2 + \frac{1}{|\hat{\mathcal{V}}|}\sum_{\hat{\mathbf{x}} \in \hat{\mathcal{V}}}\text{min}_{{\mathbf{x}} \in {\mathcal{V}}} \| \hat{\mathbf{x}} - {\mathbf{x}} \|^2,
\end{equation}
{where $\mathcal{V}$ and $\hat{\mathcal{V}}$ are the set of vertices in the ground-truth and predict mesh, respectively. $|\mathcal{V}|$ and $|\hat{\mathcal{V}}|$ denote the number of vertices in each mesh. }

\textbf{Mesh Continuity} -- Mesh Continuity \citep{knupp2007remarks} evaluates the uniformity of predicted mesh cell sizes to ensure stability and is defined as 
\begin{equation}
    \text{MC} = \frac{1}{C}\sum_{i=1}^C \frac{\max_{c_j\in \text{Adj}(c_i)} V(c_j)}{\min_{c_j\in \text{Adj}(c_i)} V(c_j)}, \label{eq:MC}
\end{equation}
where $\text{Adj}(c_i)$ is the neighboring cells of cell $c_i$, and $V(c_i)$ calculates the volumetric area for $c_i$. 

{\textbf{Aspect Ratio (error)} -- The Aspect Ratio \citep{zienkiewicz2005finite} assesses the shape quality of individual 2D or 3D mesh elements and is widely used in finite element method (FEM) literature to evaluate how closely each element approaches the ideal shape, such as an equilateral triangle or a regular tetrahedron. For example, for triangular meshes, the aspect ratio is defined as $\frac{L_{\text{max}}}{2\sqrt{\sqrt{3}A}}$, where $L_\text{max}$ is the longest edge length, $A$ is the area of the triangle. For tetrahedra mesh, it is defined as $\frac{\sqrt{6}L_{\text{max}}}{V^{1/3}}$, where $V$ the volume of the tetrahedron.  High aspect ratios indicate elongated or distorted elements, which can cause numerical instability and reduce simulation accuracy. By analyzing the aspect ratios across all elements, we can assess the overall uniformity and regularity of the mesh. To evaluate the accuracy of the predicted mesh compared to the ground truth, we calculate the aspect ratio for both the predicted and actual meshes. The Aspect Ratio Error is then determined as the $L_1$ distance between these two values. This error metric quantifies the deviation in shape quality between the predicted and true meshes, providing a direct measure of how well the prediction preserves the ideal element shapes. Incorporating the Aspect Ratio Error allows for a more precise evaluation of mesh quality and prediction accuracy, ensuring that the segmented meshes maintain the necessary geometric properties for reliable simulations.}

% Below are equations for calculating aspect ratio for different meshes involved in our datasets:
% \begin{equation}
%     \text{Triangular} = , \quad \text{Tetrahedral} = \frac{\sqrt{6}L_{\text{max}}}{V^{1/3}}, \quad 
% \end{equation}

\subsection{{Segmentation Quality Metrics}} \label{appendix: segment_quality_metrics}

{In order to rigorously evaluate the quality of our hybrid mesh segmentation and its impact on the prediction of system dynamics, it is essential to consider metrics that assess both inter-segment and intra-segment characteristics. We introduce three such metrics —\textit{Conductance}, \textit{Edge Cut Ratio}, and \textit{Silhouette Score} — which provide a comprehensive assessment of segmentation quality by quantifying the cohesion within segments and the separation between segments. The necessity of these metrics arises from the need to ensure that segments are well-separated, minimizing unnecessary interactions between dissimilar regions (inter-segment quality), and that nodes within the same segment share similar properties or behaviors (intra-segment quality).}

{Moreover, in our hierarchical model architecture, the intra-segment quality pertains to the micro-level information exchange stage. High intra-segment quality facilitates accurate modeling of local dynamics within each segment by ensuring that nodes are cohesive and share similar dynamic behaviors. Conversely, the inter-segment quality directly relates to the macro-level information exchange stage. High inter-segment quality ensures efficient communication between segments by reducing redundant or irrelevant interactions, which is crucial for capturing global dynamics across the entire mesh. Below are the details of three metrics to measure segmentation quality. }

{\textbf{Conductance} -- Conductance measures the fraction of total edge connections that cross between different segments relative to the total connections of the segments. It assesses how well the segmentation minimizes inter-segment connections while maintaining intra-segment cohesion. Let $G = (\mathcal{V}, \mathcal{E})$ as an undirected graph representing the mesh, where $\mathcal{V}$ is the set of nodes and $\mathcal{E}$ is the set of edges. Let $S$ be a segment and $\Bar{S} = G \arraybackslash$ \textbackslash $S$ be its complement. The conductance of segment $S$ is defined as:}
\begin{equation}
    \text{Conductance} = \frac{ \left| \{ (u, v) \in \mathcal{E} \mid u \in S,\ v \in \bar{S} \} \right| }{ \min\left( \text{vol}(S),\ \text{vol}(\bar{S}) \right) },
\end{equation}
{where the numerator is the number of edges crossing between $S$ and $\Bar{S}$. The volumn of segment $S$ is given by $\text{vol}(S) = \sum_{u \in S} \deg(u)$, where $\deg{u}$ is the degree of node $u$ (the number of edges connected to $u$). }

{\textbf{Edge Cut Ratio} -- The Edge Cut Ratio quantifies the proportion of edges that are cut by the segmentation relative to the total number of edges in the mesh. It is defined as:}
\begin{equation}
\text{Edge Cut Ratio} = \frac{ \left| \{ (u, v) \in \mathcal{E} \mid \text{Seg}(u) \neq \text{Seg}(v) \} \right| }{E},
\end{equation}
{where the denominator is the number of edges that connect nodes in different segment. $\text{Seg}(u)$ denotes the segment to which node $u$ belongs and $E = |\mathcal{E}|$ is the total number of edges. }

{\textbf{Silhouette Score} -- For each node $i$, the Silhouette Score evaluates how similar $i$ is to nodes in its own segment compared to nodes in other segments. It is defined as:}
\begin{equation}
    \text{Silhouette Score} = \frac{1}{N} \sum_{u=1}^{N}\frac{ b(i) - a(i) }{ \max\{ a(i),\ b(i) \} },
\end{equation}
{where $N$ is the total number of nodes, $a(i)$ is the average dissimilarity of node $i$ with all other nodes in the same segment and $b(i)$ is the lowest average dissimilarity of node  $i$ to any other segment to which $i$ does not belong. To be more specific $a(i) = \frac{1}{|S_i| - 1} \sum_{\substack{j \in S_i \\ j \neq i}} d(i, j)$, $b(i) = \text{min}_{S' \neq S_i} \left( \frac{1}{|S'|} \sum_{j \in S'} d(i, j) \right)$, where $S_i$ is the segment containing node $i$ and $d(i, j)$ can be any appropriate distance metric, such as Euclidean distance based on node features or positions.}

{By combining these metrics, we achieve a comprehensive evaluation of segmentation quality that covers both the internal cohesion of segments and their external separation. Having these metrics, along with prediction result metrics, can better help us understand the effect of segmentation on the predicted system dynamics. These metrics can be used to help finding better physics-aware segment features and determining the optimal segmentation number (results and discussion in Appendix~\ref{appendix: ablation_macro_2}).}

\subsection{{Evaluation on Supplementary Datasets}}\label{appendix: supplementary_dataset_results}
{To evaluate the robustness of our approach across distinct physical regimes, we retrained all models on two additional datasets—\textit{EAGLE} and \textit{FlagSimple}—and compared the proposed M4GN architecture with the strongest baselines identified earlier, MGN and the task‑specific EAGLE solver. As summarized in Table~\ref{tab: additional_dataset_result}, M4GN attains the lowest prediction error on both datasets, lowering RMSE by 4\% on EAGLE and by 40\% on FlagSimple comparing to the second best model. For FlagSimple, M4GN also yields the most faithful meshes, reducing geometric‐quality defect metrics by 25–57\% relative to both baselines. Moreover, it halves peak training memory consumption and accelerates inference by up to 19\%. Collectively, these results demonstrate that M4GN sustains its performance advantages on the supplementary datasets, delivering higher predictive accuracy, better mesh quality, and more efficient computation.}

\begin{table*}[t]
\caption{{Comparison of results with state-of-the-art methods across two additional datasets, where each model is trained independently for each dataset. Prediction accuracy is evaluated using Root Mean Square Error
(RMSE), with the output being the 2D velocity and pressure field for EAGLE and the 3D position for FlagSimple. Results are averaged over three experiments with different random seeds and presented as mean and standard deviation.}}
\label{tab: additional_dataset_result}
\begin{adjustbox}{width=\textwidth}
%\vskip 0.15in
\begin{scriptsize}
\begin{sc}
{
\begin{tabular}{llccccccc} 
\toprule
  &  & \multicolumn{4}{c}{\textnormal{Mesh Quality Metrics $\downarrow$} }  &\multicolumn{3}{c}{\textnormal{Prediction and Computation Metrics $\downarrow$}}   \\
 \cmidrule(lr){3-6} \cmidrule(lr){7-9} 
Dataset  & Model & \begin{tabular}[c]{@{}c@{}} $\text{GF}_h$ \\ ($\times 10^{-2}$) \end{tabular} & \begin{tabular}[c]{@{}c@{}} $\text{GF}_c$ \\ ($\times 10^{-5}$) \end{tabular}& \begin{tabular}[c]{@{}c@{}} MC \\ ($\times 10^{-2}$) \end{tabular} & \begin{tabular}[c]{@{}c@{}} AR \\ ($\times 10^{-2}$) \end{tabular} & \begin{tabular}[c]{@{}c@{}} RMSE-\textnormal{all} \end{tabular} &  \begin{tabular}[c]{@{}c@{}} Train \\ Memory [MB] \end{tabular}  &  \begin{tabular}[c]{@{}c@{}} Test Time\\ per step [ms] \end{tabular} \\
\midrule
\midrule
\multirow{3}{*}{\begin{tabular}[c]{@{}l@{}} EAGLE \end{tabular}} 
& MGN     & -  & - & - & - & 4.13 \tiny$\pm$ 0.05 & 10525 & 35.8 \\
& EAGLE       & -  & -   & - & - &  4.24 \tiny$\pm$ 0.06 & 7254 & 35.2 \\
& M4GN (Ours) & -  & -  & - & -&  \textbf{3.95 \tiny$\pm$ 0.05} & \textbf{5308} & \textbf{28.4}\\
\midrule
\multirow{3}{*}{\begin{tabular}[c]{@{}l@{}} Flag\\Simple\\ \end{tabular}}
& MGN    & 1.82 \tiny$\pm$ 0.08 & {5.01 \tiny$\pm$ 0.34} & 6.02 \tiny$\pm$ 0.59 & {4.16 \tiny$\pm$ 0.65} &  0.25\tiny$\pm$ 0.11 & 1060 & 36.5\\
& EAGLE       & 1.73 \tiny$\pm$ 0.08 & {5.22 \tiny$\pm$ 0.50} & 6.71 \tiny$\pm$ 0.55 & {5.49 \tiny$\pm$ 1.05} &   1.01 \tiny$\pm$ 1.14 & 1336 & 41.9 \\
& M4GN (Ours)         & \textbf{0.98 \tiny$\pm$ 0.05} & {\textbf{2.23 \tiny$\pm$ 0.19}} & \textbf{4.06 \tiny$\pm$ 0.40} & {\textbf{3.11 \tiny$\pm$ 0.62 }} & \textbf{0.15 \tiny$\pm$ 0.01} & \textbf{549} & \textbf{30.6}\\

\bottomrule
\end{tabular}}
\end{sc}
\end{scriptsize}
\end{adjustbox}
\vskip -0.1in
\end{table*}

\subsection{{Evaluation on the Effectiveness of Segmentation Algorithm}} \label{appendix: segmentation_transfer}

{To isolate the impact of the proposed hybrid segmentation algorithm, we replace EAGLE’s original segmentation with those generated by our method while holding all other settings unchanged. Table~\ref{tab: m4gn_segment_on_eagle} shows a clear gain: mesh‑quality metrics improve by 28\%–35\% and prediction errors fall by 15\%–23\%. When the same segmentation is paired with the full M4GN architecture, prediction errors are reduced even further and every mesh‑quality metric reaches its best value. These results indicate that (i) the proposed hybrid mesh segmentation alone contributes a portion of the improvement, and (ii) the architectural changes in M4GN provide an additional, complementary boost. Hence both components—better segmentation and the modified model—are necessary for the overall performance gains.}

\begin{table*}[t]
\caption{{Transfer‑segmentation experiment on the \textit{DeformingBeam} dataset.  Replacing EAGLE’s native segmentation with the proposed hybrid M4GN segmentation (second row) isolates the effect of segmentation quality, while the full M4GN row (third row) shows the additional benefit of our modified hierarchical architecture. Lower values indicate better performance.}}
\label{tab: m4gn_segment_on_eagle}
\begin{adjustbox}{width=\textwidth}
%\vskip 0.15in
\begin{scriptsize}
\begin{sc}
{
\begin{tabular}{lllccccccc} 
\toprule
  &  &  & \multicolumn{4}{c}{\textnormal{Mesh Quality Metrics $\downarrow$} }  &\multicolumn{3}{c}{\textnormal{Prediction Error Metrics $\downarrow$}}   \\
 \cmidrule(lr){4-7} \cmidrule(lr){8-10} 
Dataset  & Model & Segmentation & \begin{tabular}[c]{@{}c@{}} $\text{GF}_h$ \\ ($\times 10^{-3}$) \end{tabular} & \begin{tabular}[c]{@{}c@{}} $\text{GF}_c$ \\ ($\times 10^{-6}$) \end{tabular}& \begin{tabular}[c]{@{}c@{}} MC \\ ($\times 10^{-3}$) \end{tabular} & \begin{tabular}[c]{@{}c@{}} AR \\ ($\times 10^{-3}$) \end{tabular} & \begin{tabular}[c]{@{}c@{}} RMSE-1 \\ ($\times 10^{-5}$) \end{tabular} &  \begin{tabular}[c]{@{}c@{}} RMSE-50 \\ ($\times 10^{-4}$) \end{tabular}  &  \begin{tabular}[c]{@{}c@{}} RMSE-\textnormal{all} \\ ($\times 10^{-4}$) \end{tabular} \\
\midrule
\midrule
\multirow{3}{*}{\begin{tabular}[c]{@{}l@{}} Deforming\\Beam \end{tabular}}
& EAGLE   & EAGLE    & 0.64 \tiny$\pm$ 0.04 & {0.17 \tiny$\pm$ 0.01} & 5.98 \tiny$\pm$ 0.43 & {5.17 \tiny$\pm$ 0.37} &   1.51 \tiny$\pm$ 0.04 & 0.67 \tiny$\pm$ 0.12 & 4.22 \tiny$\pm$ 0.30\\
& EAGLE  & M4GN    & 0.46 \tiny$\pm$ 0.02 & {0.11 \tiny$\pm$ 0.00} & 5.43 \tiny$\pm$ 0.09 & {4.02 \tiny$\pm$ 0.10} &   1.28 \tiny$\pm$ 0.01 & 0.57 \tiny$\pm$ 0.04 & 3.27 \tiny$\pm$ 0.25\\
& M4GN  & M4GN       & \textbf{0.31 \tiny$\pm$ 0.01} & {\textbf{0.05 \tiny$\pm$ 0.00}} & \textbf{5.26 \tiny$\pm$ 0.04} & {\textbf{3.08 \tiny$\pm$ 0.06 }} & {\textbf{1.17 \tiny$\pm$ 0.01}} & {\textbf{0.34 \tiny$\pm$ 0.02}} & {\textbf{1.87 \tiny$\pm$ 0.12}}\\
\bottomrule
\end{tabular}}
\end{sc}
\end{scriptsize}
\end{adjustbox}
\vskip -0.1in
\end{table*}

\section{{Hyperparameter Sensitivity Analysis}}\label{appendix: ablation}

{This section presents a systematic sensitivity study, detailing how each hyperparameter affects performance and offering practical tuning guidelines. Importantly, the model already outperforms all baselines with default or minimally adjusted settings; additional tuning only refines an existing advantage rather than creating it. The following subsections analyze each hyperparameter in detail and Table~\ref{tab:hyperparam_summary} summarizes the key findings.}

\begin{table}[h]
\centering
\small
\begin{tabular}{p{2.5cm} p{2.2cm} p{5.7cm} p{4.5cm}}
\toprule
{\textbf{Hyper‐parameter}} & {\textbf{Ranges/ Variants}} & {\textbf{Observed impact on performance}} & {\textbf{Practical tuning guideline}} \\ \midrule
Message–passing steps ($n_{\text{MP}}$) & 1–8 & Few steps $\rightarrow$ under‑reach, large RMSE and discontinuities; moderate steps $\rightarrow$ improve both RMSE and mesh quality; very high steps $\rightarrow$ over‑smoothing and slight accuracy decay. & Start with 3-5 steps; increase while RMSE drops, stop when gains plateau or mesh quality stalls. \\ \addlinespace
Hybrid segmentation flavour & METIS, SLIC variants & SLIC variants outperform METIS; physics‑aware features give 7–35\,\% lower RMSE and better segment metrics; best overall: \textit{SLIC‑MDOD$_e$}. & Use \textit{SLIC‑MDOD$_e$} by default; avoid single‑cue variants unless domain knowledge shows otherwise. \\ \addlinespace
Number of segments ($N_{\text{seg}}$) & 3–51 & Accuracy fairly stable; too few segments $\rightarrow$ coarse resolution, RMSE rise; too many segments $\rightarrow$ diminishing returns followed by possible degradation and added overhead. & Compute Silhouette over candidate $N_{\text{seg}}$, then train 3–4 values near the peak; pick the point where RMSE/Chamfer stop improving. \\ \addlinespace
Positional encoding (PE) & off / on & Improves RMSE when $N_{\text{seg}}$ small and geometry simple; neutral or harmful when segments already fine or geometry complex. & Enable PE for coarse segmentations ($N_{\text{seg}}\!<\!15$) or 2‑D flows; disable for high‑resolution or highly deformable 3‑D meshes. \\ \addlinespace
Segment overlap ($\delta$) & $\delta=0$ (none), $\delta=1$ (one‑ring) & Helps Eulerian or directional meshes at high $N_{\text{seg}}$ (smoother transitions); can add redundancy and hurt Lagrangian cases. & Use $\delta=1$ for high $N_{\text{seg}}$ \emph{and} the mesh is fixed; keep $\delta=0$ for low $N_{\text{seg}}$. \\
\bottomrule
\end{tabular}
\caption{{Summary of key hyperparameters, their tested ranges, performance effects, and tuning recommendations derived from the sensitivity study in Appendix~\ref{appendix: ablation}.}}
\label{tab:hyperparam_summary}
\end{table}

\subsection{Message Passing Steps in Micro-level Module} \label{appendix: ablation_micro}
\begin{figure}[htbp]
    \centering
    \includegraphics[width=1\textwidth]{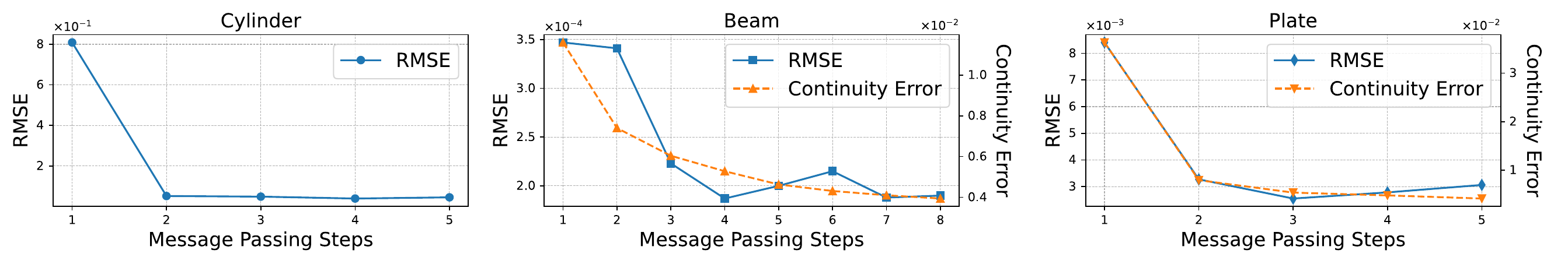}
    %\vskip -0.1in
    \caption{Ablation study on the impact of varying message-passing steps in the micro-level information exchange on prediction performance across three datasets.}
    \label{fig: ablation_micro_level}
\end{figure}

According to Figure~\ref{fig: ablation_micro_level}, with fewer message passing steps, each node updates only based on immediate neighbors, resulting in higher prediction errors and mesh discontinuities. As more steps are introduced, nodes gather information from a broader neighborhood, leading to more accurate predictions and smoother mesh transitions. The early iterations of message passing yield the most noticeable improvements, as nodes rapidly gather useful information from their surrounding environment. Later iterations primarily serve to fine-tune the mesh continuity and reduce local errors, but the impact on overall accuracy diminishes. Interestingly, increasing the number of message-passing steps beyond a certain point continues to improve mesh quality, but prediction accuracy may degrade. This suggests the occurrence of oversmoothing, where the model excessively homogenizes node features, or overfitting, where the model starts to memorize local information rather than generalize. This phenomenon highlights the importance of carefully selecting the number of message-passing steps during the micro-level information exchange step to strike the right balance between improving prediction accuracy and maintaining mesh quality.

\subsection{Variations of Hybrid Segmentation} \label{appendix: ablation_meso_1}
{As detailed in Appendix~\ref{appendix: mesh_segmentation}, the hybrid segmentation admits six variants depending on the selected feature set and transformation.  Table~\ref{tab: ablation_patch_method} summarizes the resulting model performance.  Compared with METIS alone, SLIC‑based variants lower \textit{RMSE\textsubscript{all}} by 16\,\%, 27\,\%, and 7\,\% on \textit{CylinderFlow}, \textit{DeformingPlate}, and \textit{DeformingBeam}, respectively, with {SLIC‐MDOD\textsubscript{e}} emerging as the best choice across all three datasets. In terms of the impact of modal decomposition features in SLIC, for the two solid‑mechanics cases (DeformingPlate and DeformingBeam), modal‑decomposition features enrich the descriptor space with physics‑relevant mode shapes, and  every OD+MD variant—SLIC‑MDOD$_l$ and SLIC‑MDOD$e$—beats its OD‑only counterpart across all metrics, delivering 24–35\% lower RMSE and up to 25\% better mesh‑quality scores. These consistent gains confirm that modal information and boundary‑distance cues are complementary for solid mechanics. On CylinderFlow, however, the flow is dominated by rapidly varying vortical patterns; the MD basis, derived purely from geometry in this Eulerian setting, adds little new information and can perturb the SLIC clusters, so SLIC‑MD alone shows a slight RMSE rise. When MD is combined with the (exponentially weighted) distance cue in SLIC‑MDOD$e$, the boundary‑aware term stabilizes the segmentation while the modal vectors still provide complementary detail, giving a net improvement over distance or modal information used in isolation.}

{Moreover, refining the coarse METIS partitions with SLIC improves accuracy only when the added SLIC features better align local cuts with the true physics; otherwise, the refinement can fragment physically coherent regions and hurt performance. SLIC‑OD, for example, uses only geometric distance to boundaries; on DeformingPlate this over‐weights proximity and splits mode‑consistent areas, so RMSE increases relative to the original METIS segmentation. Likewise, on DeformingBeam the distance‑only (SLIC‑OD) and modal‑only (SLIC‑MD) variants either ignore contact boundaries or long‐range bending modes, producing finer—but less meaningful—segments and therefore higher error than METIS. Only the combined SLIC‑MDOD\textsubscript{e}, which couples modal information with an exponentially weighted distance term, strikes the right balance between global coherence and local adaptation. }

{To thoroughly evaluate the different segmentation methods, we utilize the three metrics -Conductance, Edge Cut Ratio, and Silhouette Score - introduced in Appendix~\ref{appendix: segment_quality_metrics} to assess both inter-segment and intra-segment qualities of mesh partitions, providing a comprehensive understanding of each method's effectiveness. We then analyzed the correlation between these segmentation metrics and overall dynamic system performance, including mesh quality and prediction error, as illustrated in Figure~\ref{fig: segmentation_analysis_main} (a-c). Our findings indicate that segmentation methods incorporating physics-aware features, particularly those utilizing obstacle distances with exponential transformations, generally enhance model performance across various datasets. This improvement can be attributed to three key factors: (1) \textit{Alignment with Dynamics}, where segmentation reflecting physical influences enables more effective learning of the system's dynamics; (2) \textit{Enhanced Segment Quality}, achieved through improved intra-segment cohesion and minimized inter-segment interactions, facilitating better learning of localized patterns; and (3) \textit{Benefit to Learning}, where emphasizing critical regions via exponential transformations allows the model to focus on areas with significant dynamic changes, thereby enhancing prediction accuracy. These results demonstrate that the choice of segmentation method impacts the model's ability to learn dynamic behaviors, and the introduction of additional metrics reveals that physics-aware segmentation effectively aligns mesh partitions with the system's inherent physical properties, thereby benefiting the learning process.}

\begin{table}
\caption{Ablation study on different segment extraction methods over different dataset.}
\label{tab: ablation_patch_method}
%\vskip 0.05in
\begin{center}
\begin{scriptsize}
\begin{sc}
\centering
\begin{tabular}{llcccccc}
    \toprule
    Segmentation Method & Dataset & GF$_h$ $\downarrow$  & {GF$_c$ $\downarrow$}  & MC $\downarrow$ & {Aspect Ratio $\downarrow$} & RMSE-1 & RMSE-all \\
    % \midrule \midrule
    % \multirow{3}{*}{Grid}  & Cylinder  & - & 3.05e-03 & 4.05e-02 \\
    %                        & Plate & 4.86e-03 & 2.58e-04 & 2.05e-03 \\
    %                        & Beam & 5.20e-03 & 1.13e-05 & 2.33e-04 \\
    \midrule
    \multirow{3}{*}{METIS} & Cylinder & -&- & - & - & 3.44e-03 & 4.59e-02 \\
                           & Plate & 5.32e-03 & {1.36e-05} & 5.33e-03 & {2.97e-03} & 2.67e-04 & 3.29e-03 \\
                           & Beam & 3.88e-04 & {5.61e-08} & 5.18e-03 & {3.09e-03} & 1.15e-05 & 2.16e-04 \\
    \midrule
    \multirow{3}{*}{SLIC-OD} & Cylinder &- &- & - & - & 3.28e-03 & 4.40e-02\\
                           & Plate & 5.24e-03 & 1.50e-05 & 5.16e-03 & 3.04e-03  & 2.69e-04 & 3.70e-03 \\
                           & Beam & 3.78e-04 & 5.41e-08  & 5.39e-03 & 3.60e-03 & 1.21e-05 & 2.31e-04\\
    \midrule
    \multirow{3}{*}{SLIC-OD$_{l}$} &  Cylinder &- &- & - & - & 3.33e-03 & 4.37e-02\\
                           & Plate & 5.11e-03 & {8.01e-06} & 5.09e-03 & {2.90e-03} & 2.81e-04 & 3.44e-03 \\
                           & Beam & 3.95e-04& {5.44e-08} & 5.33e-03 & {3.30e-03} & 1.18e-05 & 2.68e-04\\
    \midrule
    \multirow{3}{*}{SLIC-OD$_{e}$} &  Cylinder &- &- & - & - &3.21e-03 & 3.95e-02\\
                           & Plate & 5.27e‑03 & {1.31e‑05} & 4.58e‑03 & {2.54e‑03} & 2.61e‑04 & 3.51e‑03 \\
                           & Beam & 3.81e-04 & {5.68e-08} & 5.40e-03 & {3.32e-03} & 1.20e-05 & 2.51e-04\\
    % \midrule
    % \multirow{3}{*}{SLIC-ODBD-log}  & Cylinder & - & - & - & - & 2.99e-03 & 5.57e-02\\
    %                                & Plate & 5.11e-03 & \YH{1.26e-05} & 4.93e-03 & \YH{2.76e-03} & 2.78e-04 & 3.57e-03\\
    %                                & Beam &3.74e-04 & \YH{5.48e-08} & 5.28e-03 & \YH{3.26e-03} & 1.18e-05 & 2.46e-04\\                           
    % \midrule
    % \multirow{3}{*}{SLIC-ODBD-exp}  & Cylinder  &- & - & - & - & 2.95e-03 & 4.37e-02 \\
    %                                & Plate & 4.33e-03 & \YH{6.85e-06} & 4.66e-03 & \YH{2.63e-03} & 2.66e-04 & 2.57e-03 \\
    %                                & Beam & 3.21e-04 & \YH{4.59e-08} & 5.28e-03 & \YH{3.20e-03} & 1.17e-05 & 1.87e-04\\
    \midrule
    \multirow{3}{*}{{SLIC-MD}}   & Cylinder  & - & - & - & - & 3.16e-03 & 5.62e-02 \\
                                    & Plate & 5.10e-03 & 8.38e-6 & 4.67e-03 & 2.53e-03 & 2.74e-04 & 3.02e-03\\
                                    & Beam  & 3.81e-04 & 5.45e-08 & 5.32e-03 & 3.20e-03 & 1.17e-05 & 2.32e-04\\
    \midrule
    \multirow{3}{*}{{SLIC-MDOD$_{l}$}} & Cylinder  & - & - & - & - & 4.16e-03 & 5.29e-02 \\ 
                                            & Plate & 4.84e-03 & 7.23e-06 & 4.56e-03 & 2.47e-03 & 2.68e-04 & 2.82e-03 \\
                                             & Beam  & 3.53e-03 & 5.10e-08 & 5.29e-03 & 3.40e-03 & 1.22e-05 & 2.25e-04 \\
    \midrule
    \multirow{3}{*}{{SLIC-MDOD$_{e}$}} & Cylinder  & - & - & - & - & 3.09e-03 & 3.86e-02 \\ 
                                            & Plate & 4.26e-03 & 6.49e-06 & 4.73e-03 & 2.58e-03 & 2.71e-04 & 2.40e-03 \\
                                             & Beam  & 3.02e-03 & 4.47e-08 & 5.31e-03 & 3.08e-03 & 1.17e-05 & 2.01e-04 \\
    % \multirow{3}{*}{Potential} & Cylinder  & - & 3.44e-03 & 5.23e-02\\
    %                        & Plate & 5.89e-03 & 2.82e-04 & 3.57e-03\\
    %                        & Beam & 5.25e-03 & 1.20e-05 & 2.30e-04\\
    %\midrule
    \bottomrule
\end{tabular}
\end{sc}
\end{scriptsize}
\end{center}
%\vskip -0.1in
\end{table}

\subsection{Number of Segmentation} \label{appendix: ablation_meso_2}
Table~\ref{tab:ablation_mp_hop} and Table~\ref{tab:ablation_PE} present the RMSE-1, RMSE-all, and various mesh quality metrics as the total number of mesh segments is varied during training on three different datasets. In general, M4GN maintains stable performance with relatively low variance, indicating that results are not highly sensitive to segment count. This robustness ensures reliable accuracy across different mesh granularities. However, increasing the number of segments—thereby reducing finite elements per segment—can lead to slight decreases in accuracy and performance.

\begin{figure}[htbp]
    \centering
    \includegraphics[width=0.8\textwidth]{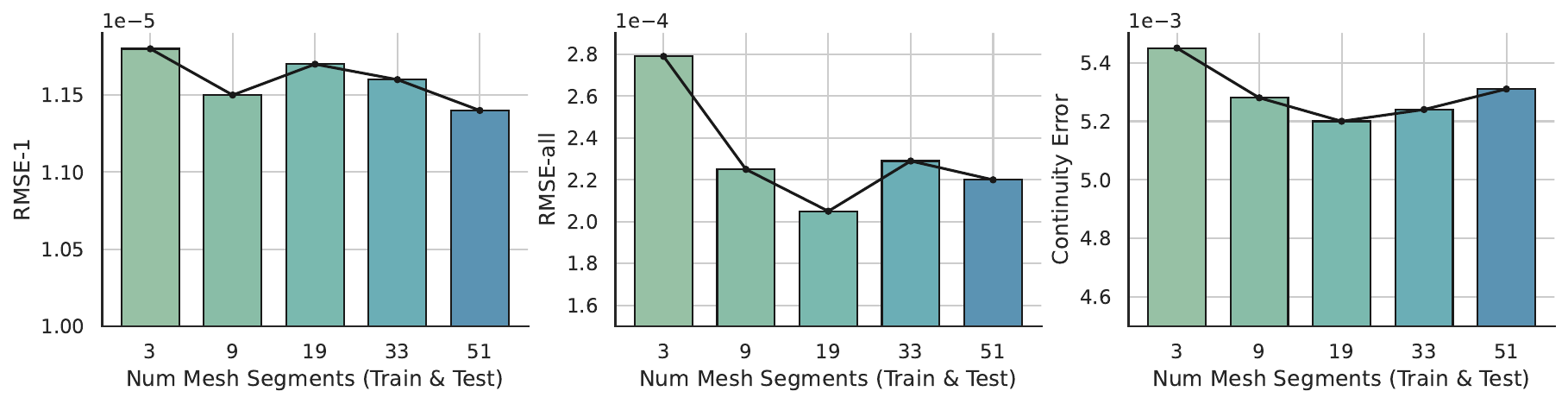}
    %\vskip -0.1in
    \caption{ Impact of varying mesh segment numbers during training on prediction accuracy under the DeformingBeam dataset. The number of mesh segments remains consistent during both training and testing. In general, M4GN maintains stable performance with relatively low variance, indicating that results are not highly sensitive to segment count. This robustness ensures reliable accuracy across different mesh granularities. However, increasing the number of segments—thereby reducing finite elements per segment—can lead to slight decreases in accuracy and performance. {More detailed analysis on the effect of segmentation numbers on various metrics can be found in Figure~\ref{fig: rebuttal_segment_analysis}.}}
    \label{fig: deformingBeam_RMSEall&MC_vs_nSeg}
\end{figure}

\begin{figure}[htbp]
    \centering
    \includegraphics[width=0.55\textwidth]{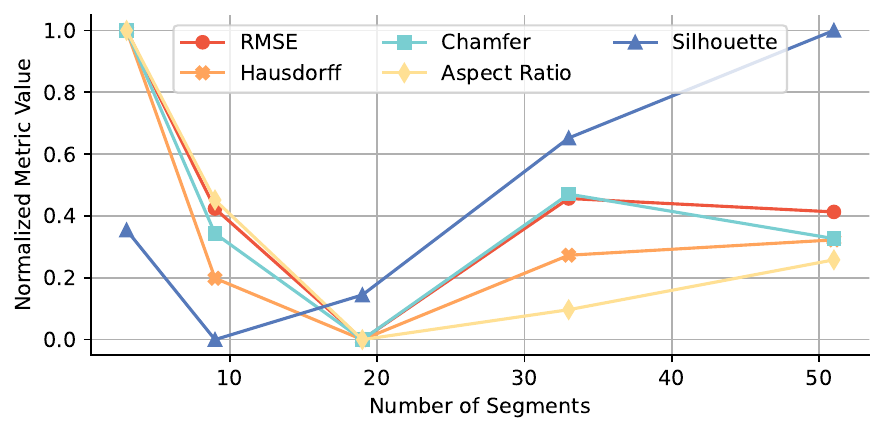}
    %\vskip -0.1in
    \caption{Dependence of various performance metrics on the number of segments in M4GN under Deforming Beam dataset. The plot illustrates how the normalized values of several performance metrics vary with the number of segments. Each metric is represented by a distinct curve, demonstrating the relationship between segment number and overall performance. This figure evaluates the effect of segment number and guides the selection of the optimal number of segments for balanced performance across all metrics.}
    \label{fig: rebuttal_segment_analysis}
\end{figure}

{To comprehensively evaluate the effect of segment number and determine the optimal segmentation for a given dataset, we analyzed prediction accuracy across a wide range of segment counts (from 3 to 51) during training on the DeformingBeam dataset. The impact of varying the number of mesh segments on prediction accuracy is illustrated in Figure~\ref{fig: deformingBeam_RMSEall&MC_vs_nSeg} and Figure~\ref{fig: rebuttal_segment_analysis}. According to the plots, we identify 19 segments as the optimal number. At this segmentation level, the model achieves the lowest RMSE and Chamfer Distance, indicating high prediction accuracy and precise shape representation. The Hausdorff Distance is also minimized, reflecting excellent alignment between the predicted and true meshes. While the Silhouette score peaks at 9 segments—suggesting well-defined and compact clusters—the slight decrease at 19 segments is offset by significant gains in other performance metrics. Choosing a lower number of segments, such as 3 or 9, may result in higher Silhouette scores but can compromise mesh detail and prediction accuracy due to insufficient spatial granularity. Conversely, selecting a higher number of segments beyond 19 shows diminishing returns, with only marginal improvements or slight degradations in some metrics and a continued decline in Silhouette scores, potentially indicating over-segmentation and unnecessary computational complexity.}

{In conclusion, when presented with a new dataset, the optimal number of segments can be determined by first computing Silhouette scores for various segment counts to assess cluster cohesion and separation without requiring model training. This provides initial guidance on meaningful segmentation levels. Subsequently, training the model with different segment numbers and evaluating performance metrics like RMSE, Hausdorff Distance, and Chamfer Distance will help identify the point where performance improvements plateau or begin to reverse, indicating the optimal balance between segmentation detail and model efficacy.}
%The results demonstrate that our method maintains relatively consistent performance regardless of the number of segments. The error metrics are largely insensitive to the segment count, with the exception of a slightly higher RMSE-all when fewer segments are used. \YH{[Refer to the new figure and add analysis.]}

\subsection{Influence of Positional Encoding}\label{appendix: ablation_macro_1}
Table~\ref{tab:ablation_PE} and Figure~\ref{fig: rebuttal_PE_delta_ablation}(a) show the effect of adding positional encoding for small and large numbers of segments across three datasets. According to the results, we identified several key findings. Firstly, the effectiveness of PE depends on the number of segments: in the CylinderFlow and Deforming Plate datasets, incorporating PE with fewer segments improves performance across multiple metrics by reducing positional ambiguity. With low segment counts, each segment covers larger, more diverse areas, limiting the model’s spatial detail and understanding of segment relationships. PE provides explicit positional information, allowing the model to distinguish distinct regions within the same segment and better comprehend their interactions. However, as the number of segments increases and spatial resolution improves, the benefits of PE diminish and may even introduce unnecessary complexity that hinders performance. Additionally, dataset-specific factors influence PE’s effectiveness; for example, the DeformingBeam dataset, with its complex geometry and deformation, did not benefit from PE. This indicates that PE’s success depends not only on segment count but also on how well the PE implementation aligns with the dataset’s unique characteristics. Consequently, tailored PE approaches that consider specific geometry and deformation patterns are necessary for complex systems to achieve performance gains. In summary, while PE enhances the performance of graph-based networks, further advancements are needed to develop optimal encoding strategies that consistently improve performance across diverse dynamic systems.

\subsection{Influence of Segment Overlap}\label{appendix: ablation_macro_2}
\begin{figure}[htbp]
    \centering
    \includegraphics[width=1\textwidth]{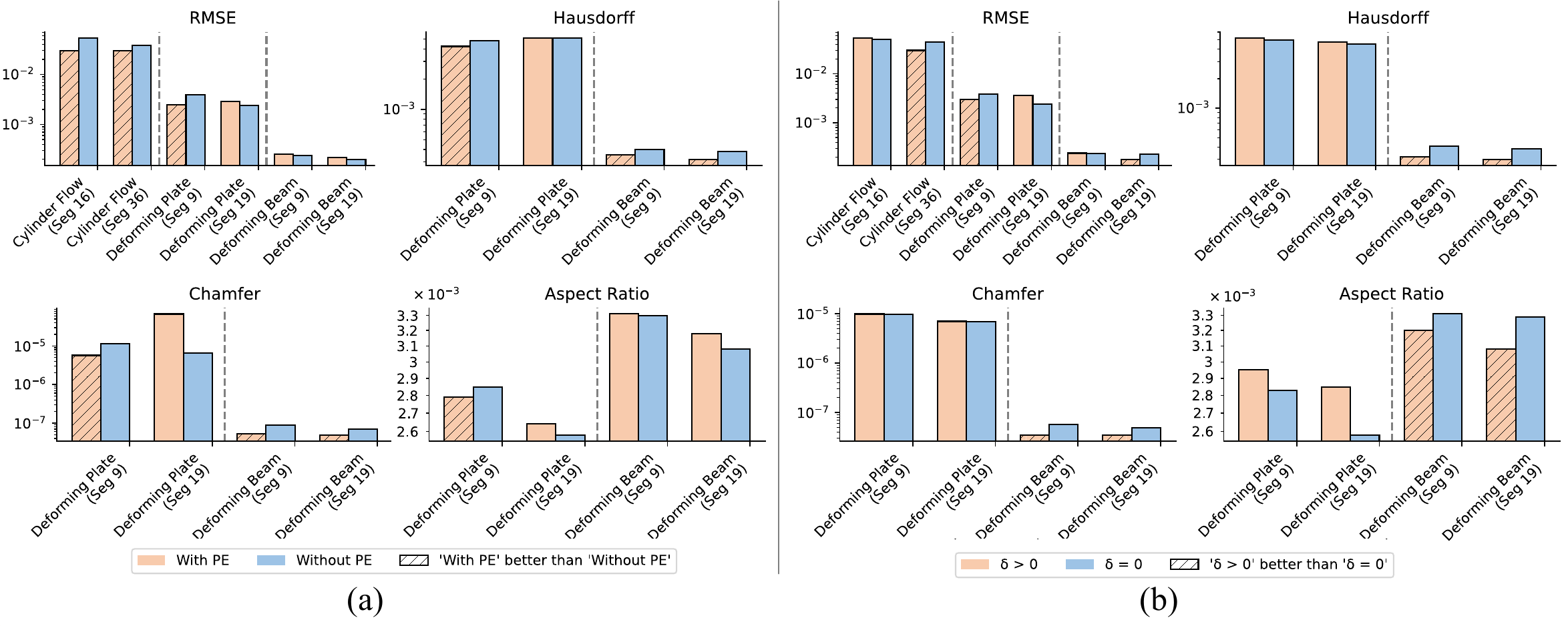}
    %\vskip -0.1in
    \caption{{Ablation study on the effects of position encoding and segment overlap across datasets with varying segment numbers.} The figure presents the performance metrics for models both with and without the position encoder (a), and with and without considering segment overlap (b) across three distinct datasets, each characterized by a different number of segments. By comparing these conditions, the study highlights how the inclusion of position encoding and the handling of segment overlap influence overall performance, thereby informing the selection of optimal model configurations.}
    \label{fig: rebuttal_PE_delta_ablation}
\end{figure}

{Table~\ref{tab:ablation_mp_hop} and Figure~\ref{fig: rebuttal_PE_delta_ablation}(b) illustrate the effect of adding segment overlap for small and large number of segments across three datasets. According to the results, the effectiveness of adding overlap between segments ($\delta$ = 1) depends on both the segment count and the characteristics of the dataset, such as dimensionality, mesh type, and system dynamics. Overlapping segments are more beneficial with higher segment counts where discontinuities are more prevalent. In Eulerian systems, overlaps enhance the capture of complex interactions and smooth transitions on fixed meshes, leading to improved representation of fluid dynamics. Conversely, in Lagrangian systems where meshes move with the material, overlaps can create redundancy and complicate connectivity, with their impact on model performance varying based on mesh structures and deformation behaviors. For example, in the Deforming Beam dataset, which uses a prism mesh suited for directional deformation, overlapping segments improve performance by facilitating smooth transitions along its mesh surface, especially with a higher number of segments. In contrast, the Deforming Plate dataset employs a tetrahedral mesh with complex, isotropic deformations, where overlaps introduce unnecessary complexity and redundancy, resulting in decreased performance. Therefore, despite both being 3D Lagrangian systems, the different mesh types and deformation patterns explain why overlapping segments benefit the Deforming Beam but not the Deforming Plate.}

\begin{table}
%\begin{minipage}{0.5\linewidth}
\caption{Ablation study of number of segments, and effect of adding segment overlap.}
\label{tab:ablation_mp_hop}
%\vskip 0.05in
\centering
\begin{scriptsize}
\begin{sc}
%\resizebox{\linewidth}{!}{
\begin{tabular}{lcccccccc}
\toprule
   Dataset & $N_\text{seg}$  & Overlap & {GF$_h$ $\downarrow$} & {GF$_c$ $\downarrow$} & MC $\downarrow$ & {Aspect Ratio} $\downarrow$ & RMSE-1 & RMSE-all\\
\midrule
   \multirow{4}{*}{Cylinder}    %& 16  & \xmark  &- &- & - & - & 3.07e-03 & 3.96e-02 \\
                                &16 & \xmark  &- &- & - & - & 3.16e-03 & 5.03e-02 \\
                                %&16  & \cmark  & -&- & - & - & 3.02e-03  & 4.09e-02 \\
                                &16 & \cmark  &- &- & - & - & 3.19e-03 & 5.35e-02 \\
                                %&36 & \xmark  &- &- & - & - & 3.58e-03 & 4.63e-02 \\
                                &36 & \xmark  &- &- & - & - & 3.41e-03 & 4.42e-02 \\
                                %&36 & \cmark &- & -& -  & - & 2.88e-03  & 3.34e-02 \\ 
                                &36 & \cmark  &- &- & - & - & 3.09e-03 & 3.86e-02 \\
\midrule                                
   \multirow{4}{*}{Plate} 
                                %&9  & \xmark  & 4.89e-03 & \YH{9.56e-06} & 5.01e-03  & \YH{2.78e-03} & 2.83e-04 &  3.05e-03 \\
                                & 9 & \xmark & 4.98e-03 & 9.58e-06 & 5.01e-03 & 2.83e-03 & 2.77e-04 & 3.88e-03\\ 
                                %&9  & \cmark  & 5.96e-03 & \YH{1.34e-05} & 5.10e-03 & \YH{2.90e-03} & 2.75e-04 & 3.64e-03 \\
                                &9 & \cmark & 5.32e-03 & 9.87e-06 & 5.24e-03 & 2.95e-03 & 2.83e-04 & 2.98e-03 \\ 
                                %&19 & \xmark & 4.33e-03 & \YH{6.85e-06} & 4.66e-03 & \YH{2.63e-03} & 2.66e-04 & 2.57e-03 \\
                                &19 & \xmark & 4.51e-03 & 6.91e-06  & 4.73e-03 & 2.58e-03 & 2.71e-04 & 2.40e-03\\ 
                                %&19 & \cmark & 5.08e-03 &  \YH{2.18e-05} & 4.76e-03 & \YH{2.72e-03} & 2.65e-04 & 4.49e-03 \\ 
                                &19 & \cmark & 4.76e-03 & 7.01e-06 & 4.81e-03 & 2.85e-03 & 2.77e-04 & 3.59e-03 \\ 

\midrule
   \multirow{4}{*}{Beam}        
                                %&9 & \xmark & 3.45e-04 & \YH{5.21e-08} & 5.18e-03 & \YH{3.30e-03} & 1.14e-05 & 2.25e-04 \\
                                & 9 & \xmark & 3.46e-04 & 5.23e-08 & 5.17e-03 & 3.31e-03 & 1.14e-05 & 2.39e-04 \\ 
                                %&9 & \cmark  & 3.67e-04 & \YH{5.38e-08} & 5.25e-03 & \YH{3.32e-03} & 1.18e-05 & 2.37e-04 \\
                                & 9 & \cmark & 3.38e-04 & 5.07e-08 & 5.31e-03 & 3.30e-03 & 1.15e-05 & 2.40e-04 \\ 
                                %&19 & \xmark & 3.29e-04 & \YH{4.81e-08}  & 5.18e-03 & \YH{3.26e-03} & 1.19e-05 & 2.05e-04 \\
                                & 19 & \xmark & 3.57e-04 & 4.92e-08 & 5.24e-03 & 3.29e-03 & 1.18e-05 & 2.28e-04 \\ 
                                %&19 & \cmark & 3.21e-04 & \YH{4.59e-08} & 5.28e-03 & \YH{3.20e-03} & 1.17e-05 & 1.87e-04 \\ 
                                & 19 & \cmark & 3.19e-04 & 4.73e-08 & 5.31e-03 & 3.08e-03 & 1.17e-05 & 2.01e-04 \\ 
\bottomrule
\end{tabular}
\end{sc}
\end{scriptsize}
%\end{minipage}
%\hfill
\end{table}
%\begin{minipage}{0.48\linewidth}
\begin{table}
\caption{Ablation study of number of segments and whether to add PE or not.}
\vspace{0.05in}
\label{tab:ablation_PE}
\centering
\begin{scriptsize}
\begin{sc}
%\resizebox{\linewidth}{!}{
\begin{tabular}{lcccccccc}
\toprule
   Dataset & $N_\text{seg}$  & PE & GF$_h$ $\downarrow$  & {GF$_c$ $\downarrow$} & MC $\downarrow$ & {Aspect Ratio $\downarrow$} & RMSE-1 & RMSE-all\\
\midrule
\multirow{4}{*}{Cylinder}   %& 16  & \xmark &-& -& - & - & 3.02e-03 & 4.09e-02 \\
                            &16 & \xmark  &- &- & - & - & 3.19e-03 & 5.35e-02 \\
                            %& 16  & \cmark &- &- &- & - & 2.95e-03 & 3.69e-02\\
                            &16 & \cmark  &- &- & - & - & 3.34e-03 & 4.76e-02 \\
                            %& 36 & \xmark & -&- & - & - & 2.88e-03 & 3.34e-02\\
                            &36 & \xmark  &- &- & - & - & 3.09e-03 & 3.86e-02 \\
                            %& 36 & \cmark & -&- & - & - & 3.23e-03 & 4.10e-02\\
                            &36 & \cmark  &- &- & - & - & 3.00e-03 & 3.80e-02 \\

\midrule
\multirow{4}{*}{Plate} 
                            %& 9  & \xmark & 4.89e-03 & \YH{9.56e-06} & 5.01e-03 & \YH{2.78e-03} & 2.83e-04 & 3.05e-03\\
                            &9 & \xmark & 4.81e-03 & 9.33e-06 & 5.01e-03 & 2.83e-03 & 2.77e-04 & 3.88e-03 \\ 
                            %& 9  & \cmark & 4.39e-03 & \YH{6.94e-06} & 5.11e-03 & \YH{2.98e-03} & 2.86e-04 & 2.57e-03 \\
                            &9 & \cmark & 4.24e-03 &6.72e-06 & 5.04e-03 & 2.81e-03 & 2.84e-04 & 2.72e-03 \\
                            %& 19 & \xmark & 4.33e-03 & \YH{6.85e-06} & 4.66e-03 & \YH{2.63e-03} &2.66e-04 & 2.57e-03 \\
                            &19 & \xmark & 5.10e-03 & 6.51e-06 & 4.73e-03 & 2.58e-03 & 2.71e-04 & 2.40e-03 \\ 
                            %& 19 & \cmark & 5.21e-03 & \YH{1.34e-05} & 4.76e-03 & \YH{2.62e-03} & 2.81e-04 & 3.67e-03 \\
                            &19 & \cmark & 5.13e-03 & 6.78e-06 & 4.74e-03 & 2.64e-03 & 2.68e-04 & 2.91e-03 \\ 
\midrule
\multirow{4}{*}{Beam} 
                            %& 9  & \xmark & 3.67e-04 & \YH{5.38e-08} & 5.25e-03 & \YH{3.32e-03} & 1.18e-05 & 2.37e-04\\
                            &9 & \xmark & 3.75e-04 & 5.89e-08 & 5.31e-03 & 3.30e-03 & 1.15e-05 & 2.40e-04 \\ 
                            %& 9  & \cmark & 3.93e-04 & \YH{6.54e-08} & 5.32e-03 & \YH{3.43e-03} & 1.15e-05 & 2.58e-04 \\
                            &9 & \cmark & 3.51e-04 &5.33e-08  & 5.18e-03 & 3.31e-03 & 1.17e-05 & 2.56e-04 \\ 
                            %& 19 & \xmark & 3.21e-04 & \YH{4.59e-08} & 5.28e-03 & \YH{3.20e-03} &1.17e-05 & 1.87e-04\\
                            &19 & \xmark & 3.22e-04 & 4.86e-08 & 5.31e-03 & 3.08e-03 & 1.17e-05 & 2.01e-04 \\ 
                            %& 19 & \cmark & 3.33e-04 & \YH{5.15e-08} & 5.22e-03 & \YH{3.27e-03} &1.17e-05 & 1.88e-04\\
                            &19 & \cmark & 3.19e-04 & 4.84e-08 & 5.27e-03 & 3.18e-03 & 1.15e-05 & 2.21e-04 \\ 
                            
\bottomrule
\end{tabular}
\end{sc}
\end{scriptsize}
%\end{minipage}
%\vskip -0.1in
\end{table}

\section{Generalization Studies}\label{appendix: generalization}
To evaluate the generalizability of our M4GN model, we created a larger-scale DeformingBeam dataset, detailed in Appendix~\ref{appendix: dataset}.
\subsection{Performance on Larger-Scale Datasets} \label{appendix: generalization_large_scale}

\begin{figure}[htbp]
    \centering
    \includegraphics[width=0.8\textwidth]{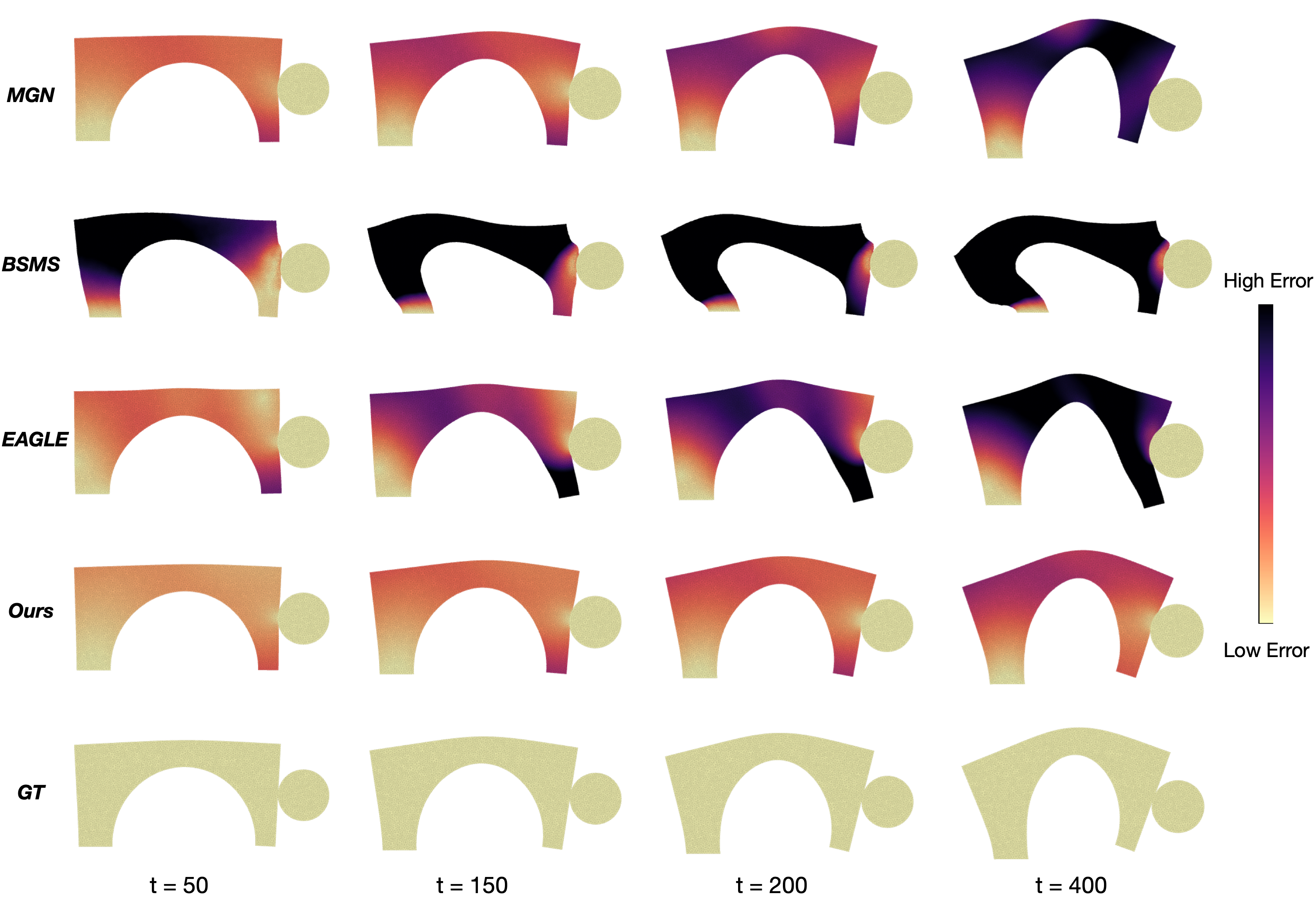}
    %\vskip -0.1in
    \caption{{Generalization results for different models under DeformingBeam(large) dataset.}}
    \label{fig: generalization_rollout_1}
\end{figure}

Table~\ref{tab: generalization_result} summarizes the generalization performance of various models trained on the DeformingBeam dataset and directly applied to DeformingBeam(large), a scaled-up version. The results demonstrate that M4GN consistently outperforms all other models across all metrics. In terms of mesh quality, M4GN achieves a 53\% improvement over both EAGLE and BSMS for Geometric Fidelity (GF). Similarly, for Mesh Continuity (MC), M4GN achieves the best performance with a value of 1.08e-02, representing a 45\% improvement over EAGLE, the next-best model. For the RMSE metrics, M4GN delivers the lowest RMSE-1, RMSE-50, and RMSE-all. Notably, M4GN's RMSE-all is 46\% lower than EAGLE. These findings suggest that M4GN not only preserves prediction accuracy but also enhances mesh quality when generalizing to larger-scale data, significantly surpassing state-of-the-art models in both accuracy and mesh quality. This demonstrates M4GN’s robust generalization ability, making it highly suitable for complex, large-scale dynamical systems. {Figure~\ref{fig: generalization_rollout_1} is a visualization of generalization results on DeformingBeam(large) dataset for different models. }
\begin{table}[h]
\caption{Generalization performance of our method and five baseline models on the DeformingBeam(large) dataset. M4GN demonstrates great accuracy and mesh quality when generalizing to an unseen dataset with a denser mesh and more extensive long-range dynamic effects.}
\label{tab: generalization_result}
\begin{center}
\begin{scriptsize}
\begin{sc}
\centering
\begin{tabular}{lcccccccc}
\toprule
   Method    &  GF$_h$ $\downarrow$ &  {GF$_c$ $\downarrow$} & MC $\downarrow$ & {Aspect Ratio $\downarrow$} &  RMSE-1  & RMSE-50   & RMSE-all   \\
\midrule
   GCN & 2.18e-02 & {3.28e-05} & 1.21e-01 & {1.69e-01} & 2.57e-04 & 1.95e-03 & 1.11e-02\\
   $g$-U-Net  & 1.94e-02 & {2.80e-05} & 4.56e-02 & {7.01e-02} & 1.60e-04 & 1.87e-03 & 1.01e-02\\
   MGN   & 2.32e-02 & {1.43e-05} & 2.00e-02  & {2.57e-02} & 1.34e-04 & 1.43e-03 & 6.42e-03\\
   BSMS  & 1.72e-02 & {3.34e-05} & 1.35e-01 & {1.17e-01} & 4.47e-04 & 3.19e-03 & 1.03e-02\\
   EAGLE  & 1.69e-02 & {2.20e-05} & 1.98e-02 & {5.15e-02} & 8.42e-05 & 1.45e-03 & 8.37e-03\\
   %M4GN & \textbf{8.25e-03} & \YH{\textbf{5.59e-06}} & \textbf{1.13e-02} & \YH{\textbf{2.16e-02}} & \textbf{5.84e-05} & \textbf{9.43e-04} & \textbf{4.59e-03}\\
   M4GN & \textbf{7.96e-03} & \textbf{5.35e-06} & \textbf{1.08e-02} & \textbf{2.24e-02} &\textbf{5.47e-05} & \textbf{9.20e-04} & \textbf{4.58e-03} \\
\bottomrule
\end{tabular}
\end{sc}
\end{scriptsize}
\end{center}
\end{table}

\subsection{Effect of Mesh Segment Count on Generalization}\label{appendix: generalization_num_seg}
\begin{figure}[htbp]
    \centering
    \includegraphics[width=\textwidth]{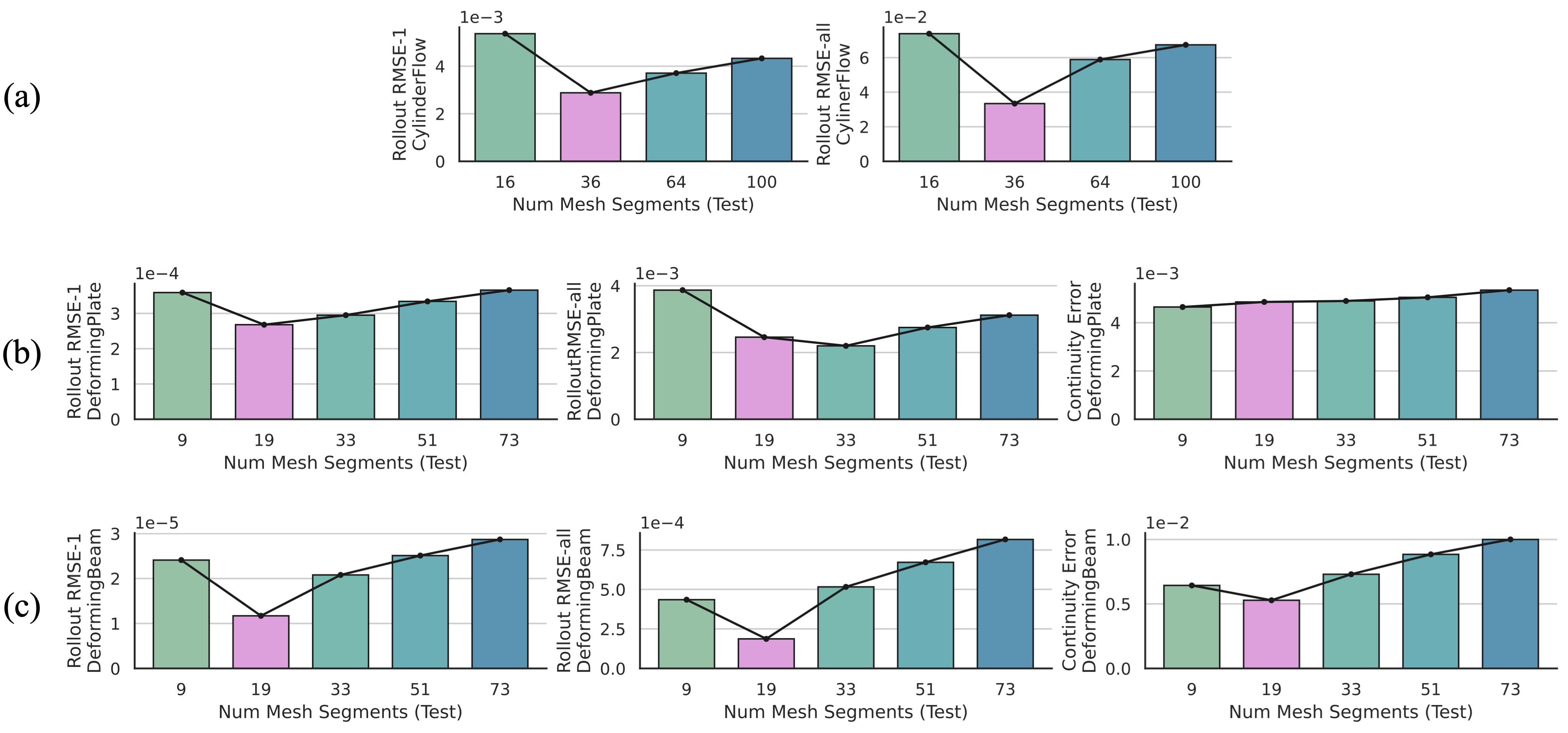}
    %\vskip -0.1in
    \caption{Generalization performance of our method under varying segment counts during testing over three datasets. (a) CylinderFlow: effect of number of segments for test set on different metrics, where model is trained under 36 segments (colored in pink); (b) DeformingPlate: effect of number of segments for test set on different metrics, where model is trained under 19 segments (colored in pink); (c) DeformingBeam: effect of number of segments for test set on different metrics, where model is trained under 19 segments (colored in pink). This figure illustrates that our M4GN model, despite being trained with a fixed number of mesh segments, maintains strong accuracy and mesh quality when tested with varying numbers of mesh segments.}
    \label{fig: all_dataset_test_nSeg}
\end{figure}

\textbf{Generalization with Varying Segment Counts During Testing} --
Across three datasets, we perform generalization studies where the model is tested using a varying number of segments. The results in Figure~\ref{fig: all_dataset_test_nSeg} illustrate the generalization performance. Pink columns are the references for regular testing and the others are generalization to different number of segments from training. Overall, the M4GN model can generalize very well to different number of segments during testing. 

\begin{figure}[htbp]
    \centering
    \includegraphics[width=0.8\textwidth]{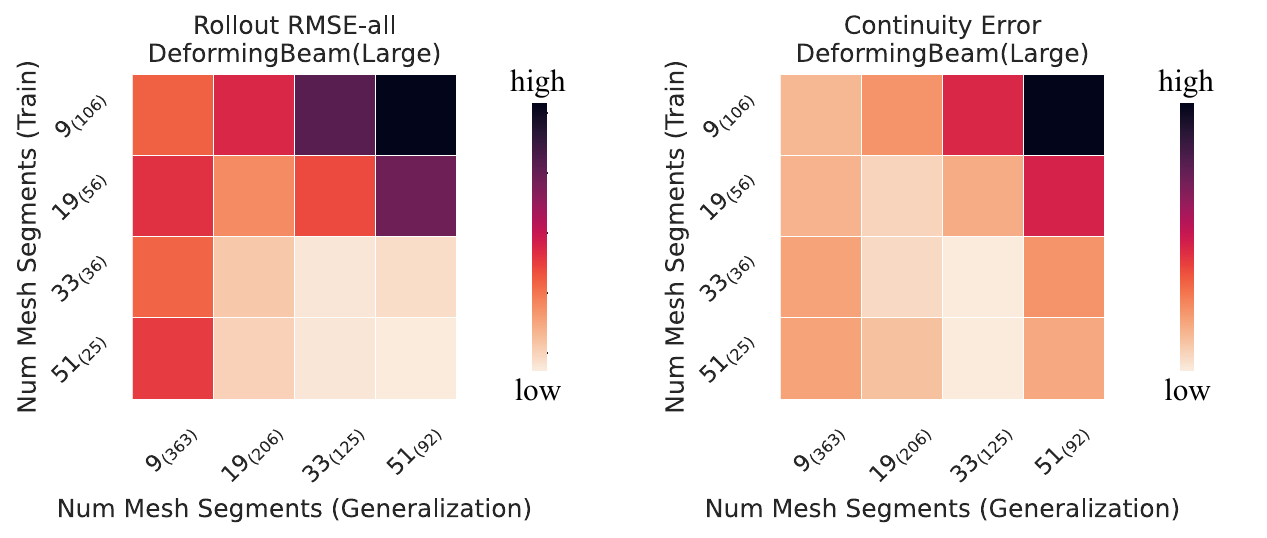}
    %\vskip -0.1in
    \caption{Generalization performance of our method on larger domains under different number of mesh segmentation during training and testing. The subscript of each mesh segment indicating the average number of nodes per segment. M4GN demonstrates robustness and adaptability in handling larger domains with varying mesh segments, making it well-suited for real-world applications involving large and complex mesh structures.}
    \label{fig: generalization_heatmap}
\end{figure}
\textbf{Impact of Segment Count During Training and Testing} --  Equipped with message passing and transformer mechanisms, M4GN can handle an arbitrary number of segments. Figure~\ref{fig: generalization_heatmap} shows the generalization performance of our M4GN model to larger domain as heatmaps, where models trained with a specific number of segment under deformingBeam dataset are tested with varying number of segments under deformingBeam (large). We observe that better results are seen when the number of nodes per segment during training is less than or equal to that in the generalizing domain, or when the number of segments is greater. Overall, we demonstrate M4GN's robustness and adaptability in generalizing to larger domains with varying mesh segments, making it highly suitable for real-world applications involving large and diverse mesh graphs.

\begin{table}
\renewcommand{\arraystretch}{1.1}
\caption{Comprehensive evaluation of our method alongside MGN, BSMS, and EAGLE under three datasets. M4GN consistently delivers stable, competitive efficiency while maintaining high accuracy and mesh quality.}
\label{tab: ablation_complexity}
\begin{sc}
\begin{center}
\begin{scriptsize}
\centering

\begin{tabular}{llccccccc}
\toprule
   Dataset & Model & RMSE-all & MC $\downarrow$ & 
   \begin{tabular}[c]{@{}l@{}} Train Time\\ per step [ms] $\downarrow$\end{tabular} & 
   \begin{tabular}[c]{@{}l@{}} Train \\ Memory [MB] $\downarrow$\end{tabular} & 
   \begin{tabular}[c]{@{}l@{}} Test Time\\ per step [ms] $\downarrow$\end{tabular} & 
   \begin{tabular}[c]{@{}l@{}} Test \\ Memory [MB] $\downarrow$\end{tabular} & 
   {\begin{tabular}[c]{@{}l@{}} Train Time\\ total [h] $\downarrow$\end{tabular}}\\
\midrule
\multirow{4}{*}{Cylinder} & MGN              & 4.81e-02 & - & 66.7 & 698.5 & 20.2 & 67.2 & {37.1}\\
                          & BSMS             & 1.37e-01 & - & {54.7} & 430.3 & 23.8 & {57.9} & {30.4}\\
                          & EAGLE            & 5.83e-02 & - & 69.5 & 618.7 & 28.8 & 230.8 & {38.6}\\
                          & M4GN            & {3.80e-02} & -  & 56.2 & {366.6} & {20.0} & 65.0 & {31.2}\\
\midrule
\multirow{4}{*}{Plate}    & MGN              & 1.47e-02 & 9.25e-03  & 131.9 & 6021.5 & 36.2 & 445.5 & {73.3}\\
                          & BSMS             & 1.18e-02 & 1.83e-02  & 83.9 & 910.1 & 37.7 & {77.9} & {46.6}\\
                          %& EAGLE            & 6.06e-03 & 6.18e-03  & 80.2 & 1012.5 & 33.0 & 360.2\\
                          & EAGLE            & 3.87e-03 & 5.56e-03  & 81.2 & 1090.8 & 32.4 & 362.7 & {45.1}\\
                          & M4GN            & {2.65e-03} & {4.82e-03}  & {76.5} & {648.1} & {29.3} & 103.3 & {42.5}\\
\midrule
\multirow{4}{*}{Beam}     & MGN              & 4.72e-04 & 1.69e-02  & 79.1 & 1074.4 & 28.6 & 83.8 & 22.0\\
                          & BSMS             & 4.98e-04 & 3.25e-02  & 61.8 & {213.7} & 30.7 & {35.6} & {17.2}\\
                          %& EAGLE            & 5.36e-04 & 6.33e-03 & 45.4 & 402.6 & 24.3 & 150.2\\
                          & EAGLE            & 4.22e-04 & 5.98e-03  & 53.5 & 410.3 & 26.0 & 153.5 & {14.9}\\
                          & M4GN            & {1.87e-04} & {5.26e-03}  & {53.4} & 234.5 & {24.2} & 47.1 & {14.8}\\
% \midrule
% \multirow{4}{*}{Beam-Large} & MGN            & 6.42e-03 & 2.00e-02 & 3850754 & - & - & 43.9 & 120.9\\
%                           & BSMS             & 1.03e-02 & 1.35e-01 & 2004738 & - & - & 42.1 & 88.5\\
%                           & EAGLE            & 9.33e-03 & 1.53e-02 & 10317314 & - & - & 42.2 & 391.1\\
%                           & EAGLE (wedge)    & 8.37e-03 & 1.98e-02 & 11041794 & - & - & 47.7 & 365.2\\
%                           & MMSGN            & 4.59e-03 & 1.13e-02 & 2416706 & - & - & 41.2 & 113.1\\
\bottomrule
\end{tabular}
\end{scriptsize}
\end{center}
\end{sc}
%\vskip -0.1in
\end{table}
\begin{table}[h]
    \centering
    \begin{footnotesize}
    \begin{sc}
    \caption{{The per-step timing of our model against ground-truth simulators across datasets. Since CylinderFlow and DeformingPlate are datasets from MGN paper, we adopt their reported values for simulator timing ($t_{GT})$. The time of our model $t_{ours}$ is computed by adding the time used for segmentation and inference on a single NVIDIA Tesla P100 GPU.}}
    \label{tab: gt_simulator_compare}
    \begin{tabular}{lccccc}
    \toprule      
    Dataset & solver & $t_{GT}$ ms/step & $t_{ours}$ ms/step & speedup \\
    \midrule
        CylinderFlow & COMSOL & 820 & 20.4 & 40.2 \\
        DeformingPlate & COMSOL & 2893 & 29.7 & 97.4 \\
        DeformingBeam & OpenFOAM  & 3261 & 24.6 & 132.6 \\
    \bottomrule
    \end{tabular}
    \end{sc}
    \end{footnotesize}
    
\end{table}

\section{Computational Efficiency Analysis} \label{appendix: complexity}
% \YH{Discussion on Table~\ref{tab: ablation_complexity}}

\subsection{Performance Comparison}
Table~\ref{tab: ablation_complexity} lists the training time, test time, and memory usage for four models MGN, BSMS-GNN, EAGLE, and M4GN across three datasets. The RMSE-all is also listed as a performance reference. Our M4GN model has comparable or better efficiency compared with other models. Notably, the M4GN model has the best efficiency with RMSE-all compared to other baselines. {We also compare the per-step timing of our model against ground-truth simulators across datasets in Table~\ref{tab: gt_simulator_compare}. Since CylinderFlow and DeformingPlate are datasets from the MGN paper, we adopt their reported values for simulator timing ($t_{GT})$. The time of our model $t_{ours}$ is computed by adding the time used for segmentation and inference on a single NVIDIA Tesla P100 GPU. }

\subsection{Complexity Analysis}

M4GN is composed of four key components: an Encoder-Process-Decoder (EPD) network operating on mesh graphs, modal decomposition, hybrid mesh segmentation, and a mesh segment transformer. For the learnable part of the model, the computational complexity mainly depends on the EPD and mesh segment transformer components. The complexity of the EPD is: $O(L_1|\mathcal{V}|d^2 + L_1|\mathcal{E}|d^2)$, where $L_1$ is the number of message passing layers, $d$ is the feature dimension, $|\mathcal{V}|$ is the number of mesh nodes, and $|\mathcal{E}|$ is the number of mesh edges. The complexity of the mesh segment transformer is $O(L_2 K^2 d + L_2 K d^2)$, where $L_2$ is the number of multi-head attention layers, $K$ is the number of segments, and $d$ is the feature dimension. The overall time complexity is $O(L_1|\mathcal{V}|d^2 + L_1|\mathcal{E}|d^2 + L_2 K^2 d + L_2 K d^2)$.

{Modal decomposition and mesh segmentation are performed only once at the initial time step. For modal decomposition, with sparse finite–element matrices, the setup steps—basis construction, matrix  assembly, and application of boundary conditions—each require  $O(|\mathcal{V}|)$ time. The dominant cost is extracting the first $m$ eigenmodes via a Lanczos/Arnoldi  solver; because each Krylov iteration involves one sparse matrix–vector product,  the eigen-solve scales as $O(m\,\kappa\,|\mathcal{V}|)$, with $\kappa\approx10\text{–}100$ iterations per converged mode and $m$ the number of modes requested. Hence, for a fixed number of modes, the overall algorithm is linear in  mesh size. Our hybrid segmentation approach consists of a graph-based method for initial mesh segmentation (METIS) and a superpixel-based method guided by features for refinement (SLIC). In the case of Lagrangian systems where segmentation varies with time, only the refinement part is needed since the initialization segmentation is only based on connectivity and is invariant to feature and coordinate variations. SLIC  is based on a local search in k-means clustering, resulting in a time complexity of $O(|\mathcal{V}|)$}.

\section{Qualitative Results} \label{appendix: rollout}

\begin{figure*}[htbp]
    \centering
    \includegraphics[scale=0.1]{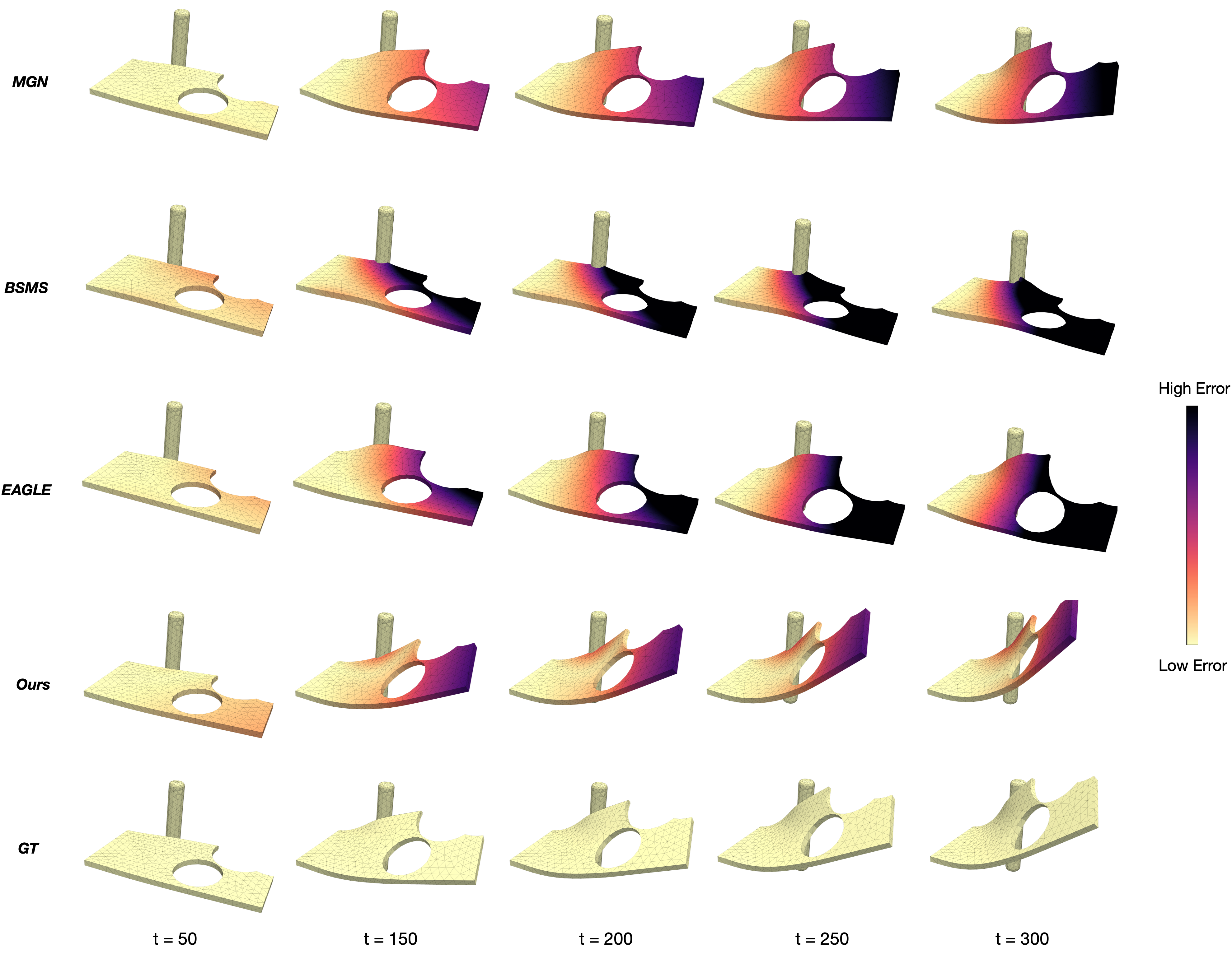}
    %\vskip -0.1in
    \caption{Additional simulation results for different models under DeformingPlate dataset.}
    \label{fig: deformingplate_rollout_1}
\end{figure*}

\begin{figure*}[htbp]
    \centering
    \includegraphics[scale=0.098]{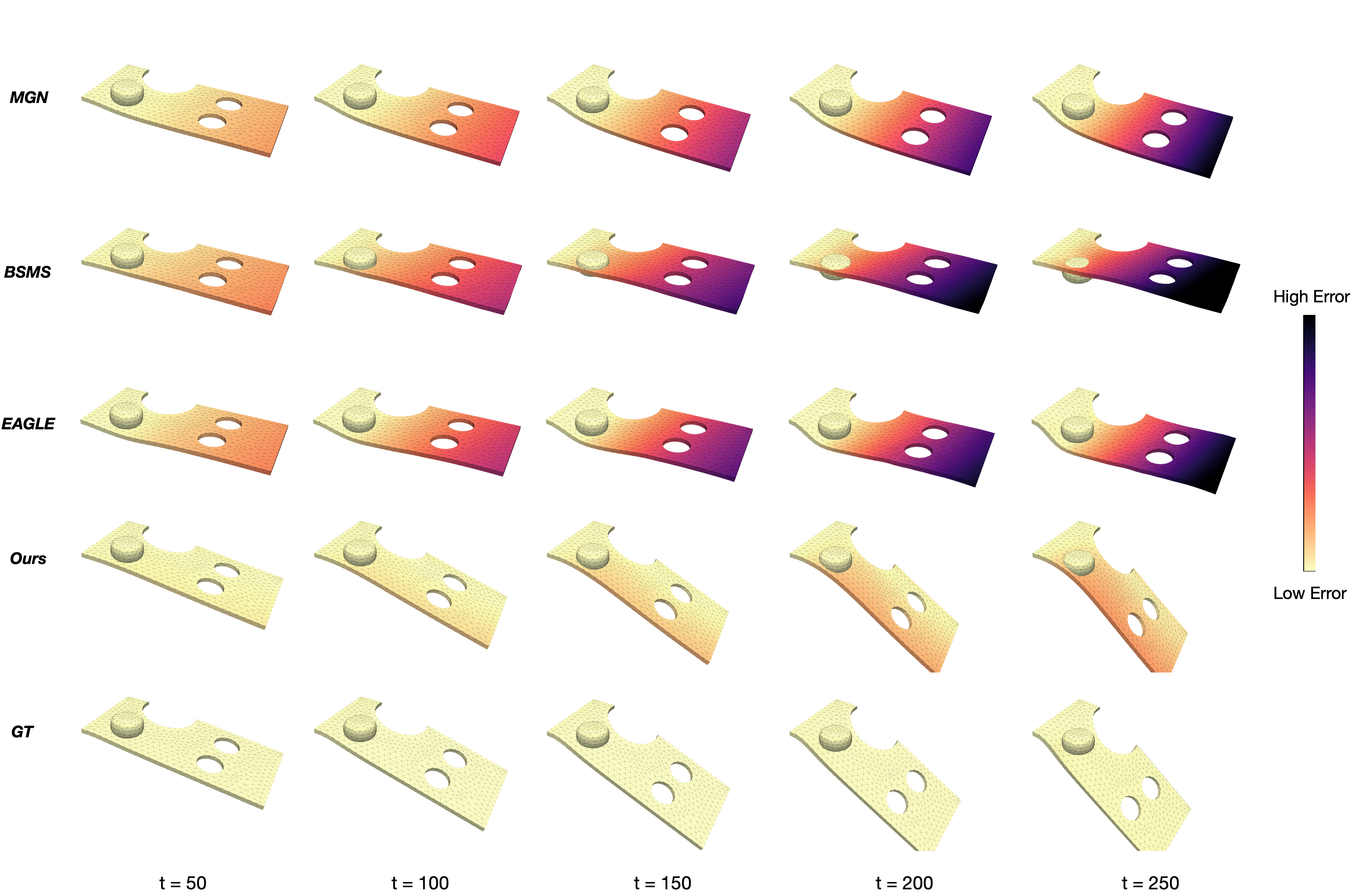}
    %\vskip -0.1in
    \caption{Additional simulation results for different models under DeformingPlate dataset.}
    \label{fig: deformingplate_rollout_2}
\end{figure*}

\begin{figure*}[htbp]
    \centering
    \includegraphics[scale=0.5]{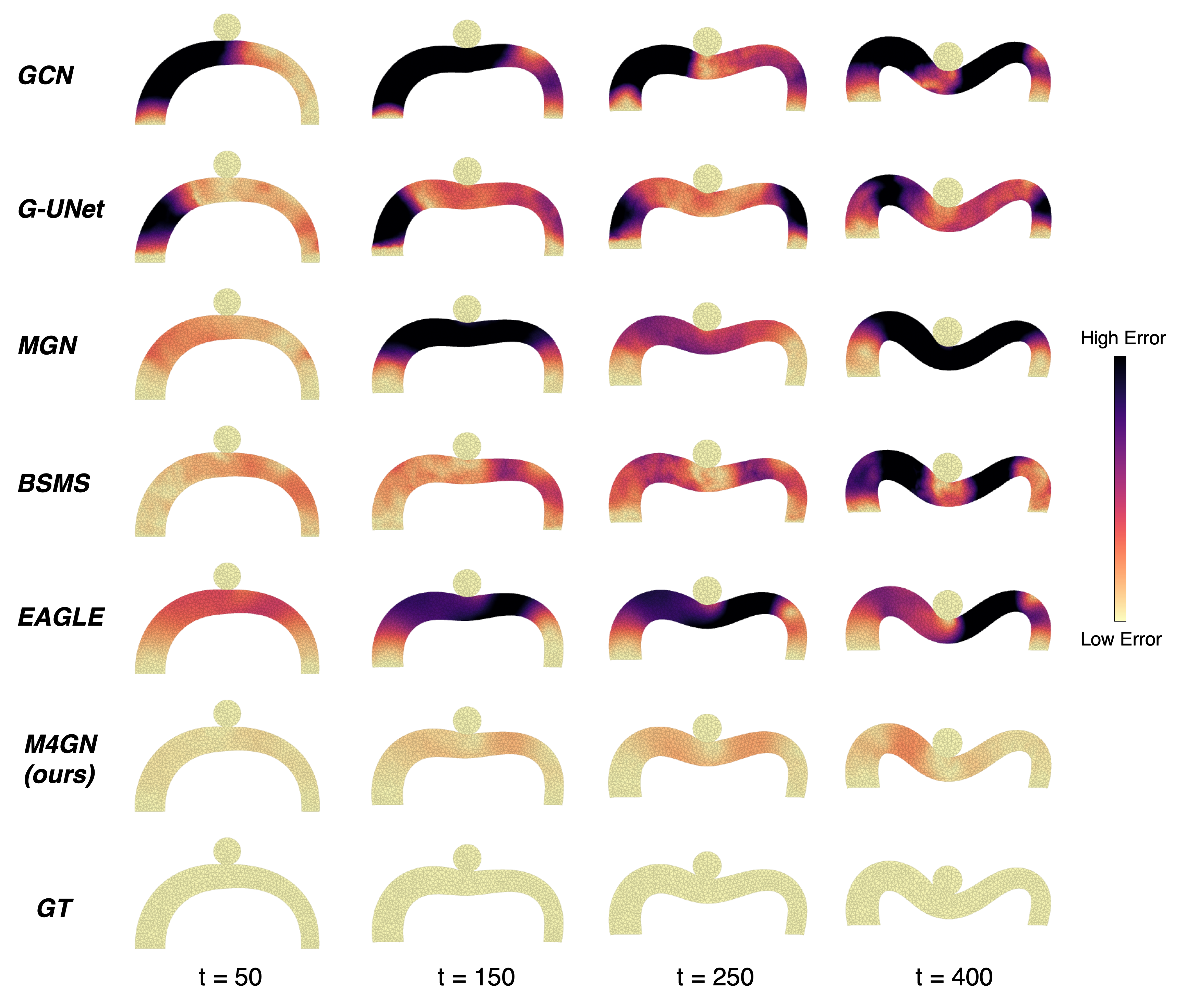}
    %\vskip -0.1in
    \caption{Additional simulation results for different models under DeformingBeam dataset}
    \label{fig: deformingbeam_rollout}
\end{figure*}

\begin{figure*}[htbp]
    \centering
    \includegraphics[scale=0.15]{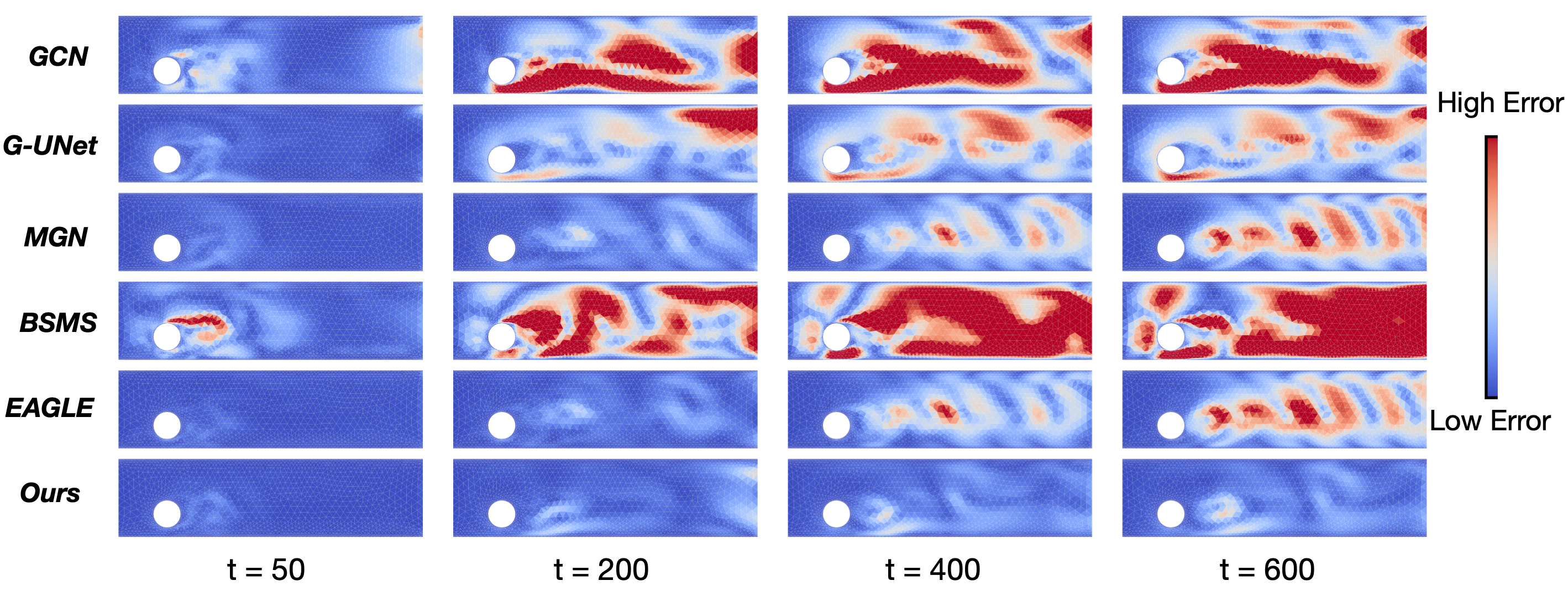}
    %\vskip -0.1in
    \caption{Additional simulation results for different models under CylinderFlow dataset}
    \label{fig: cylinderflow_rollout}
\end{figure*}

Figure~\ref{fig: deformingplate_rollout_1}, ~\ref{fig: deformingplate_rollout_2}, ~\ref{fig: deformingbeam_rollout}, and ~\ref{fig: cylinderflow_rollout} illustrate selected rollout results for all three datasets under different models. 

\end{document}